%% file: evorepair.tex
\PassOptionsToPackage{table,dvipsnames}{xcolor}
\documentclass{article} 
\usepackage{iclr2027_conference,times}
\usepackage[T1]{fontenc}

\input{math_commands.tex}

\usepackage{xcolor}
\usepackage{hyperref}
\ifdefined\pdfsuppressptexinfo\pdfsuppressptexinfo=-1\fi
\usepackage{url}
\usepackage{booktabs}
\usepackage{subcaption}
\usepackage{multirow}
\usepackage{tabularx}
\usepackage{graphicx}
\usepackage{float}
\usepackage{amsmath}
\usepackage{wrapfig}
\usepackage{needspace}
\usepackage{colortbl}
\usepackage[most]{tcolorbox}
\usepackage{listings}
\usepackage{pifont}
\usepackage{algorithm}
\usepackage{algpseudocode}
\usepackage{amssymb}
\usepackage{fontawesome5}
\usepackage{tikz}
\definecolor{linktext}{HTML}{1F3A63}
\newcommand{\hflogo}{\raisebox{-0.2\height}{\includegraphics[height=1.15em]{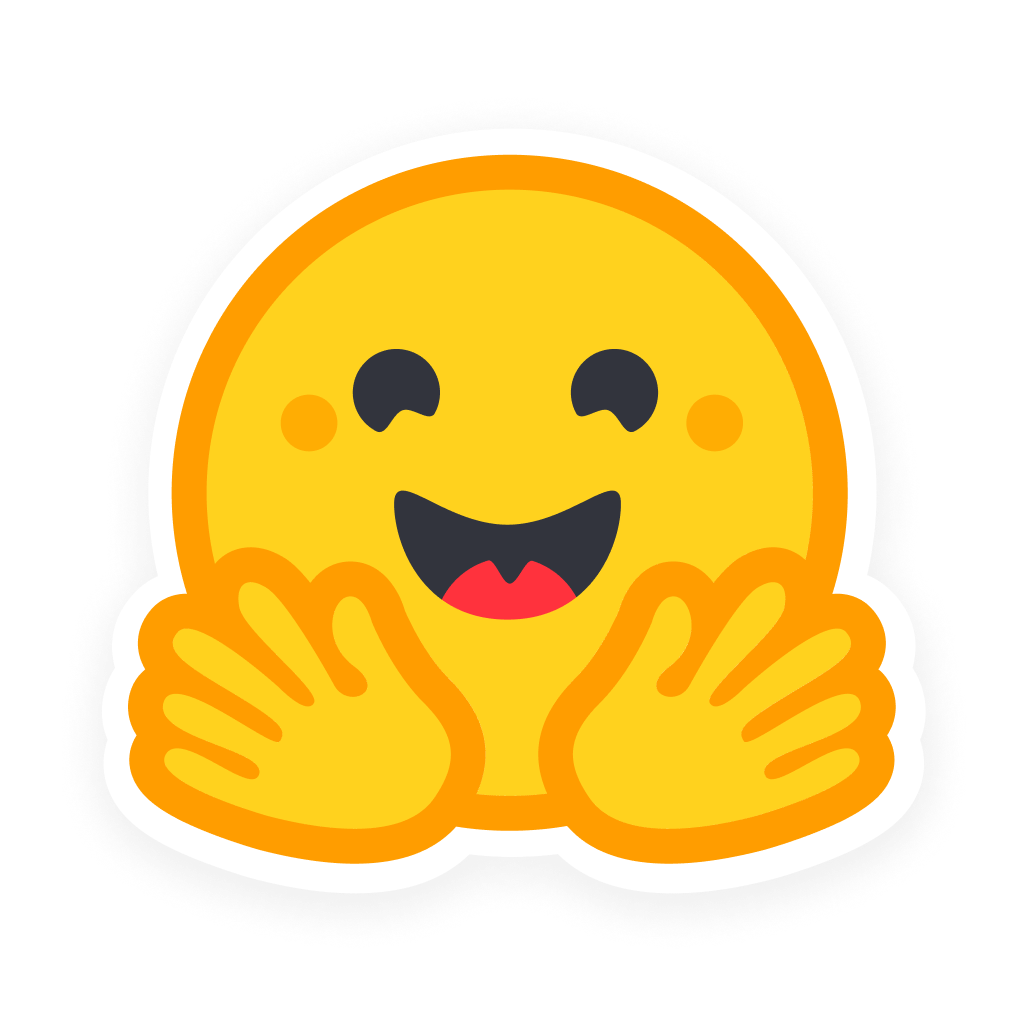}}}
\newcommand{\headlogo}[2]{\raisebox{-0.12\height}{\includegraphics[height=#1]{#2}}}
\newcommand{\institutionlogos}{%
  \headlogo{12pt}{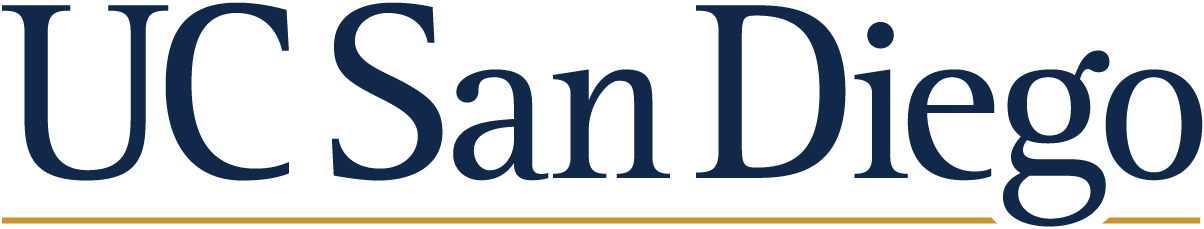}\hspace{9pt}%
  \headlogo{12pt}{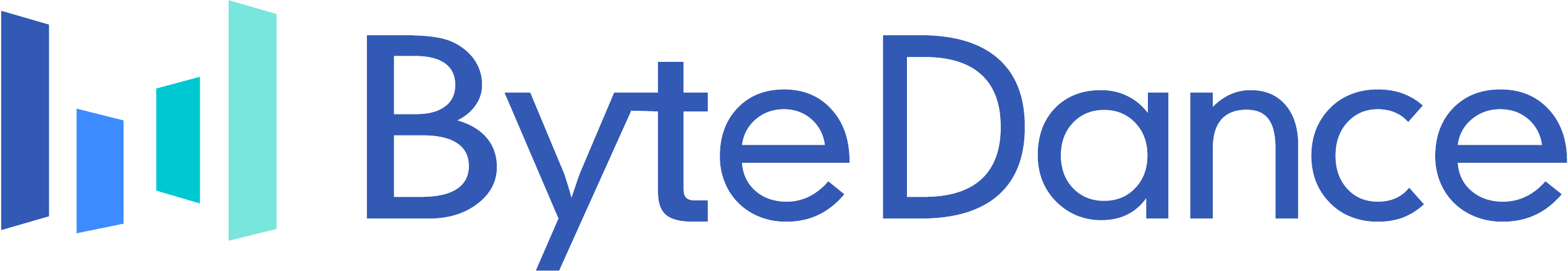}\hspace{9pt}%
  \headlogo{10pt}{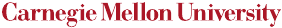}}
\newcommand{\linkbadge}[3]{%
  \begingroup\hypersetup{pdfborder={0 0 0}}%
  \href{#3}{{\small\bfseries#1\hspace{4pt}\textcolor{linktext}{#2}}}%
  \endgroup}

\definecolor{RegRed}{HTML}{C4442E}

\definecolor{PromptAccent}{HTML}{1664FF}
\definecolor{PromptBack}{HTML}{F7F9FE}
\definecolor{PromptLine}{HTML}{D8E2F5}
\definecolor{PromptInk}{HTML}{1F2430}
\newtcblisting{promptlisting}[1][]{%
  enhanced, breakable, listing only,
  listing engine=listings,
  listing options={basicstyle=\ttfamily\scriptsize, breaklines=true,
                   breakatwhitespace=true, columns=fullflexible,
                   keepspaces=true, showstringspaces=false, upquote=true},
  colback=PromptBack, colframe=PromptLine, coltitle=PromptInk,
  fonttitle=\bfseries\footnotesize,
  boxrule=0.45pt, arc=1.2mm,
  left=2mm, right=2mm, top=1.1mm, bottom=1.1mm,
  borderline west={1.2pt}{0pt}{PromptAccent!70},
  #1}

\definecolor{PromptPlan}{HTML}{EFF5FC}
\definecolor{PromptProbe}{HTML}{FCF5E8}
\definecolor{PromptRepair}{HTML}{F4F0FA}
\definecolor{PromptVerify}{HTML}{EFF8F1}
\definecolor{PromptCompare}{HTML}{ECF8F7}
\newtcblisting{rsipromptbox}[2]{%
  enhanced, breakable, listing only,
  listing engine=listings,
  listing options={basicstyle=\ttfamily\scriptsize, breaklines=true,
                   breakatwhitespace=true, columns=fullflexible,
                   keepspaces=true, showstringspaces=false, upquote=true},
  colback=#1, colbacktitle=#1,
  colframe=PromptAccent!22, coltitle=PromptInk,
  boxrule=0.45pt, arc=2pt,
  left=8pt, right=8pt, top=6pt, bottom=6pt,
  before skip=9pt, after skip=10pt,
  fonttitle=\small\bfseries,
  title={#2}}

\definecolor{EvoHighlight}{HTML}{DCEBF7}
\newcommand{\evorow}{\rowcolor{EvoHighlight}}
\definecolor{EvoBest}{HTML}{8FC1E3}
\newcommand{\evobest}{\rowcolor{EvoBest}}
\title{RSIGame: Autonomous Agentic Game Development with Recursive Self-improvement}

\newcommand{\authorsep}{\hspace{1em}}
\author{%
Wenyi Wu$^{1*}$\authorsep
Minghao Fu$^{1,2*}$\authorsep
Jieyu You$^{2}$\authorsep
Kun Zhou$^{1\dagger}$\authorsep
Siqi Liu$^{1}$\\
\bf Aayush Salvi$^{1}$\authorsep
Yiheng Lin$^{2}$\authorsep
Ce Zhang$^{3}$\authorsep
Xiaohan Lan$^{2}$\authorsep
Jiahui Zhu$^{2}$\\
\bf Yujie Zhong$^{2\dagger}$\authorsep
Qi She$^{2}$\authorsep
Biwei Huang$^{1}$\\[3pt]
{\normalfont $^{1}$University of California San Diego\quad $^{2}$ByteDance Inc.\quad $^{3}$Carnegie Mellon University}%
}

\newcommand{\rsigame}{\textsc{RSIGame}}

\iclrfinalcopy 
\begin{document}

\maketitle
\lhead{}
\rhead{\ifnum\value{page}=1 \institutionlogos\fi}
\renewcommand{\headrulewidth}{\ifnum\value{page}=1 0.4pt\else 0pt\fi}
{\renewcommand{\thefootnote}{}\footnotetext{%
$^{*}$Equal contribution.\quad $^{\dagger}$Corresponding authors.\quad
Contact: \texttt{franciskunzhou@gmail.com}.}}
\vspace{-1.1em}

\begin{abstract}
Recent advances in large language models have made automatic game generation increasingly feasible, yet reliably improving generated games beyond a playable version remains challenging. Naive iterative refinement can easily overfit a small set of test cases, producing fragile games with unresolved bugs, missing behaviors, and poor generalization to broader player interactions. We introduce \rsigame{}, an autonomous agentic game development framework with recursive self-improvement. \rsigame{} organizes development into complementary local and global loops. Concretely, a local \emph{explore-diagnose-improve} loop broadly explores the executable game, diagnoses and prioritizes discovered issues, and performs evidence-grounded revision, where an evolving checklist continually accumulates new testing and improvement guidance. A global loop tracks overall quality, preserves the best checkpoint, and detects saturation or regression over long-horizon development. Beyond test-time improvement, \rsigame{} further internalizes successful development experience into the generator through training. Across 140 GameCraft-Bench tasks, two game engines, and five generators, \rsigame{} consistently improves game quality under matched development budgets. 
Notably, experience internalization enables Qwen3.8-27B to reach 61.38 on Godot and 58.53 on Phaser, exceeding GPT-5.5 one-shot scores while reducing Qwen's generation tokens by \(11\) times.

\end{abstract}

\par\vspace{0.3em}
{\centering
\linkbadge{\faGithub}{Code}{https://github.com/WenyiWU0111/RSIGame}%
\hspace{2em}%
\linkbadge{\hflogo}{Dataset}{https://huggingface.co/RSIGame}%
\hspace{2em}%
\linkbadge{\faGlobe}{Project Page}{https://huggingface.co/spaces/RSIGame/rsigame-page}\par}
\vspace{0.2em}

\begin{figure}[h]
    \centering
    \includegraphics[width=\linewidth]{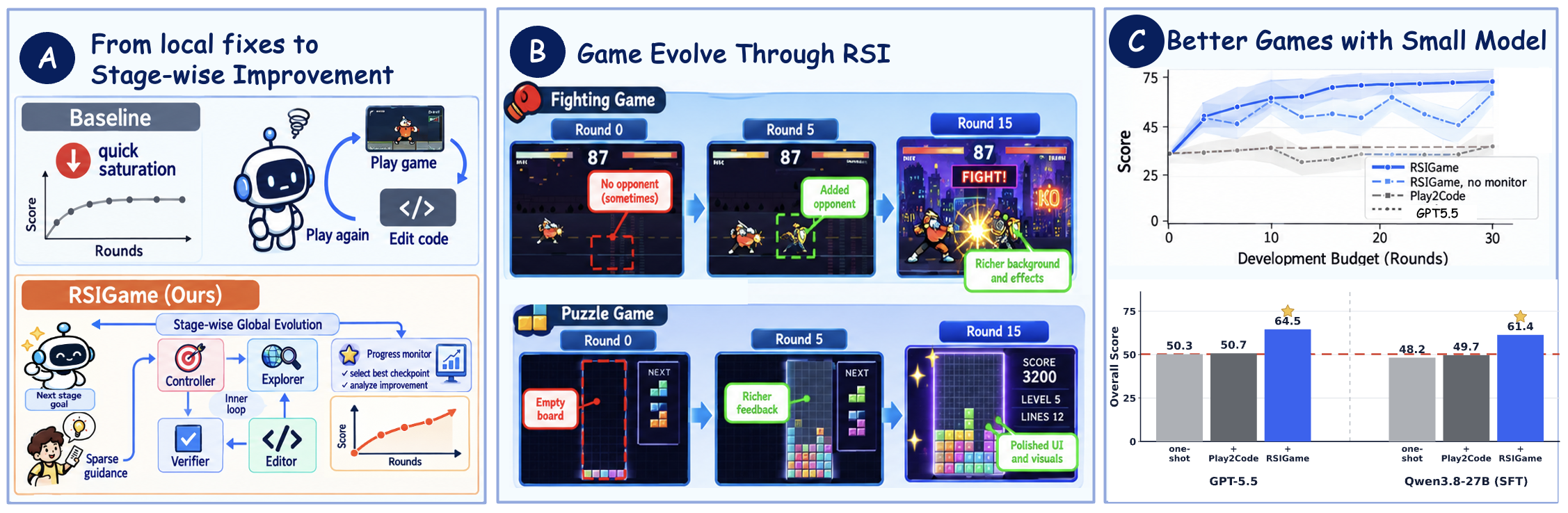}
    \caption{
\textbf{\rsigame{} turns agentic game development into autonomous recursive self-improvement.} It combines a local explore-diagnose-improve loop with global progress monitoring and control, enabling Qwen3.8-27B approach GPT-5.5-level performance.
}
    \label{fig:intro}
\end{figure}

\section{Introduction}
Recent advances in large language models (LLMs) and vision-language models (VLMs), have substantially expanded the capability of autonomous agents to perform complex real-world tasks, ranging from software engineering to computer-use automation~\cite{sweagent,rsiagent}. Automatic game development represents another promising domain, as building a complete game requires jointly designing game mechanics and progression, create source code and multimodal assets, and organize these heterogeneous components into a coherent executable project~\cite{gamecraftbench}. It has become increasingly realistic with the rapid improvement of long-horizon generation, multimodal understanding, and tool-use capabilities in foundation models. 
Accordingly, recent benchmarks show that VLM-based agents can already turn natural-language specifications into playable games with recognizable mechanics and visual content~\citep{gamecraftbench,gamedevbench}.

However, game development remains highly challenging due to its nature as a complex multimodal engineering process. Even in mature human teams, unexpected bugs, missing functionalities, and overlooked corner cases frequently arise during development~\cite{roque2025gametesting}. A common solution is therefore to adopt an iterative development workflow, in which developers repeatedly test the current build, collect feedback, and revise the project until it reaches the desired quality. This process can be naturally viewed as a form of recursive self-improvement~\cite{metarewarding}, where each development round builds upon the previous version based on newly observed evidence. Inspired by this workflow, we formulate agentic game development as an iterative loop: an agent first generates an initial game, 
then repeatedly tests and improves it until the target requirements are met.

Despite its natural fit for agentic game development, we find that recursive self-improvement can easily converge to fragile solutions that roughly satisfy the target while leaving many untested bugs, missing behaviors, and corner cases unresolved. In effect, the development loop may overfit a small set of test cases rather than improve the game as a whole, leading to poor generalization to broader player interactions. To address this issue, we strengthen the testing stage to broadly explore remaining failures and potential improvements, with the goal of developing a high-quality game rather than merely a playable one. Concretely, we devise an \emph{\textbf{explore-diagnose-improve loop}}: the agent first broadly explores the executable game to uncover bugs, weaknesses, and opportunities for enhancement; it then diagnoses the discovered issues and prioritizes them to plan the next round of improvement. Throughout this process, we maintain an \emph{evolving checklist} that continually accumulates newly identified issues and actionable guidance for subsequent testing and revision. By expanding as new evidence is discovered, this checklist persistently pushes the development process beyond the limited test cases and toward more generalizable improvements in overall game quality.

\begin{figure}[t]
\centering
\includegraphics[width=0.98\textwidth]{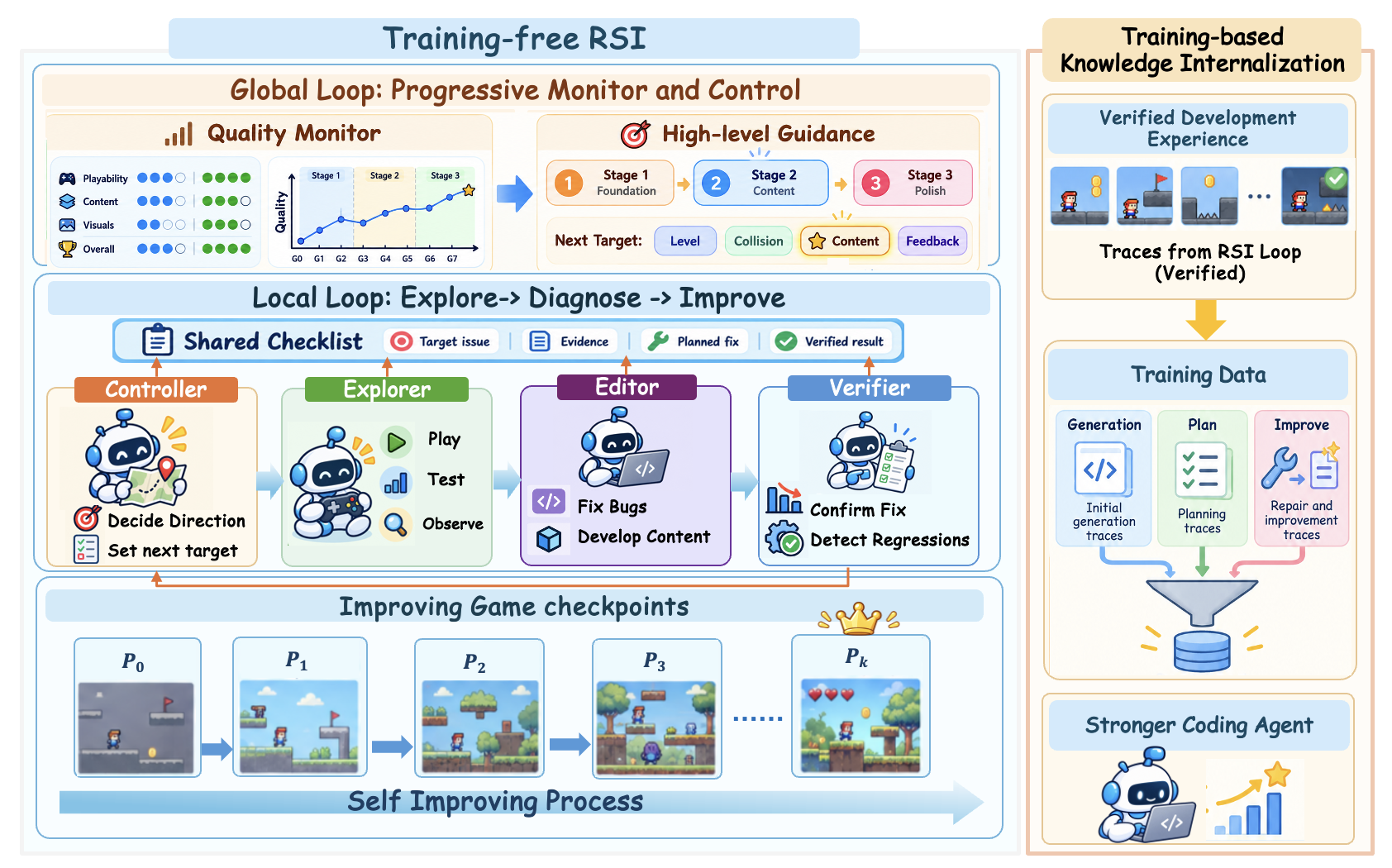}
\caption{Overview of \rsigame{}. 
The local loop autonomously evolves the game through direction decision, active exploration, evidence-grounded editing, and agentic verification. The global loop monitors global quality, preserves the best checkpoint, and introduces sparse high-level guidance when progress saturates, enabling progressive multi-stage game evolution.}
\label{fig:overview}
\end{figure}

To this end, we introduce \rsigame{} (Figure~\ref{fig:overview}), an autonomous agentic game development framework with recursive self-improvement. At its core, \rsigame{} organizes development into complementary local and global loops. The local loop instantiates our explore-diagnose-improve process, repeatedly exploring the executable game, diagnosing and prioritizing discovered issues, and revising the project based on accumulated evidence. The global loop monitors development across iterations, tracks overall game quality, preserves the best checkpoint, and detects saturation or regression, enabling the system to maintain long-horizon progress rather than blindly continue editing. Beyond training-free self-improvement, \rsigame{} further introduces training-based experience internalization, converting successful development trajectories, diagnostic decisions, and verified revisions into supervision for the generator. Such a way enables a broader RSI loop to boost the underlying generator. Across 140 GameCraft-Bench tasks, two engines, and five generators, \rsigame{} consistently improves games under matched development budgets. 
With iterative development, Qwen3.8-27B reaches 47.77 on Godot versus Kimi-K2.6's 44.77~\citep{kimik26}; experience internalization lifts it to 61.38 versus Codex$+$GPT-5.5's 50.26 one-shot score~\citep{codex,gpt55}, while cutting generation tokens by \(11\) times.
On Phaser, it rises to 50.24, past the 49.44 one-shot score of GPT-5.5.
\section{Preliminaries}
\label{sec:preliminaries}

\paragraph{Agentic Game Development.}
It aims to automate game creation with intelligent agents. Given a natural-language game specification, an agent translates user intent into a complete and playable game.
Similar to human game development, agentic game development must satisfy two objectives: following user requirements and producing a functionally correct, bug-free game. Formally, given a game specification $x$, we aim to learn a generator policy $\pi_\theta$ that produces a game project
\begin{equation}
P_0 \sim \pi_\theta(\cdot \mid x),
\end{equation}
where $P_0$ should both conform to the intended game design and execute correctly.
Such a project consists of source code, visual and textual assets, and configuration files. Its construction therefore requires joint generation and integration of coding, text, and visual elements into a coherent interactive system.
Importantly, game quality is ultimately reflected in runtime behavior and player experience. Each generated game includes a set of replayable demonstration traces
\(\mathcal{T}=\{\tau_1,\tau_2,\ldots,\tau_N\}\), which are replayed during evaluation and scored against a hidden task-specific rubric.

\paragraph{Our Focus.}
Because game generation involves many tightly coupled components, producing a fully correct project in one pass is difficult. Existing workflows therefore commonly adopt an iteration-based strategy: $P_0 \rightarrow P_1 \rightarrow \cdots \rightarrow P_T$. At each iteration, the current project is executed and tested, observed failures are diagnosed, and the project is revised accordingly. In conventional game development, this loop relies on human developers and playtesters. In this work, we bring the same principle into agentic game development, but replace human-driven testing and revision with a \emph{fully autonomous process} that recursively tests, diagnoses, and improves its own generated game.



\section{\rsigame{}}
\label{sec:method}

Our \rsigame{} incorporates both training-free and training-based strategies. The training-free strategy builds on local and global loops for iterative development, while the training-based strategy further internalizes accumulated development experience into the backbone game generation model.



\subsection{Training-Free Recursive Self-Improvement}
\label{sec:training_free}
Our training-free RSI strategy consists of a local loop that follows an iterative explore-diagnose-improve process to identify issues and refine the game, and a global loop that monitors long-horizon progress, preserves the best, and detects saturation.

\subsubsection{Local Loop: Explore-Diagnose-Improve}
\label{sec:local_loop}
The local loop performs fine-grained recursive game improvement through three stages: autonomous exploration, issue diagnosis, and iterative improvement. The loop continuously explores the executable game, accumulates newly discovered issues and improvement opportunities in an evolving checklist, and uses them to guide subsequent revisions.

\paragraph{Controller and Explorer for Autonomous Exploration.}
The autonomous exploration stage involves two agents with complementary roles. The controller is responsible for deciding what aspect of the game should be explored next, while the explorer interacts with the executable game to collect concrete behavioral evidence. Specifically, given the game specification $x$, the current project $P_t$, the development checklist $\mathcal{C}_t$, and optional stage-level guidance $\gamma_s$, the controller proposes a development direction $d_t$. Conditioned on this direction, the explorer performs targeted interaction with the game and produces an exploration trajectory $\tau_t$:
\begin{equation}
d_t \sim \pi_{\mathrm{controller}}(\cdot \mid x,P_t,\mathcal{C}_t,\gamma_s),
\qquad
\tau_t \sim \pi_{\mathrm{explorer}}(\cdot \mid P_t,d_t).
\end{equation}
Through this process, the controller provides high-level exploration intent, while the explorer grounds it in observed gameplay, uncovering concrete bugs, missing behaviors, and potential improvement opportunities for subsequent diagnosis and revision.


\paragraph{Editor and Verifier for Issue Diagnosis.}
The issue diagnosis stage involves two agents with complementary responsibilities. The editor converts the discovered issues and interaction evidence into concrete game modifications, while the verifier evaluates whether these modifications resolve the intended problems without introducing regressions. Specifically, conditioned on the current project, development direction, and observed playtest trajectory, the editor proposes an edit
\begin{equation}
\Delta_t \sim
\pi_{\mathrm{editor}}(\cdot \mid P_t,d_t,\tau_t),
\qquad
P_{t+1}=\operatorname{Apply}(P_t,\Delta_t).
\end{equation}
The verifier then interacts with the updated project and produces a verification outcome
\begin{equation}
v_t \sim
\pi_{\mathrm{verifier}}(\cdot \mid P_{t+1},d_t,\tau_t),
\end{equation}
which determines whether the targeted issue has been successfully addressed and whether the edit causes unintended side effects. The verified outcome is subsequently written back to the evolving checklist, providing updated evidence for the controller to plan the next development round.

\paragraph{Iterative Improvement with Checklist Update.}
To connect exploration, diagnosis, and revision across iterations, \rsigame{} maintains a shared development checklist \(\mathcal{C}_t\) as an evolving working state. The checklist records discovered issues, improvement opportunities, priorities, and verification outcomes, allowing all agents to operate on a consistent view of the current development status. At each round, the controller reads \(\mathcal{C}_t\) to determine the next development direction \(d_t\), the explorer augments it with newly observed evidence, the editor addresses prioritized items, and the verifier updates their status according to the resulting gameplay behavior.
These interactions produce the next-round checklist
\begin{equation}
\mathcal{C}_{t+1}=U(\mathcal{C}_t,d_t,\tau_t,\Delta_t,v_t),
\end{equation}
which is passed to the controller together with \(P_{t+1}\). In this way, \rsigame{} continually accumulates development knowledge and ensures that each iteration builds on verified progress rather than restarting from scratch.




\subsubsection{Global Loop: Progress Monitoring and Control}
\label{sec:global_loop}
While the local loop focuses on improving the game within each development round, the global loop governs the overall development process across stages. It consists of two components: the game quality monitor that tracks accumulated game quality and preserves the strongest checkpoint, and the progress and convergence control mechanism that determines whether development should continue, terminate, or enter a new stage under additional high-level guidance.

\paragraph{Game Quality Monitor.}
The game quality monitor tracks game-level progress across local development rounds. A stage begins from the checkpoint the previous stage retained, $P_s^\star \gets P_{s-1}^\star$. At each global evaluation point, the monitor compares the checkpoint just produced against the retained one and updates
\begin{equation}
P_s^\star \gets
\operatorname{SelectBest}
\left(
P_s^\star,P_{t+1}
\right).
\end{equation}
Unlike the local Verifier, which assesses whether a specific edit resolves its targeted issue, the game quality monitor evaluates the accumulated quality of the game as a whole and maintains a persistent best state throughout development. Importantly, this monitoring process is strictly isolated from the benchmark evaluator, including its scores, rubrics, and feedback. Details are in Appendix~\ref{app:quality_monitor}.

\paragraph{Progress and Convergence Control.}
The evolution of the retained checkpoint $P_s^\star$ provides a natural signal for controlling long-horizon development. If successive global evaluations fail to produce a better checkpoint, \rsigame{} treats the persistent lack of progress as convergence. The system then either terminates and returns $P_s^\star$ as the final game, or optionally introduces new high-level guidance $\gamma_{s+1}$ from a human or stronger model. Such guidance provides strategic directions or creative suggestions rather than explicit edit instructions. The next development stage therefore starts from the retained best checkpoint $P_s^\star$ under $\gamma_{s+1}$, allowing the local loop to explore new improvement directions without discarding previously verified gains. Repeating this process enables \rsigame{} to control long-horizon progress while preserving the strongest solution reached so far.

\subsection{Training-Based Knowledge Internalization}
\label{sec:co_evolution}
During training-free RSI, \rsigame{} also accumulates reusable experience about how games are planned, implemented, tested, and refined. Thus, we further internalize successful RSI knowledge into the backbone model, extending self-improvement from context optimization to parameter optimization. This broader RSI loop improves the underlying generator, enabling stronger initial generation and reducing the number of refinement iterations required for high-quality game development.

\paragraph{Development Experience Collection and Curation.}
We collect three types of supervision: \emph{generation traces} from GPT-5.5~\citep{gpt55} constructing games in Godot~\citep{godot} and Phaser~\citep{phaser}, \emph{planning traces} collected from successful generations, and \emph{improvement traces} produced by running \rsigame{} with GLM-5.3-Flash~\citep{glm53}. We retain only executable generations and improvements that are verified through post-edit interaction without breaking previously functional behavior. In total, we obtain 2{,}213 generation traces, 2{,}108 planning traces, and 2{,}013 verified improvement rounds from 4{,}003 candidates.

\paragraph{Experience Internalization.}
We apply supervised fine-tuning to Qwen3.8-27B~\citep{qwen38} on the curated planning, generation, and improvement trajectories. Beyond learning from final game artifacts, the model internalizes the intermediate planning, tool-use, diagnosis, and revision decisions that drive successful development. This transfers reusable RSI knowledge into model parameters, strengthening the refinement ability and reducing the required number of iterations to reach high-quality game generation. Such a way opens the door to continual RSI, in which newly acquired development experience can be repeatedly internalized to further improve the generator over time. Training details are provided in Appendix~\ref{app:training}.

\begin{table}[t]
\centering
\caption{
\textbf{Main results on GameCraft-Bench (top: Godot; bottom: Phaser).}
Methods are compared from matched base games under one backbone and budget, averaged over 140 tasks;
\textbf{Tok.} and \textbf{Cost} are mean billable tokens and cost per task (Appendix~\ref{app:cost_and_protocol}).
\textbf{Bold} marks the best value per column within a group;
{\setlength{\fboxsep}{1pt}\colorbox{EvoBest}{deeper}} rows internalize development experience.
}
\label{tab:godot_main}\label{tab:phaser_main}
\small

\begin{tabular*}{\textwidth}{@{\extracolsep{\fill}}lcccccrr}
\toprule
Method
& Mechanics
& Depth
& Visuals
& Art
& Overall $\uparrow$
& Tok. $\downarrow$
& Cost $\downarrow$ \\
\midrule
\multicolumn{8}{c}{\textbf{Godot}~\citep{godot}} \\
\midrule

\multicolumn{8}{l}{\emph{Generator: Codex $+$ GPT-5.5 (high)}} \\
Base (frozen $P_0$)
& 58.9 & 51.7 & 51.4 & 44.6 & 50.26 & 0.26M & -- \\
\quad + Play2Code
& 59.2 & 51.3 & 52.2 & 45.8 & 50.74 & 3.20M & \$1.06 \\
\evorow \quad + \rsigame{}
& \textbf{72.8} & \textbf{60.3} & \textbf{65.9} & \textbf{64.5} & \textbf{64.53} & \textbf{3.12M} & \textbf{\$0.88} \\

\midrule
\multicolumn{8}{l}{\emph{Generator: Codex $+$ Kimi-K2.6}} \\
Base (frozen $P_0$)
& 40.2 & 30.3 & 35.4 & 21.9 & 29.63 & 3.22M & -- \\
\quad + Play2Code
& 44.8 & 35.4 & 40.2 & 28.6 & 35.19 & \textbf{7.07M} & \$1.27 \\
\evorow \quad + \rsigame{}
& \textbf{51.7} & \textbf{40.1} & \textbf{47.1} & \textbf{45.4} & \textbf{44.77} & 7.39M & \textbf{\$1.18} \\

\midrule
\multicolumn{8}{l}{\emph{Generator: Codex $+$ GLM-5.3-Flash}} \\
Base (frozen $P_0$)
& 35.7 & 30.2 & 32.0 & 24.9 & 29.46 & 0.44M & -- \\
\quad + Play2Code
& 47.2 & 39.7 & 41.1 & 32.7 & 38.59 & \textbf{3.18M} & \textbf{\$0.99} \\
\evorow \quad + \rsigame{}
& \textbf{51.9} & \textbf{44.4} & \textbf{49.9} & \textbf{53.9} & \textbf{49.72} & 5.06M & \$1.25 \\

\midrule
\multicolumn{8}{l}{\emph{Generator: Codex $+$ Qwen3.8-27B}} \\
Base (frozen $P_0$)
& 41.2 & 33.4 & 38.6 & 38.3 & 37.07 & 6.41M & -- \\
\quad + Play2Code
& 44.6 & 36.1 & 42.1 & 42.5 & 40.53 & 9.92M & \$1.32 \\
\evorow \quad + \rsigame{}
& \textbf{51.2} & \textbf{40.4} & \textbf{47.5} & \textbf{53.8} & \textbf{47.77} & \textbf{9.13M} & \textbf{\$1.09} \\

\midrule
\multicolumn{8}{l}{\emph{Generator: Codex $+$ Qwen3.8-27B (SFT)}} \\
Base (frozen $P_0$)
& 56.1 & 47.5 & 49.8 & 44.8 & 48.22 & 0.57M & -- \\
\quad + Play2Code
& 57.4 & 48.1 & 51.6 & 47.2 & 49.71 & 3.74M & \$1.08 \\
\evobest \quad + \rsigame{}
& \textbf{68.7} & \textbf{55.8} & \textbf{62.8} & \textbf{63.2} & \textbf{61.38} & \textbf{3.49M} & \textbf{\$0.94} \\

\midrule[\heavyrulewidth]
\multicolumn{8}{c}{\textbf{Phaser}~\citep{phaser}} \\
\midrule

\multicolumn{8}{l}{\emph{Generator: OpenGame $+$ GPT-5.5}} \\
Base (frozen $P_0$)
& 49.1 & 38.6 & 53.6 & 58.6 & 49.44 & 5.36M & -- \\
\quad + Play2Code
& 57.1 & 42.0 & 60.0 & 65.2 & 55.10 & \textbf{8.34M} & \textbf{\$1.43} \\
\evorow \quad + \rsigame{}
& \textbf{59.8} & \textbf{44.2} & \textbf{62.4} & \textbf{69.7} & \textbf{58.21} & 8.38M & \$1.44 \\
\midrule
\multicolumn{8}{l}{\emph{Generator: OpenGame $+$ Qwen3.8-27B}} \\
Base (frozen $P_0$)
& 40.6 & 29.5 & 43.7 & 48.7 & 40.03 & 9.03M & -- \\
\quad + Play2Code
& 51.0 & 36.4 & 53.2 & 58.0 & 48.70 & 12.99M & \$1.75 \\
\evorow \quad + \rsigame{}
& \textbf{52.0} & \textbf{36.5} & \textbf{54.6} & \textbf{61.3} & \textbf{50.24} & \textbf{12.78M} & \textbf{\$1.50} \\

\midrule
\multicolumn{8}{l}{\emph{Generator: OpenGame $+$ Qwen3.8-27B (SFT)}} \\
Base (frozen $P_0$)
& 47.1 & 35.4 & 48.9 & 52.4 & 45.14 & 5.42M & -- \\
\quad + Play2Code
& 55.6 & 41.7 & 57.3 & 61.8 & 53.15 & 9.58M & \$1.61 \\
\evobest \quad + \rsigame{}
& \textbf{60.4} & \textbf{46.2} & \textbf{62.3} & \textbf{68.4} & \textbf{58.53} & \textbf{9.31M} & \textbf{\$1.53} \\
\bottomrule
\end{tabular*}

\end{table}

\section{Experiments}

\subsection{Experimental Setup}

\paragraph{Benchmark and Engines.}
We evaluate on GameCraft-Bench~\citep{gamecraftbench}, which contains 140
game-development tasks spanning 15 game families. We instantiate the same tasks in both Godot~\citep{godot} and Phaser~\citep{phaser}, keeping the task specifications, evaluation rubrics, and development protocol fixed across
engines. Phaser therefore serves as a cross-engine robustness evaluation of the findings on Godot.

\paragraph{Evaluation Protocol.}

Following GameCraft-Bench, each generated game is packaged with replayable demonstration traces provided by the game generator, which are replayed and scored against a hidden
task-specific rubric over \emph{Mechanics} (\(M\)), \emph{Depth} (\(D\)),
\emph{Visuals} (\(V\)), and \emph{Art} (\(A\)):
\[
Q = \mathrm{BUILD}\times(0.15M+0.35D+0.15V+0.35A),
\]
where \(\mathrm{BUILD}\in\{0,1\}\) assigns zero score to non-playable games.
We use Qwen3.8-27B~\citep{qwen38} as the judge and average three independent replay-and-score runs. The hidden rubric, scores, and judge feedback are never exposed to the development agents. Evaluation stability and free-play robustness are detailed in Appendices~\ref{app:stability} and~\ref{app:freeplay}.

\paragraph{Baselines and Development Budget.}

We compare the frozen initial project \(P_0\), with Play2Code~\citep{play2code},
and with \rsigame{} across multiple generators. For each generator, all development
methods start from an identical clone of \(P_0\) and use the same per-round tool
budget. Both Play2Code~\citep{play2code} and \rsigame{} are allowed up to 26 per-round tool calls and 30 development rounds. Full model configurations, row-specific budgets, cost accounting, and implementation details are provided in Appendix~\ref{app:config} and Appendix~\ref{app:cost}.

\subsection{Godot Game Results}
\label{sec:godot_results}

Table~\ref{tab:godot_main} reveals three main findings.
\textit{First, \rsigame{} delivers large and consistent gains across generators.}
Across all five settings, it improves the frozen initial games by
\(10.7\)--\(20.3\) Overall points, with gains consistently spanning Mechanics,
Depth, Visuals, and Art.
\textit{Second, 
under the same compute budget, \rsigame{} performs significantly better than other recursive-based methods.
}
Under matched development budgets, \rsigame{} outperforms Play2Code by
\(7.2\)--\(13.8\) Overall points. Play2Code brings almost no improvement to the
strong Codex+GPT-5.5 initialization and 
Qwen3.8-27B variants, whereas \rsigame{} substantially improves the same frozen games. This contrast shows that the benefit comes from how development is organized, rather than from iteration alone.
\textit{Finally, experience internalization further improves both quality and efficiency.}
Fine-tuning raises Qwen3.8-27B's one-shot score from \(37.07\) to \(48.22\), within \(2.0\) points of Codex$+$GPT-5.5, while reducing generation tokens by \(11\) times (\(6.41\)M to \(0.57\)M). With \rsigame{}, the score further rises to \(61.38\) and total token usage decreases by \(2.6\) times (\(9.13\)M to \(3.49\)M).

\subsection{Phaser Game Results}

The Phaser block of Table~\ref{tab:phaser_main} shows that the gains of \rsigame{} transfer across engines, improving all three generators by \(8.8\)--\(13.4\) Overall points over their frozen bases and achieving the strongest final quality in every setting. Interestingly, Play2Code is considerably stronger on Phaser than on Godot. The category breakdown reveals why: Phaser initializations tend to have
weaker Mechanics and Depth but stronger Visuals and Art, yielding similar Overall scores while leaving more readily improvable functional headroom. Play2Code can therefore simply find these obvious deficiencies and recover substantial quality, narrowing its gap to \rsigame{}. Nevertheless, \rsigame{} still consistently produces the best final games, indicating that its advantage extends beyond low-level improvement even when such improvement already captures much of the available headroom.

\section{Further Analysis}
\label{sec:further}

\begin{figure}[!t]
\centering
\includegraphics[width=\textwidth]{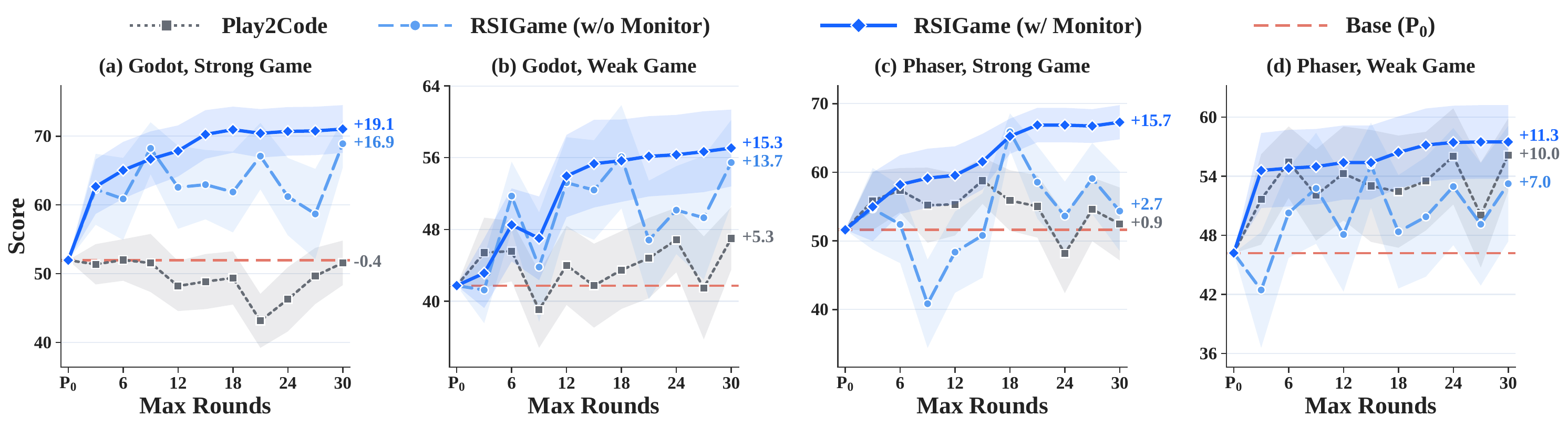}
\caption{
\textbf{Development-time scaling on GameCraft-Bench (40 tasks).}
Across Godot and Phaser with strong (GPT-5.5) and weak (Qwen3.8-27B) initial
generators, iterative development can plateau or regress, while \rsigame{}
achieves sustained improvement across development budgets. The Global Quality
Monitor preserves the best checkpoint reached so far, improving the
quality--cost tradeoff over returning the last checkpoint.
Shaded regions denote $\pm1$ standard error.
}
\label{fig:devscaling}
\end{figure}

\begin{figure}[t]
\centering
\includegraphics[width=\textwidth]{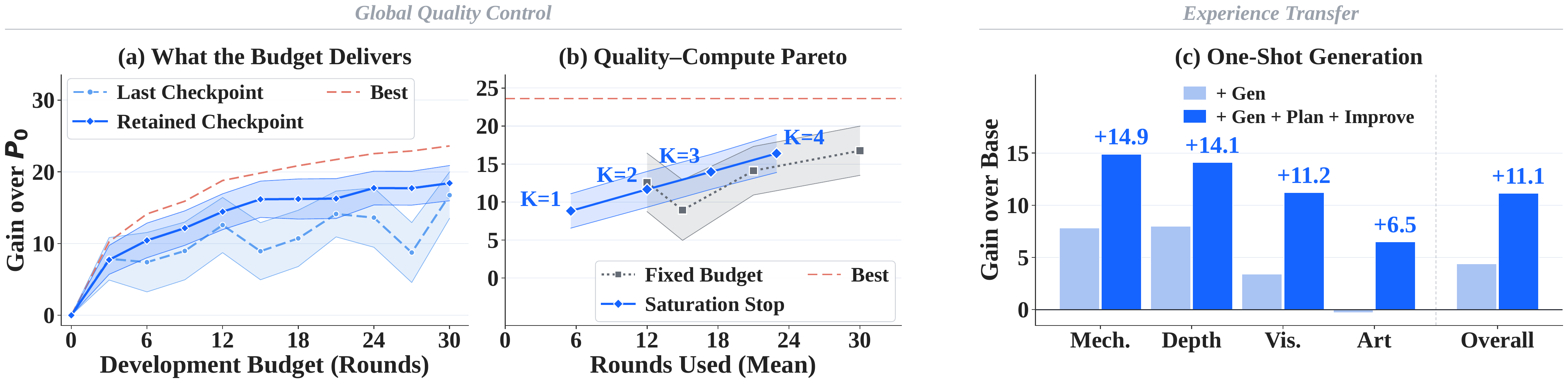}
\caption{
\textbf{Global quality control and experience transfer.}
(\textbf{a}) Best-checkpoint tracking protects earlier gains from later regressions.
(\textbf{b}) Saturation-aware stopping achieves comparable quality with fewer development rounds.
(\textbf{c}) Internalizing verified development experience improves one-shot generation across all quality dimensions.
}
\label{fig:global_analysis}
\end{figure}

\begin{figure}[t]
\centering
\includegraphics[width=\textwidth]{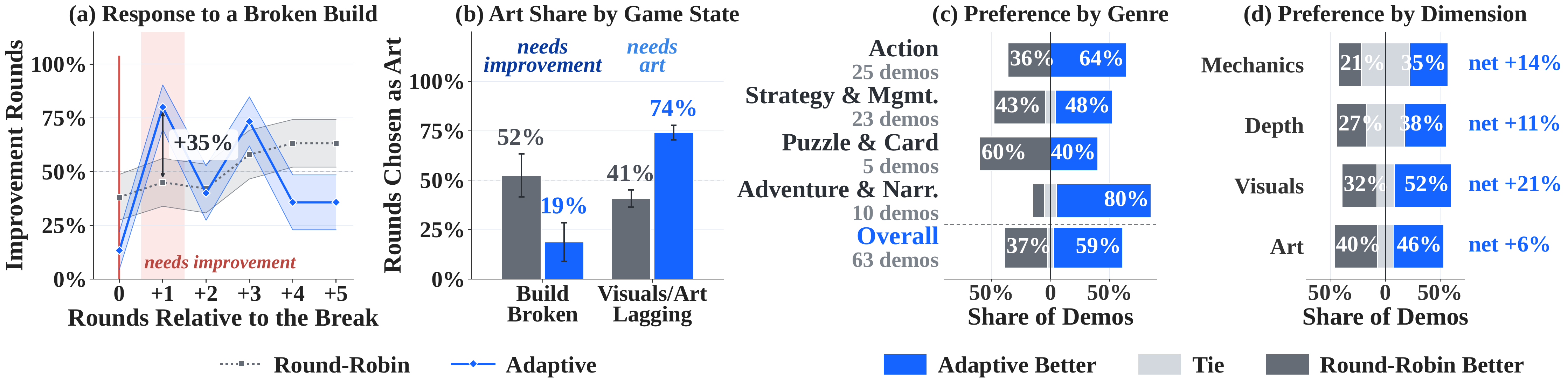}
\caption{
\textbf{Adaptive development follows the game state and improves final quality.}
(\textbf{a, b}) Adaptive development reallocates effort according to the current
bottleneck, shifting toward improvement after build failures and toward art
when visual quality lags.
(\textbf{c, d}) Blind pairwise evaluation consistently favors adaptive development
over round-robin scheduling.
}
\label{fig:direction}
\end{figure}

\paragraph{Can Global Quality Monitoring Improve Development-Time Scaling?}
\label{sec:quality_monitor}

Figure~\ref{fig:devscaling} reveals a clear difference in how development methods scale with additional compute. Simply extending the budget does not guarantee better games: Play2Code quickly plateaus and often regresses as more rounds are added. In contrast, \rsigame{} consistently converts additional development rounds into higher game quality across both Godot and Phaser and for both strong and weak initializations. The local loop alone already produces substantial gains, but its trajectory remains volatile; the Global Quality
Monitor stabilizes this process by retaining the best state reached so far, yielding sustained improvement throughout the development budget.

Figure~\ref{fig:global_analysis}(a)(b) further isolates the benefit of this global control. The retained checkpoint closely tracks the oracle best one within each budget, preventing later edits from erasing earlier gains, while saturation-aware stopping achieves comparable quality to much longer fixed-budget runs with fewer rounds. Together, these results show that the monitor makes development-time scaling both more reliable and more compute-efficient.
Detailed stopping criteria are provided in Appendix~\ref{app:quality_monitor}.

\paragraph{Does Adaptive Evolution Matter?}
Figure~\ref{fig:direction} examines whether the local loop benefits from choosing its own development focus adaptively rather than following a fixed round-robin schedule.
Adaptive Evolution puts its effort to the current game state: after a build failure it shifts strongly toward improvement, while once visual quality becomes the bottleneck it allocates substantially more rounds to art (Figure~\ref{fig:direction}(a) (b)).
This adaptive behavior also leads to better final games: blind pairwise evaluation prefers adaptive development on \(59\%\) of comparisons versus \(37\%\) for round-robin, with consistent advantages across all four quality dimensions (Figure~\ref{fig:direction}(c)(d)).
These results show that long-horizon development benefits from letting the local loop respond to the evolving needs of the game rather than following a fixed schedule.
Full evaluation details are provided in Appendix~\ref{app:direction_eval}.

\paragraph{Does Agentic Verification Make Local Improvement More Reliable?}
Table~\ref{tab:verification} evaluates verification both before and after editing. Evidence-grounded pre-improvement verification raises grounded precision from \(58.6\%\) to \(72.3\%\) and reduces unsupported targets from
\(1.93\) to \(0.50\) per round. After editing, replay-based verification detects \(76.2\%\) of unsuccessful improvements and reaches \(84.4\%\) balanced accuracy, whereas build-only checking detects none of these behavioral failures. Together, these results show that agentic verification improves both the quality of development decisions and the reliability of the feedback returned to subsequent rounds.
The audit protocol and metric definitions are given in Appendix~\ref{app:verification_eval}.

\begin{table}[t]
\centering
\small
\setlength{\tabcolsep}{4pt}
\begin{minipage}[t]{0.49\textwidth}
\centering
\begin{tabular}{@{}lcc@{}}
\toprule
Method
& \shortstack{Grounded\\Prec. $\uparrow$}
& \shortstack{Ungrounded\\/ Round $\downarrow$} \\
\midrule
Free critic & 58.6\% & 1.93 \\
Agentic verification & \textbf{72.3\%} & \textbf{0.50} \\
\bottomrule
\end{tabular}
\subcaption{Pre-improvement target grounding, 40 sessions.}
\label{tab:preverify}
\end{minipage}\hfill
\begin{minipage}[t]{0.49\textwidth}
\centering
\begin{tabular}{@{}lccc@{}}
\toprule
Method
& \shortstack{Failure\\Recall $\uparrow$}
& \shortstack{Spec.\\$\uparrow$}
& \shortstack{Bal.\\Acc. $\uparrow$} \\
\midrule
Build only & 0.0\% & \textbf{100.0\%} & 50.0\% \\
Replay $+$ verification & \textbf{76.2\%} & 92.6\% & \textbf{84.4\%} \\
\bottomrule
\end{tabular}
\subcaption{Post-improvement failure detection, 48 adjudicated rounds.}
\label{tab:postverify}
\end{minipage}
\caption{
\textbf{Agentic verification improves both target grounding and post-edit reliability.}
Before editing, it filters unsupported improvement targets; after editing, replay-based verification detects failed changes and regressions that compilation alone cannot reveal.
}
\label{tab:verification}
\end{table}

\paragraph{Can Development Experience Transfer to Future Generation?}
Figure~\ref{fig:global_analysis}(c) shows that development experience transfers back to the generation model. Training on generation traces alone yields uneven gains across quality dimensions, whereas incorporating planning and verified improvement experience produces consistent improvements in Mechanics, Depth, Visuals, and Art. The gains are particularly pronounced in Mechanics \(+14.9\) and Depth \(+14.1\), leading to a \(+11.1\)-point Overall improvement before any test-time development. This suggests that planning and improvement trajectories provide reusable knowledge beyond simply imitating successful game generations.

\section{Related Work}

\paragraph{Game Generation and Development Benchmarks.}
Recent benchmarks increasingly evaluate agents on complete, executable games rather than isolated code snippets. OpenGame-Bench~\citep{opengame} evaluates web-based games through build validity, visual usability, and instruction
alignment; GameCraft-Bench~\citep{gamecraftbench} extends evaluation to engine-based games with interaction-grounded measures of mechanics, content, visuals, and art; and GameDevBench~\citep{gamedevbench} studies multimodal game development over larger codebases and assets. Related work further examines interactions in game-playing agents and broader web- or code-based game generation, through both benchmarks and generation systems~\citep{omnigamearena,gamegenllm,vgamegym,webgamebench,creativegame}.
Together, these efforts shift evaluation from static code correctness toward the quality of executable, interactive game artifacts. However, these works score a single submitted build, leaving open how an agent should keep improving a game across rounds of development, which is the setting we study.

\paragraph{Iterative Agentic Game Development.}
Recent systems extend game generation into iterative development.
Play2Code~\citep{play2code} alternates between gameplay and code revision, using interaction feedback from each build to guide subsequent edits, while VibeGame~\citep{vibegame} coordinates multiple specialized agents for continued game development. OpenGame~\citep{opengame} further accumulates reusable development skills across projects to support subsequent generation. 
These efforts follow broader agentic coding paradigms based on repeated reasoning, execution, and feedback ~\citep{sweagent,metagpt,react,reflexion}.
Rather than introducing iteration itself, \rsigame{} focuses on how iterative development is sustained reliably over long horizons. It combines adaptive, evidence-grounded local improvement with global best-checkpoint tracking and saturation detection. High-level guidance can further reopen development after autonomous progress stalls, enabling multi-stage game evolution.

\paragraph{Verification and Learning from Development Experience.}
Agentic verification has been studied through general LLM-as-judge protocols~\citep{mtbench}, reliable process control method in agent loop~\citep{structagent}, and game-specific methods that validate behavior through runtime interaction~\citep{gamegenverifier,play2code}.
In \rsigame{}, independent verification closes each local development round: the updated game is replayed to confirm the intended improvement and detect new regressions, with verified outcomes written back into the shared development state.
Beyond individual tasks, prior work learns from feedback, memory, or verified trajectories~\citep{reflexion,selfrefine,selfdebug,cure,mstar,star,verifiercurriculum}, while game systems reuse experience across generations~\citep{opengame,vibegame}.
\rsigame{} extends this idea by jointly internalizing planning, generation, and verified improvement experience to strengthen future game generation and development.

\section{Conclusion}
We presented \rsigame{}, an autonomous agentic game development framework that formulates game creation as a recursive self-improvement process. \rsigame{} strengthens the development loop with broad exploration, structured diagnosis, prioritized improvement, and an evolving checklist that continuously expands the set of identified issues and improvement opportunities. Its local explore-diagnose-improve loop enables evidence-grounded revision of the executable game, while the global loop tracks long-horizon progress, preserves the best checkpoint, and detects saturation or regression. We further introduced training-based experience internalization to transfer successful development trajectories back into the backbone game generation model, improving future game creation beyond test-time refinement alone. 
Experiments across 140 GameCraft-Bench tasks, two game engines, and five generators demonstrated that \rsigame{} consistently improves game quality under matched development budgets, enabling smaller open models to reach or even surpass substantially stronger generators after iterative development. 


\clearpage

\subsection*{AI use statement}

Large language models are integral to the research methodology of this work. GPT-5.5 is used for game generation and planning-data collection; GLM-5.3-Flash is used for game improvement, verification, and improvement-data collection; and Qwen3.8-27B serves as the gameplay test agent, the benchmark evaluation judge, and the base model for experience internalization.
Claude Opus 5 is additionally used for the independent verification audit.

LLM-based assistants were also used during research and manuscript preparation for language editing, presentation refinement, and programming assistance.
The authors designed the experiments, reviewed the resulting artifacts and measurements, and verified \textbf{all} reported results and scientific claims.
The authors take full responsibility for the content of the paper.

\subsection*{Ethics statement}

Our study focuses on general-purpose game-generation and game-development
tasks. We apply safety constraints during generation and exclude artifacts
containing unsafe or sensitive content from evaluation and data curation.
The experiments do not use private user data, personal information, or
user-generated content.

Human involvement is limited to game evaluation and high-level development
guidance. We study the resulting game assessments and guidance rather than
participant attributes, and collect no sensitive personal information.
Released datasets and artifacts are screened to remove
machine-specific paths, credentials, and personal information.

\subsection*{Reproducibility Statement}

All reported results are traceable to retained experimental artifacts and documented protocols. Table~\ref{tab:repro} provides a single index to the released artifacts and to the appendix sections specifying each component of the experimental pipeline.

\textbf{Controlled comparisons.}
For each comparison, the initial project \(P_0\) is frozen and every development method receives an identical clone, the same improvement backbone, and the same per-round tool budget. Row-specific development budgets are reported explicitly in Appendix~\ref{app:config}. Each resulting game is independently replayed and scored three times, while benchmark rubrics, scores, and judge outputs remain hidden from all development agents. Replay and judge variability are quantified in Appendix~\ref{app:stability}.

\textbf{Artifact availability.}
The full implementation, prompts, and evaluation pipeline are available in our code repository\footnote{\url{https://github.com/WenyiWU0111/RSIGame}}, and representative evolved games are playable from the project page\footnote{\url{https://huggingface.co/spaces/RSIGame/rsigame-page}}. Frozen base projects, training trajectories, and per-round run trees are undergoing internal review and will be released upon approval. We also
release on Hugging Face\footnote{\url{https://huggingface.co/datasets/RSIGame/RSIGame-TableArtifacts}} the complete evaluation artifacts underlying the reported results, including 51{,}644 scoring files containing per-task scores, judge outputs, and replay reports. Their release status and corresponding protocol documentation are summarized in Table~\ref{tab:repro}.

\clearpage
\bibliography{iclr2027_conference}
\bibliographystyle{iclr2027_conference}

\clearpage
\appendix

\newcommand{\apxsec}[4]{%
  \noindent\hyperref[#1]{%
    \makebox[1.9em][l]{\textcolor{black!70}{\bfseries\ref{#1}}}%
    \textbf{#2}}%
  \nobreak\hfill\nobreak\hyperref[#1]{\textbf{\pageref{#1}}}\par
  \nobreak\vspace{0.5pt}%
  \noindent\hspace{1.9em}\begin{minipage}{\dimexpr\linewidth-1.9em\relax}
    \footnotesize\textcolor{black!45}{#3}\par
    #4
  \end{minipage}\par\vspace{3.5pt}}
\newcommand{\apxsub}[2]{%
  \nobreak\vspace{1pt}\noindent\hyperref[#1]{\small\ref{#1}\hspace{0.6em}#2}%
  \nobreak\hfill\nobreak\hyperref[#1]{\small\pageref{#1}}\par}

\begingroup
\hypersetup{pdfborder={0 0 0}, linkbordercolor=white}
\setlength{\parindent}{0pt}
{\Large\bfseries Appendix Overview}\par\vspace{3pt}
{\small\textcolor{black!60}{This appendix provides implementation details,
experimental protocols, training specifications, and extended analyses
supporting the main paper.}}\par\vspace{9pt}

\apxsec{app:repro}{Reproducibility Index}
  {Artifacts, release status, and where each protocol detail is specified.}{}

\apxsec{app:limit}{Limitations}
  {Scope of the autonomous pipeline, the dimensions evaluated, and project size.}{}

\apxsec{app:implementation}{\rsigame{} System Design and Implementation}
  {Complete specification of the recursive development system and its agent
   interfaces.}
  {\apxsub{app:algorithm}{Recursive Development Procedure}%
   \apxsub{app:quality_monitor}{Global Quality Monitoring}%
   \apxsub{app:director}{High-Level Director Interface}%
   \apxsub{app:prompts}{Agent Prompts and Output Contracts}}

\apxsec{app:protocol}{Experimental Protocols and Evaluation Reliability}
  {Configuration, accounting, and reliability protocols underlying the reported
   experiments.}
  {\apxsub{app:config}{Experimental Configuration}%
   \apxsub{app:cost}{Token and Cost Accounting}%
   \apxsub{app:stability}{Replay and Scoring Stability}%
   \apxsub{app:freeplay}{Free-Play Evaluation}%
   \apxsub{app:significance}{Statistical Reliability of Main Results}%
   \apxsub{app:direction_eval}{Adaptive Direction Policy Evaluation}%
   \apxsub{app:verification_eval}{Agentic Verification Audit}%
   \apxsub{app:stage_eval}{Multi-Stage Guidance Evaluation}}

\apxsec{app:training}{Training Data and Experience Internalization}
  {Construction, curation, and internalization of game-development experience.}
  {\apxsub{app:training_corpus}{Training Corpus Construction and Curation}%
   \apxsub{app:finetuning}{Fine-Tuning Configuration}%
   \apxsub{app:internalize_ablation}{Training-Data Ablation}}

\apxsec{app:extra}{Extended Evaluation and Case Studies}
  {Extended comparisons, qualitative trajectories, and task-family breakdowns.}
  {\apxsub{app:vibegame}{Comparison with a Multi-Agent Generate-and-Verify System}%
   \apxsub{app:cases}{Qualitative Evolution Case Studies}%
   \apxsub{app:families}{Per-Family Results}}
\endgroup
\clearpage

\section{Reproducibility Index}
\label{app:repro}
Table~\ref{tab:repro} is the single index referred to by the
Reproducibility Statement: for each artifact it gives what the artifact
contains, whether it is released, and where the corresponding protocol or
implementation detail is specified.

\begin{table}[h!]
\centering
\caption{
\textbf{Reproducibility index.}
Artifacts, release status, and locations of the corresponding implementation
and protocol details.
}
\label{tab:repro}
\small
\begin{tabular*}{\textwidth}{
@{\extracolsep{\fill}}
p{0.24\textwidth}
p{0.46\textwidth}
p{0.20\textwidth}}
\toprule
\textbf{Item} & \textbf{Contents / specification} & \textbf{Location} \\
\midrule

\multicolumn{3}{l}{\emph{Released artifacts}} \\
\midrule
Implementation
& Local/global loops, Global Quality Monitor, director-review tool, and scoring harness
& Code, live \\

Prompts and output contracts
& Complete prompts and structured outputs used by all agents
& Code, live; App.~\ref{app:prompts} \\

Playable evolution
& Six representative games with retained checkpoints
& Project page, live \\

Frozen base projects \(P_0\)
& Starting project for each task and generator
& On release \\

Evaluation artifacts
& Per-task rubric scores, judge outputs, and replay reports (51{,}644 files)
& HF, live$^\dagger$ \\

Training corpus
& 2{,}213 generation trajectories, 2{,}108 plans, and 2{,}013 verified improvement rounds
& On release \\

Run trees
& Per-round project snapshots for long-horizon development runs
& On release \\

\midrule
\multicolumn{3}{l}{\emph{Protocol and implementation details}} \\
\midrule

Recursive development loop
& One local round, global checkpoint update, and stopping logic
& App.~\ref{app:algorithm} \\

Global Quality Monitor
& Best-checkpoint selection, saturation criterion, and patience \(K\)
& App.~\ref{app:quality_monitor} \\

Model assignment
& Models used for generation, exploration, editing, verification, and judging
& Tab.~\ref{tab:cfg-models} \\

Hyperparameters
& Loop, monitor, and scoring parameters
& Tab.~\ref{tab:cfg-params} \\

Development budgets
& Rounds and tool-call budgets for each reported row
& App.~\ref{app:config} \\

Token and cost accounting
& Billable tokens, excluded costs, and pricing
& App.~\ref{app:cost} \\

Scoring stability
& Replay and judge variability
& App.~\ref{app:stability} \\

Corpus construction
& Extraction, verification filtering, and contamination checks
& App.~\ref{app:training_corpus} \\

Fine-tuning
& Training mixture and hyperparameters
& App.~\ref{app:finetuning} \\

Director review
& Director inputs, interaction budget, and guidance schema
& App.~\ref{app:director} \\

Per-family results
& Main-table results broken down by task family
& App.~\ref{app:families} \\

\bottomrule
\end{tabular*}

\vspace{1mm}
{\footnotesize
$^\dagger$ Released on Hugging Face:
\href{https://huggingface.co/datasets/RSIGame/RSIGame-TableArtifacts}{\texttt{RSIGame/RSIGame-TableArtifacts}}.
Machine-specific paths and identifying terms are removed before release, and
the exported files are re-parsed and re-scanned.
}
\end{table}

\section{Limitations}
\label{app:limit}
The core \rsigame{} pipeline is fully autonomous: adaptive local development, global quality monitoring, best-checkpoint tracking, and saturation-aware stopping require no human intervention. We additionally study an optional directed extension in which sparse high-level guidance from a human or a
stronger model is introduced only after autonomous improvement saturates.
Our current study evaluates this mechanism on a limited set of multi-stage development cases; a broader investigation of when guidance should be invoked, how much guidance is beneficial, and how human and model guidance differ remains future work.

Our evaluation is also centered on the four dimensions of Mechanics, Depth, Visuals, and Art. These dimensions do not fully capture higher-level properties such as originality, narrative quality, long-term player engagement, or subjective enjoyment.

Finally, our experiments focus on relatively compact games that can be iteratively developed within practical agent budgets. Extending autonomous recursive development to substantially larger projects with longer horizons, richer assets, and more complex cross-system dependencies remains an important direction.

\section{\rsigame{} System Design and Implementation}
\label{app:implementation}

\subsection{Recursive Development Procedure}
\label{app:algorithm}

Algorithm~\ref{alg:rsigame} is the loop of Section~\ref{sec:method} written out: one
round of the inner loop, the checkpoint at which the Global Quality Monitor runs, and
the two things saturation can lead to -- returning the champion, or taking a piece of
guidance and opening the next stage.

The multi-stage structure is the counter \(c\) and the stage index \(s\). A checkpoint
either promotes a new champion, which sets \(c\) back to zero, or finds nothing better,
which advances it; \(K\) checkpoints in a row without a promotion is what the paper calls
saturation. Saturation does not by itself end development: it ends the \emph{stage}. If
guidance is available, a brief is written on the champion, the next stage restarts from
that champion rather than from the last round's build, \(c\) is reset, and the loop
continues with \(\gamma_{s+1}\) steering every subsequent round. Development ends only
when a stage saturates and no further guidance is given. The runs in this paper use
\(E = 3\) and \(K = 3\).

\begin{algorithm}[H]
\caption{\rsigame{} Recursive Game Development}
\label{alg:rsigame}
\begin{algorithmic}[1]
\Require Game specification \(x\), initial project \(P_0\), initial checklist \(\mathcal{C}_0\), checkpoint interval \(E\), patience \(K\)
\State \(P_0^\star \gets P_0,\ \gamma_0 \gets \varnothing,\ t \gets 0,\ s \gets 0,\ c \gets 0\) \Comment{champion, guidance, round, stage, idle checkpoints}
\While{development budget remains}
    \State \(d_t \sim \pi_{\mathrm{controller}}(\cdot \mid x,P_t,\mathcal{C}_t,\gamma_s)\) \Comment{the checklist and guidance steer the round}
    \State \(\tau_t \sim \pi_{\mathrm{explorer}}(\cdot \mid P_t,d_t)\)
    \State \(\Delta_t \sim \pi_{\mathrm{editor}}(\cdot \mid P_t,d_t,\tau_t)\),\quad \(P_{t+1} \gets \operatorname{Apply}(P_t,\Delta_t)\)
    \State \(v_t \sim \pi_{\mathrm{verifier}}(\cdot \mid P_{t+1},d_t,\tau_t)\) \Comment{is the improvement observable, and what regressed}
    \State \(\mathcal{C}_{t+1} \gets U(\mathcal{C}_t,d_t,\tau_t,\Delta_t,v_t)\) \Comment{the working memory the next round builds on}
    \If{\(t \bmod E = 0\)} \Comment{a checkpoint: the Monitor compares, the loop does not}
        \State \(P' \gets \operatorname{SelectBest}(P_s^\star,P_{t+1})\)
        \If{\(P' \neq P_s^\star\)}
            \State \(P_s^\star \gets P'\);\ \(c \gets 0\) \Comment{a new champion; the stage is still paying}
        \Else
            \State \(c \gets c + 1\) \Comment{another checkpoint with nothing better}
        \EndIf
        \If{\(c \geq K\)} \Comment{saturated: \(K\) checkpoints without a new champion}
            \State \textbf{if} no guidance is available \textbf{then return} \(P_s^\star\)
            \State \(\gamma_{s+1} \gets \textsc{Guidance}(x,P_s^\star)\) \Comment{a brief on the champion opens the next stage}
            \State \(P_{s+1}^\star \gets P_s^\star,\ P_{t+1} \gets P_s^\star,\ s \gets s+1,\ c \gets 0\) \Comment{the next stage restarts here}
        \EndIf
    \EndIf
    \State \(t \gets t+1\)
\EndWhile
\State \Return \(P_s^\star\)
\end{algorithmic}
\end{algorithm}

\raggedbottom

\subsection{Global Quality Monitoring}
\label{app:quality_monitor}

\paragraph{What it compares.} The Monitor never scores a version on its own.
At each checkpoint it replays the retained champion $P_s^\star$ and the candidate
build on the \emph{same} set of demonstrations, pairs the two recordings
scenario by scenario, and asks only what changed between them. The comparator
is not told which round produced the candidate, what the improvement was trying to
achieve, or how much budget has been spent: direction and budget are decided
one layer up, and supplying them here would let development intent colour what
the model reports seeing.

\paragraph{Scale.} Each criterion of the proxy rubric is read on three
anchored levels --- $1.0$ works as described, $0.5$ works partly,
intermittently, or only in some cases, $0.0$ does not work --- for the champion
and the candidate separately, and only on criteria that \emph{both} recordings
show, so that a difference is a difference in the game rather than in what the
two replays happened to reach. Criterion readings are combined into a weighted
overall value in $[0, 1]$, reported as two halves: the criteria from the current
stage brief, and those from the task document.

\paragraph{Adoption.} The candidate replaces the champion when the brief half
improves by at least $0.05$ and the task half falls by no more than $0.05$;
otherwise the champion is kept. The task half acts as a floor rather than a
target, so a build may be adopted for advancing the stage's own goals provided
it does not regress what the task already required. Without stage criteria the
rule reduces to a $0.05$ gain on the overall value.

\paragraph{Checkpoints and stopping.} Checkpoints are taken every three rounds,
which is the shortest interval that contains an improvement, its verification and a
replay, and therefore the shortest window in which a difference can be
observed at all. Development stops once the champion has gone unchanged for
$K$ consecutive checkpoints; we use $K = 3$. Figure~\ref{fig:global_analysis} (b)
places $K = 1 \ldots 4$ against fixed budgets on the same runs: smaller $K$
stops sooner and delivers less, larger $K$ approaches the fixed-budget score at
a fraction of its rounds, and $K = 3$ sits at the knee.

\paragraph{Isolation from the evaluator.} The Monitor and the benchmark judge
share only the replay mechanism. The Monitor reads a separate proxy rubric,
never the benchmark's held-out rubric, and never its scores or feedback; the
module does not import the judge at all. The separation is also internal: the
comparator is the only component that sees pixels, and the stage that decides
direction and stopping reads its structured counts rather than the frames, so
neither layer can both interpret an image and act on its own interpretation.

\subsection{High-Level Director Interface}
\label{app:director}

Section~\ref{sec:method} describes what a stage brief does to the loop. This
appendix describes where briefs come from. Every brief reported in this paper
--- human and model alike --- was written in the interface below, under one
protocol, so that ``a person directed this run'' and ``a model directed this
run'' differ in the director and in nothing else.

\subsubsection{What the Director Is Given}
\label{app:dir_given}

\begin{figure}[t]
  \centering
  \includegraphics[width=\linewidth]{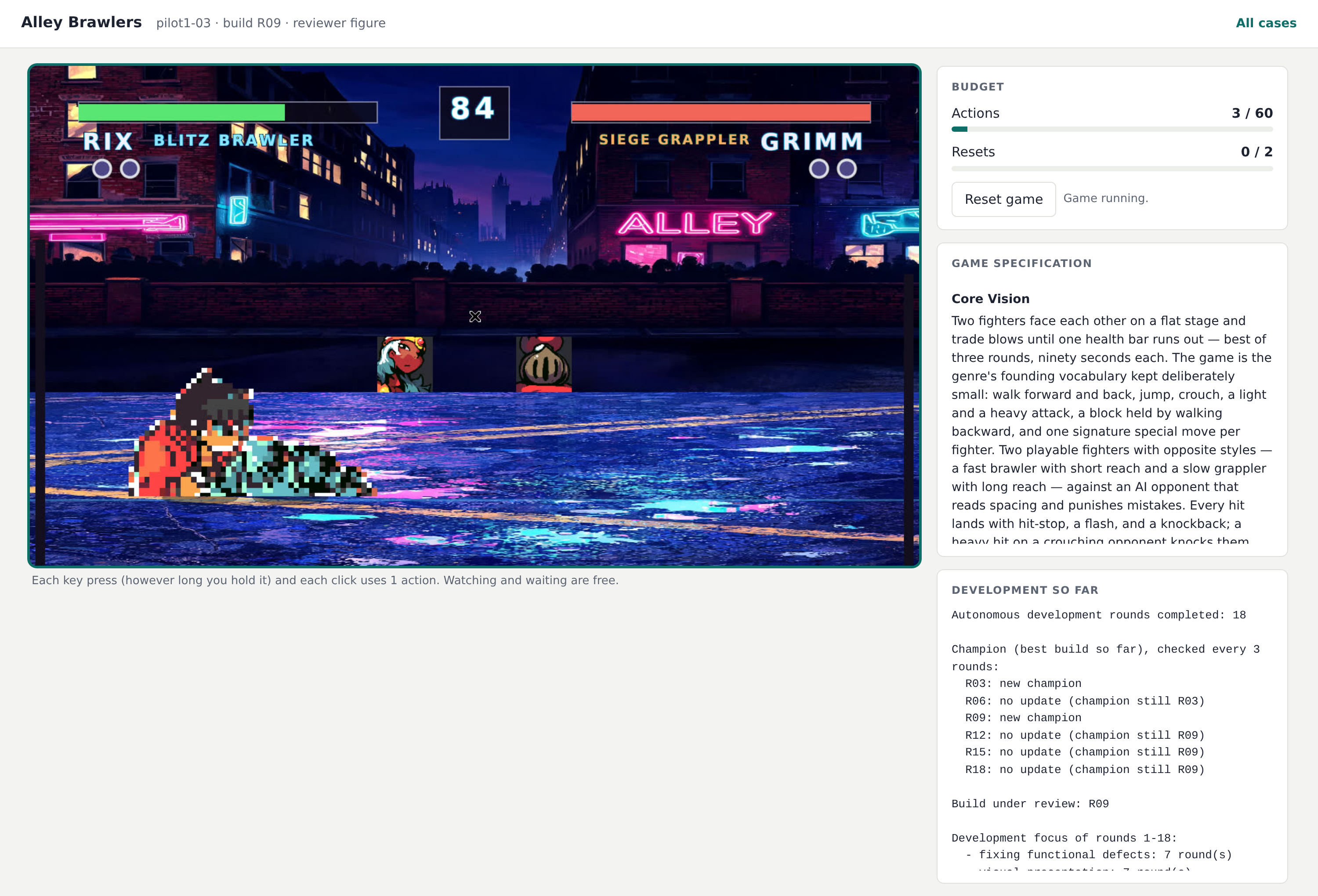}
  \caption{The director review page, mid-session, on the \emph{Alley Brawlers}
  build that had survived six monitor checkpoints: the live game, the budget,
  the task specification, and the development summary --- champions at R03 and
  R09, no update since, and the split of the eighteen rounds between functional
  and presentational work. The frame is a real moment of play and shows the
  defect both directors went on to describe in different words: the player's
  fighter has come apart into a band of scrambled sprite tiles. The brief form
  below it is Figure~\ref{fig:review_form}.}
  \label{fig:review_page}
\end{figure}

A director reviews one build of one game. Figure~\ref{fig:review_page} is the
page: the game runs live in the left panel and takes the keyboard when the
frame is focused; the right column holds the budget, the task specification the
game was generated from, and a summary of the development that has already
happened; the form below is the brief.

The summary is deliberately thin. It reports how many autonomous rounds have
run, which of them became champions at the Global Quality Monitor's
three-round checkpoints, which build is under review, and how those rounds
divided between functional and presentational work. It reports no score, at
any round, for any build.

That omission is the point. The director is asked what the game needs next, and
a score would answer a different question --- whether the last round worked ---
in a currency the loop already optimises. Nothing else about the run reaches the
page either: no source, no logs, no improvement transcripts, no other director's
review. The instrument shows a playable game and the document it was supposed
to become.

\subsubsection{One Instrument, One Budget}
\label{app:dir_budget}

A director gets 60 actions and 2 resets. A key press costs one action however
long it is held, a click costs one, and watching and waiting are free, so the
budget buys interaction rather than patience. The counters in
Figure~\ref{fig:review_page} are the ones the session enforces, and the count
each director spent is stored with the brief.

A person plays through the browser. A model plays through a command-line client
--- \texttt{open}, \texttt{press}, \texttt{hold}, \texttt{click}, \texttt{wait},
\texttt{look}, \texttt{reset}, \texttt{state}, \texttt{submit} --- that drives
the same session object on the same server, spends the same budget through the
same counters, and returns the frames captured after each action for the model
to read. Neither director can see the other's transport; both are talking to one
running game.

The two directors reviewed the same bytes. Each brief records the hash of the
project tree it was written against, and for \emph{Alley Brawlers} both read
\texttt{c12fe90b30b6c176}. The human director spent the full 60 actions and no
resets; the model spent 25 and no resets.

\subsubsection{How the Model Directs}
\label{app:dir_directs}

The model director is given one frozen prompt, versioned
(\texttt{director-prompt/2}) and unchanged within a run; editing it makes a new
run. It casts the model as a senior game director reviewing a build whose
autonomous development has plateaued, tells it to decide what the next
\emph{stage} should be rather than to enumerate defects, and forbids
implementation-level instructions.

Two rules do the methodological work. Everything the model knows about the game
must come from the session: reading, listing or searching any other file --- any
source, any log, any score, any other review --- discards the review. And the
brief is submitted through the same endpoint the form posts to, in the same
schema, once; submission ends the session.

\subsubsection{What a Brief Is}
\label{app:dir_brief}

Four fields, identical for both directors: a one-sentence
\texttt{stage\_objective}; \texttt{why\_now}, grounded in what the director saw
while playing; one to three \texttt{priorities}, stated as directions rather
than code changes; and zero to three items to \texttt{preserve}. The loop
consumes the brief in three places --- the planner, stage ranking, and the
improvement and art prompts --- and the stage checklist that the Monitor scores
against is rebuilt so that the brief's own criteria carry the most weight
(Appendix~\ref{app:quality_monitor}).

\begin{figure}[t]
  \centering
  \includegraphics[width=\linewidth]{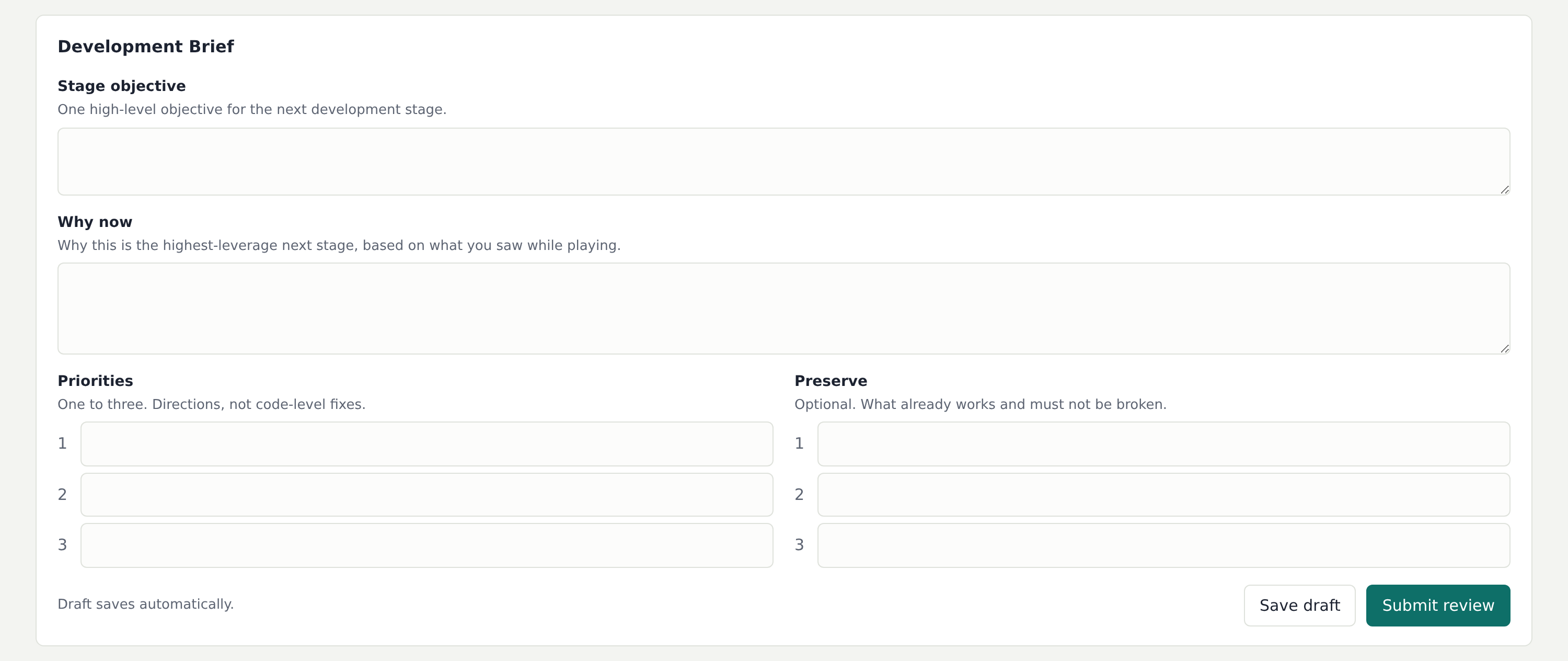}
  \caption{The brief form, the lower half of the page in
  Figure~\ref{fig:review_page}. The model director submits the same four fields
  as JSON through the endpoint this form posts to.}
  \label{fig:review_form}
\end{figure}

Table~\ref{tab:briefs} puts both briefs for \emph{Alley Brawlers} side by side.
They were written four minutes apart against the same tree, and they do not
overlap. The human read the build as a presentation problem and asked for art
and onboarding; the model played two full matches, never took a single point of
health off the opponent, and read the build as a mechanics problem. Neither is
the correct answer, and the outcome was close: from this checkpoint the
human-directed branch reached 62.0, the model-directed branch 61.5, and a
control branch restarted from the same build with no brief reached 60.0. What
the comparison shows is that a brief supplies a direction the loop's own
evidence had not produced in eighteen rounds, and that two directors looking at
one build through one instrument can supply entirely different ones.

\begin{table}[H]
  \centering
  \small
  \renewcommand{\arraystretch}{1.2}
  \begin{tabularx}{\linewidth}{@{}p{0.1\linewidth}X@{}}
    \toprule
    \multicolumn{2}{@{}l}{\textbf{Human director} --- 60/60 actions, 0 resets.} \\
    \midrule
    Objective & The art needs work: the title and game-over screens are cluttered
      and have no proper background, the character art is ugly and should be
      regenerated in a comic style, and attacks have no effects. The game also
      needs to teach how it is played --- right now I do not know what to press. \\
    Why now & The characters move and there are title and game-over screens, but
      it is still too rough to be appealing. \\
    \midrule
    \multicolumn{2}{@{}l}{\textbf{Model director} (Claude Opus 5) --- 25/60
      actions, 0 resets.} \\
    \midrule
    Objective & Turn the match from a scripted beatdown into a real two-sided
      fight: player inputs must visibly and reliably land, block and escape,
      against an AI that can be beaten, with two fighters that genuinely look
      and play differently. \\
    Why now & Across two full matches (one as each fighter) the AI never lost a
      pixel of health. Every round ended in about 18 seconds with me pinned in
      the left corner in a hit, knockdown, hit loop. Light, heavy, the special
      input, holding back, crouching and jumping all failed to trade, block or
      get me out. Nobody can learn to play when nothing they press matters.
      [\ldots] Seven rounds of visual polish have gone into the frame around the
      fight (title, background, banners, results screen), but the fight itself
      is still missing. \\
    \bottomrule
  \end{tabularx}
  \caption{The two briefs written for the same \emph{Alley Brawlers} build
  (tree \texttt{c12fe90b30b6c176}), abridged to their first two fields.
  Priorities and preserve items are omitted for space.}
  \label{tab:briefs}
\end{table}

\subsubsection{The Case Set}
\label{app:dir_cases}

\begin{figure}[t]
  \centering
  \includegraphics[width=0.78\linewidth]{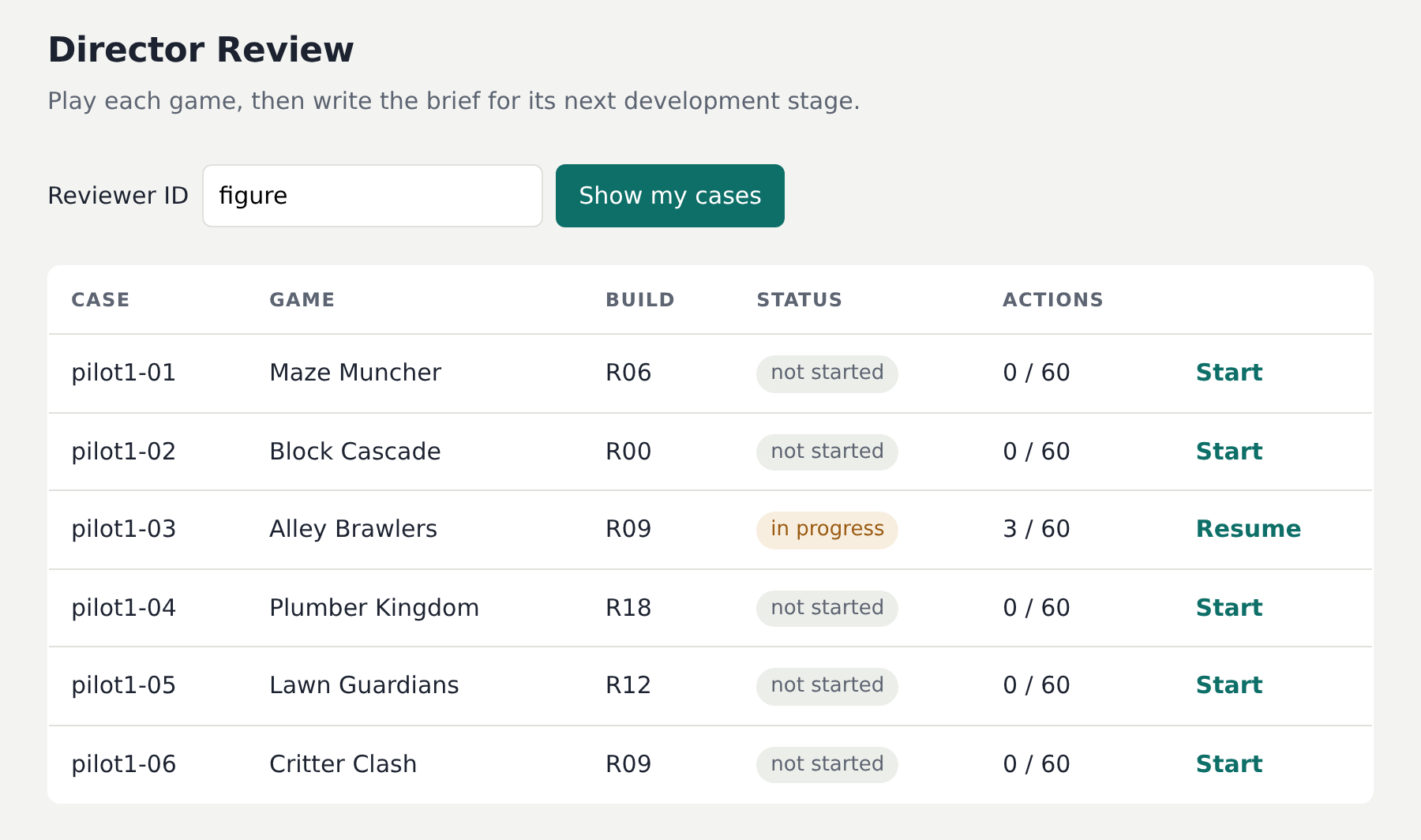}
  \caption{The case list a director opens: six games, the build under review for
  each, and the budget remaining. Case identifiers are internal labels.}
  \label{fig:review_cases}
\end{figure}

Six games were reviewed, each at the build its own run had converged on, which
ranges from the generated project to eighteen rounds in
(Figure~\ref{fig:review_cases}). Every game was reviewed independently by the
human director and by the model director, and each brief was then given to a
branch restarted from that same build, against a control branch given no brief.
Three of the six are the case studies of Appendix~\ref{app:cases}.

\subsection{Agent Prompts and Output Contracts}
\label{app:prompts}

This appendix reports the prompts of the five modules that carry the loop's
decisions, one per stage of Section~\ref{sec:method}. Each is abridged to its
decision skeleton: fixed catalogues, worked examples, frame-labelling
conventions and harness housekeeping are cut, and every cut is marked
\texttt{[...]}. Braces such as \texttt{\{items\}} are runtime slots filled
with the development state, the playtest record, or the replayed frames.

One design choice recurs across all five and is easier to see stated once. Each
prompt is told what it needs to decide its own question and deliberately not
told the rest. The planner is given the round's direction as a conclusion, not
as an argument. The Verifier is told what the round was trying to do,
because that is the claim it must check, while the comparator is told neither
that anything was improved nor which build is newer. The Editor is told
that the problem is already established, so that it starts at ``where in the
code does this live'' rather than re-deriving the diagnosis. Withholding is
what keeps a stage from confirming its own expectation.

\begin{table}[h]
  \centering\scriptsize
  \caption{The five LLM-facing modules of the \rsigame{} loop, by the stage of
  Section~\ref{sec:method} they implement.}
  \label{tab:prompt_contracts}
  \begin{tabular}{p{0.15\linewidth} p{0.14\linewidth} p{0.29\linewidth} p{0.28\linewidth}}
    \toprule
    Stage & Module & Output contract & Consumed by \\
    \midrule
    Direction Decision & Controller &
    One \texttt{development\_question}; \texttt{action} in \{\emph{reuse},
    \emph{explore}, \emph{escalate}\}; an \texttt{exploration\_plan} naming the
    cheapest mode that can answer it. &
    Sets the round's objective; \emph{escalate} reports the direction
    exhausted rather than inventing work. \\
    Agentic Exploration & Explorer &
    The 1--3 requirements a session should settle, the scenario to boot, and
    one actionable instruction for the play agent. &
    Drives the test agent's interaction with $P_t$, producing the trace
    $\tau_t$. \\
    Evidence-Grounded Editing & Editor &
    Edits to the project under a build constraint, plus a report of what was
    changed. &
    Produces $P_{t+1}$. \\
    Agentic Verification & Verifier &
    \texttt{goal\_achieved} in \{yes, no, unclear\}; \texttt{broken\_now};
    \texttt{next\_goal} when that list is non-empty. &
    Accepts or rejects the round; anything broken re-enters the state as
    pending work. \\
    Global Quality Monitoring & Global Quality Monitor &
    Per-criterion readings for both builds, changes with direction and
    magnitude, and regressions --- never a score. &
    Aggregated into the proxy value that decides the champion
    (Appendix~\ref{app:quality_monitor}). \\
    \bottomrule
  \end{tabular}
\end{table}

\subsubsection{Controller}
\label{app:prompt_direction}

The planner decides what to \emph{find out}, never what to change. It is given
the round's direction as a settled conclusion, the requirement state, what the
build did when last played, the demos available and the budget left; it returns
one question and the cheapest way to answer it. Most of the prompt's length
goes to ruling out questions that no observation can settle.

\begin{rsipromptbox}{PromptPlan}{Prompt P1. Controller}
You plan what to investigate next in a game that is being developed one round
at a time.

You do not write code, propose patches, or say what should be changed. Another
agent does that, after the evidence you ask for has been collected. Your output
is a decision about what to FIND OUT.

[... direction, requirement state, last playtest, demos, budget ...]

====================
WHAT TO DECIDE
====================
One primary question. It must be:
  - inside the direction above;
  - grounded in the state and evidence above, not in what a game like this
    usually has;
  - answerable by watching the game: someone replays it and looks. Not by
    reading the source, which nobody will do for you;
  - concrete enough that you can say what that observation would be;
  - useful for deciding whether something needs repairing.

Not a question: "Can the game be improved?" -- nothing observable answers it.
Not a question: "Improve the HUD layout." -- that is an instruction to change
something, and nothing has been observed yet.

Then choose one action:
  reuse      the evidence above already settles what the problem is.
  explore    the evidence is missing or ambiguous, and a specific observation
             would settle it.
  escalate   there is no grounded unresolved question left in this direction.
             Say so rather than inventing one.

If exploring, choose the cheapest mode that can answer the question:
  replay_existing_demo    a demo already exercises what you need to see
  extend_existing_demo    a demo reaches the state but stops before the answer
  custom_targeted_probe   no demo can reach it. The expensive option.

If a question has already been asked several rounds running without being
settled, asking it again the same way is not a plan. Either say what would be
different this time, or pick something else.

Reply with strict JSON and nothing else:
{"action": "reuse|explore|escalate",
 "development_question": {"question": "...", "why_now": "..."},
 "exploration_plan": {"mode": "...", "demo_id": "...", "focus": "...",
                      "stop_when": "..."}}
\end{rsipromptbox}

\subsubsection{Explorer}
\label{app:prompt_explore}

Exploration is aimed, not generic. The probe selector turns the requirements
the stage cannot yet answer into one instruction a play agent can carry out,
and is asked to weigh importance against what a short session can reach.

\begin{rsipromptbox}{PromptProbe}{Prompt P2. Explorer}
You are deciding what ONE play session should go and find out about this game
next.

These are the requirements this stage still cannot answer, with where each
could be checked:

[... candidate requirements, with status and scenario ...]

Pick the 1 to {cap} that matter most RIGHT NOW and write a single instruction
for the person who will play the build. Judge importance -- a requirement the
rest of the game depends on beats a detail, and something you can settle in a
few actions beats something that needs a long grind.

If the ones you pick share a place to look, say so; a session that has to boot
two different states spends half its steps travelling.

The instruction has to be ACTIONABLE: name the input to make or the screen to
reach and what to watch. "Check whether the plane rotates" is weak. "In a
sortie, press Left and then Right for about a second each and report whether
the plane's heading changes" can be carried out.

Reply with JSON and nothing else:
{"items": ["T04"], "scenario": "", "ask": "...", "why": "one sentence"}
\end{rsipromptbox}

\subsubsection{Editor}
\label{app:prompt_edit}

The Editor receives a problem that has already been established, and the
prompt's first job is to stop it from establishing it again. It is given the
source, the build command that must pass, and a way to run and watch the build;
it is not given the authority to change the harness or the evaluation traces.

\begin{rsipromptbox}{PromptRepair}{Prompt P3. Editor}
You are improving a small game. This is round {r} of at most {R}.

You did not play this build. Someone else did: a play agent drove it, the
frames were read, and what follows is what that evidence supports and what was
chosen for this round.

So the problem below has already been established. You do not have to find out
whether it is real, and you should not replay the game to confirm it -- that
work is done, and repeating it spends the calls you need for the repair. Your
job starts at "where in the code does this live".

[... the round's evidence packet, recent history, project map ...]

WHAT YOU HAVE
  - the source, in the directory you are working in
  - `{build_cmd}` -- it must pass when you stop
  - you can run the build and watch it: [...]

RULES
  - Change the game, not the harness. Do not edit `node_modules`, `dist`,
    `demo_outputs/`, or anything under `_repair_evidence/`.
  - Do not edit `demo_outputs/`. Those are the input traces the evaluator
    replays; changing them changes the exam, not the game.
  [... process and server hygiene ...]
\end{rsipromptbox}

\subsubsection{Verifier}
\label{app:prompt_verify}

After the edit, every shipped demo is replayed and one reader decides whether
the round earned its commit. The prompt separates two failures that are easy to
conflate: a feature the game never had is not a defect, and an input that
visibly does \emph{something} has been received even if the script never
reaches the outcome it was written for. Only an input that changes nothing
counts as broken.

\begin{rsipromptbox}{PromptVerify}{Prompt P4. Verifier}
A repair agent has just edited a game. Your job is to look at what the game
does now and answer two questions. You are the only check on this edit: after
you, it ships.

WHAT THE ROUND WAS TRYING TO DO
{goal}

[... the requirements this goal came from; the replayed frames ...]

ANSWER THESE TWO, IN THIS ORDER.

1. goal_achieved -- yes / no / unclear.
   Is the thing the round set out to do visibly true now? "unclear" is a real
   answer: say it when the demos never reach the part of the game the goal is
   about, rather than guessing.

2. broken_now -- a list, possibly empty.
   Anything the game does that is plainly wrong to a player watching these
   frames. Be concrete about what you SAW and in which frame. The kinds that
   matter most, because they are invisible in source code:
     - an input produces no change at all -- the before and after frames are
       the same, again and again, so the game is not listening
     - something covers the playfield: one sprite, one effect, one panel that
       hides the characters, the HUD or the action
     - the screen is frozen: the last frame equals the first
     - text is unreadable, clipped, or drawn on top of itself
   Do not list a missing feature here. A feature the game never had is a task
   requirement, not something that is broken.

AND ONE THING NOT TO REPORT. An input that visibly DOES something -- a mode
label appears, a prompt is shown -- has been received, even when the demo never
goes on to reach the outcome it was written for. That is the game asking for
more steps than this fixed script performs, and it is not a fault in the game.
Only an input that changes NOTHING belongs in broken_now.

Then, only if broken_now is not empty, write next_goal: one sentence telling
the repair agent what to fix first. Fix the worst thing, not all of them.
\end{rsipromptbox}

\subsubsection{Global Quality Monitor}
\label{app:prompt_monitor}

The comparator is the only component in the method that reads pixels. It is
shown the champion and the candidate replayed under the same fixed script and
asked what changed; it is told neither which build is newer nor that an improvement
took place. Most of the prompt defends against the two ways a pairwise reading
goes wrong: assuming the newer build is better, and mistaking recorder timing
for a behavioural change.

\begin{rsipromptbox}{PromptCompare}{Prompt P5. Global Quality Monitor}
Two builds of the same game were replayed with the SAME fixed input script.
Your job is to say what changed in this scenario, and nothing else.

[... the scenario, the criteria it can speak to, the paired frames ...]

The script is fixed, so both builds received the same inputs at the same
moments. For each moment below you get the OLD build just before and just after
the input, then the NEW build at the same two moments.

The recording is not frame-exact: replaying one build twice cuts the same
moment up to a frame or two apart, which by itself makes a score read one point
different, an animation look half a beat behind, or a falling piece sit one row
lower. That is the recorder, not the build.

You get two frames after each input for exactly this reason. A difference that
is gone by the second frame, or that later moments undo, is timing. A
difference that is still there at the next moment and at the FINAL frame is the
build. Say `same` for the first kind.

RULES
- Compare only what these frames support.
- Do not assume the newer build is better, and do not assume a difference you
  cannot explain is a defect. Both tilts are errors.
- If a behaviour is not exercised here, its criterion is `unobserved`. Not
  seeing something is not seeing it missing.
- Report anything that got worse separately, however small the rest.
- Judge relative change. Do not score either build.
- Do not propose code edits, repairs, or next steps. That is not your job.
\end{rsipromptbox}

\section{Experimental Protocols and Evaluation Reliability}
\label{app:protocol}

\subsection{Experimental Configuration}
\label{app:config}
\label{app:cost_and_protocol}

This appendix lists, in one place, which model plays which role, how each is
served and decoded, and every budget a run is given. Table~\ref{tab:cfg-models}
covers the models, and Table~\ref{tab:cfg-params} the loop and scoring parameters.

\paragraph{Scoring protocol.} Every number in Table~\ref{tab:godot_main} is the mean over the 140 tasks of a score produced by
the Qwen3.8-27B judge from three independent replay-and-score runs of the same
artefact. A task whose generator produced no project scores 0; so does a task
whose development has not finished. Base rows score the frozen $P_0$,
Play2Code rows the version thirty rounds deliver, and \rsigame{} rows the version the Global Quality Monitor is holding when the saturation stop fires.

\begin{table}[h]
\centering
\caption{Models by role. The generator differs per row group; every other role
is the same model in every row, so that a comparison between rows is a
comparison of development methods and not of the models inside them. Decoding
is greedy wherever the answer is a claim about the game: a coordinate, a
verdict, a rubric item. \textbf{Served by} matters for cost and for
reproducibility --- one model on OpenRouter is sold by a dozen providers, so
the judge and the monitor pin theirs.}
\label{tab:cfg-models}
\small
\begin{tabular*}{\textwidth}{@{\extracolsep{\fill}}llll}
\toprule
Role & Model & Served by & Decoding \\
\midrule
Generator & one per row group of & harness & harness \\
          & Table~\ref{tab:godot_main} & default & default \\
\midrule
Controller    & GLM-5.3-Flash & OpenRouter & $T=0$, top-$p$ 1 \\
Explorer    & Qwen3.8-27B & local vLLM (TP $=$ 2) & $T=0$, top-$p$ 1 \\
Verifier                & GLM-5.3-Flash & OpenRouter & $T=0$, top-$p$ 1 \\
Editor            & GLM-5.3-Flash & OpenRouter & $T=0$, top-$p$ 1 \\
Global Quality Monitor         & Qwen3.8-Flash & OpenRouter & $T=0$ \\
Asset generation & Seedream-5.0-lite & OpenRouter (images) & --- \\
\midrule
Evaluation judge        & Qwen3.8-27B & local vLLM & $T=0$, thinking off, \\
                        &             &  & $\leq$2048 output \\
\bottomrule
\end{tabular*}
\vspace{2pt}
\begin{minipage}{\textwidth}
\end{minipage}
\end{table}

\begin{table}[h]
\centering
\caption{Loop, monitor and scoring parameters. One value per knob, the same in
every run reported here.}
\label{tab:cfg-params}
\small
\begin{tabular*}{\textwidth}{@{\extracolsep{\fill}}llr}
\toprule
 & Parameter & Value \\
\midrule
\multirow{5}{*}{\emph{Round}}
 & Tool calls the Editor may make per round        & 26 \\
 & Wall-clock budget per improvement round                    & 1800 s \\
 & Improvement attempts per round before the round is given up & 3 \\
 & Verification samples per claimed fix                  & 3 \\
 & Probe steps taken before an edit is proposed          & 8 \\
\midrule
\multirow{3}{*}{\emph{Run}}
 & Generated assets per run                              & 20 \\
 & Project snapshot kept                                 & every round \\
 & Replay recording kept                                 & every 3rd round \\
\midrule
\multirow{5}{*}{\emph{Monitor}}
 & Checkpoint interval                                   & every 3 rounds \\
 & Demos replayed per checkpoint                         & $\leq$3 \\
 & Frames per demo the comparison reads                  & 16 \\
 & Margin a candidate must beat the champion by          & 0.05 \\
 & Checkpoints with an unchanged champion before the saturation stop ($K$) & 3 \\
\midrule
\multirow{3}{*}{\emph{Scoring}}
 & Replay-and-score runs per artefact                    & 3 \\
 & Uniform frames sampled per demo                       & 16 \\
 & Rubric items per task ($M$/$D$/$V$/$A$)               & 12--21 (median 18) \\
\bottomrule
\end{tabular*}
\end{table}


\subsection{Token and Cost Accounting}
\label{app:cost}

\paragraph{Scope.} We count two phases per task. \emph{Generation} is the
generator's single agent run that produces the frozen base $P_0$.
\emph{Development} is every model call made during the development rounds: the controller that determines the direction, the
explorer that plays the game, the verifier that inspects it, the editor that edits it, and the Global monitor's periodical check. Scoring by the evaluation judge is excluded for all
methods.

\paragraph{Tokens.} The \textbf{Tok.} column counts billable tokens: input
excluding cache reads, plus output. Cache reads are left out of this count
because long agent sessions re-read the same prefix on every call; they are
still paid, at the cache price, in \textbf{Cost}.

\paragraph{Cost.} Per call,
\begin{equation*}
\text{cost} = n_{\text{in}}\,p_{\text{in}} + n_{\text{cache}}\,p_{\text{cache}} + n_{\text{out}}\,p_{\text{out}},
\end{equation*}
with $n_{\text{in}}$ the uncached input tokens and OpenRouter list prices
(Table~\ref{tab:prices}). Development is not priced this way but summed from
the amount the API returns with each call, which the proxy records for every call the Controller, Explorer, Verifier, Editor, and Monitor make; the formula above is used
where only token counts survive, which is generation.
\textbf{Cost} covers development only; generation is reported separately in
Table~\ref{tab:gen-cost}. Generation cost is dominated by the harness rather
than the model: the same GPT-5.5 hits 94\% cache under Codex and 42\% under
OpenGame, and the resulting bills differ by a factor of eight.

\paragraph{Which price.} A model on OpenRouter is served by a dozen providers
whose prices differ by up to a factor of two, and the model page quotes only
one of them, which for several models is the dearest on the board. We therefore
price each model at its cheapest \emph{standard} endpoint (Table~\ref{tab:prices}).
Two exclusions make that well-defined. Latency tiers are not alternative
sellers of the same service and are left out: the flex tier halves GPT-5.5 by
deferring the request. And endpoints are compared on what these runs would
actually cost there rather than on input price alone, because a quoted cache
price of zero means the endpoint offers no prompt caching, not that cache reads
are free -- read literally it would make a provider without caching the
cheapest seller of a workload that is 42--96\% cache.

\paragraph{Sources.} Generation usage comes from each generation run's own
record (input, cache-read and output tokens), released with the base-game
corpora; for Codex $+$ GPT-5.5 on Godot only corpus-level means are recorded.
Development usage is logged per call by a proxy between the agents and the API.

\begin{table}[h]
\centering
\caption{OpenRouter list prices (\$ per million tokens), retrieved 2026-09-16.
Each model is priced at its cheapest standard endpoint for the cache mix of the
runs reported here; the serving provider is named because prices for one model
differ by up to a factor of two across providers.}
\label{tab:prices}
\small
\begin{tabular}{llrrr}
\toprule
Model & Provider & Input & Cache read & Output \\
\midrule
GPT-5.5          & OpenAI    & 5.000 & 0.500 & 30.00 \\
Kimi-K2.6        & Baidu     & 0.408 & 0.069 & \phantom{0}1.72 \\
Qwen3.8-27B      & DeepInfra & 0.150 & 0.037 & \phantom{0}1.88 \\
GLM-5.3-Flash    & DeepInfra & 0.075 & 0.015 & \phantom{0}0.25 \\
\bottomrule
\end{tabular}
\end{table}

\begin{table}[h]
\centering
\caption{Generation of the frozen base $P_0$, mean per task. \textbf{Billable}
is the \textbf{Tok.} of the Base rows; \textbf{Total} adds cache reads;
\textbf{Cost} applies Table~\ref{tab:prices} to all three token counts. The two
GPT-5.5 rows are the same model under two harnesses and differ in
price: Codex re-reads a cached prefix, OpenGame does not.
Over all 140 tasks.}
\label{tab:gen-cost}
\small
\begin{tabular*}{\textwidth}{@{\extracolsep{\fill}}llrrrrr}
\toprule
Engine & Generator & Billable & Total & Cache & Output & Cost \\
\midrule
Godot  & Codex $+$ GPT-5.5 (high) & 0.26M & \phantom{0}3.83M & 94\% & \phantom{0}28k & \$\phantom{0}3.77 \\
Godot  & GLM-5.3-Flash            & 0.44M & 11.43M & 96\% & \phantom{0}36k & \$\phantom{0}0.20 \\
Godot  & Kimi-K2.6                & 3.22M & \phantom{0}8.98M & 65\% & 123k & \$\phantom{0}1.87 \\
Godot  & Qwen3.8-27B             & 6.41M & 11.62M & 46\% & 179k & \$\phantom{0}1.47 \\
Godot  & Qwen3.8-27B (SFT)        & 0.57M & \phantom{0}8.21M & 95\% & 140k & -- \\
\midrule
Phaser & OpenGame $+$ GPT-5.5     & 5.36M & \phantom{0}9.21M & 42\% & \phantom{0}97k & \$31.17 \\
Phaser & Qwen3.8-27B & 9.03M & 26.69M & 67\% & 292k & \$\phantom{0}2.51 \\
\bottomrule
\end{tabular*}
\end{table}


\subsection{Replay and Scoring Stability}
\label{app:stability}

\paragraph{Replay Variability.}
GameCraft-Bench scores a game from its observed gameplay rather than its source code alone.
Even for the same frozen project, repeated playbacks can produce slightly different trajectories because game execution, input scheduling, and frame capture are not perfectly synchronized.
The magnitude of this variation depends on the game, and Table~\ref{tab:stability-family} measures which kind of game it depends on.
Therefore, repeatedly judging the same recording does not capture the full uncertainty of the evaluation process.

\paragraph{Evaluation Protocol.}
To account for this variability, every artifact is evaluated with three independent end-to-end runs.
Each run starts from a fresh replay, generates its own gameplay recording, and is scored independently by the same Qwen3.8-27B judge used for every score in this paper.
We report the mean score across the three runs.
The judge configuration is held fixed across all evaluations, so the repeated runs primarily capture variation introduced by gameplay replay rather than changes in the evaluation setup.

\paragraph{Observed Stability.}
Table~\ref{tab:stability} summarizes the variation across 529 frozen artifacts,
every Godot artifact in this paper that was scored three times: the frozen base
of two generators, the versions Play2Code delivered, and the versions
\rsigame{} delivered.

Most artifacts are close to deterministic. The median range across three
independent runs is 0.73 points, 225 of the 529 score identically in all three,
and a quarter of them vary by less than a hundredth of a point. Improved
artifacts vary somewhat more than frozen ones (median 0.87 against 0.36), which
is what a build with more of a game in it should do.

\begin{table}[h]
\centering
\caption{
Score variation across three independent end-to-end evaluations of the same
frozen artifact. Each run uses a fresh gameplay replay and fresh judging by the
Qwen3.8-27B judge. Range denotes the difference between the highest and lowest
Overall scores across the three runs; the columns are its quartiles.
}
\label{tab:stability}
\small
\begin{tabular*}{\textwidth}{@{\extracolsep{\fill}}lccccc}
\toprule
Subset & $n$ & p25 & Median & p75 & Identical 3/3 \\
\midrule
All artifacts &  529 & 0.00 & 0.73 & 3.33 & 225 \\
Base $P_0$    &  266 & 0.00 & 0.36 & 2.33 & 131 \\
Improved      &  263 & 0.00 & 0.87 & 3.50 & \phantom{0}94 \\
\bottomrule
\end{tabular*}
\end{table}

\begin{table}[H]
\centering
\caption{Replay variation by task family, for the families with at least twelve
scored artifacts, ordered by median range. Same 529 artifacts and same
definition of range as Table~\ref{tab:stability}.}
\label{tab:stability-family}
\small
\begin{tabular*}{\textwidth}{@{\extracolsep{\fill}}lcc@{\hskip 2em}lcc}
\toprule
Family & $n$ & Median & Family & $n$ & Median \\
\midrule
sports      & 14 & 4.66 & shooter    & 24 & 0.44 \\
horror      & 19 & 3.37 & puzzle     & 32 & 0.42 \\
openworld   & 62 & 1.41 & simulation & 24 & 0.28 \\
visualnovel & 32 & 1.17 & strategy   & 63 & 0.00 \\
platformer  & 77 & 0.87 & rhythm     & 21 & 0.00 \\
tycoon      & 58 & 0.73 & idle       & 16 & 0.00 \\
roguelike   & 53 & 0.62 & racing     & 16 & 0.00 \\
cardgame    & 18 & 0.50 &            &    &      \\
\bottomrule
\end{tabular*}
\end{table}

\paragraph{Which Games Vary.}
Splitting the 529 artifacts by task family locates the variance rather than
leaving it as a property of ``game dynamics''
(Table~\ref{tab:stability-family}). The split is not fast against slow: the
three most variable families are \emph{sports}, \emph{horror} and
\emph{openworld}, while \emph{racing} and \emph{rhythm} --- as real-time as
anything in the benchmark --- have a median range of exactly zero, alongside
\emph{strategy}. What the variable families share is that their rubric is
answered late. Whether a match was won, whether a haunting resolved, whether a
world was traversed is decided by where twenty seconds of play happens to end
up; whether a beat landed on time or a lap timer ran is decided by frames the
replay reproduces every time. Variance tracks how far into a session the
evidence for a criterion arrives, not how fast the game moves.

This also says where a single-run comparison is least safe. A one-point
difference is meaningful for a rhythm game and is noise for a sports game, and
is one reason the per-family tables of Appendix~\ref{app:families} should be
read with their family's variance in mind.

\paragraph{Cross-Judge Robustness.}
To test whether the main results depend on a particular evaluation model, we
re-score 40 family-stratified Godot tasks with GPT-5.5, using exactly the same
replays, sampled frames, task-specific rubrics, and scoring formula as the
Qwen3.8-27B judge; only the judge model is changed.

As shown in Table~\ref{tab:crossjudge}, the two judges produce the same
method-level ordering,
\(\rsigame{} > \text{Play2Code} > \text{Base}\), with similar improvement
margins. The gain of \rsigame{} over the frozen base is \(+14.04\) points under
Qwen3.8-27B and \(+12.04\) under GPT-5.5, while its gain over Play2Code is
\(+13.47\) and \(+11.76\), respectively; all four bootstrap intervals remain
well above zero. The judges also agree on the direction of the
\rsigame{}-over-base comparison for 37 of 40 tasks. Although their absolute
scores are only moderately correlated (\(\rho=0.64\)), the conclusions drawn
from the relative comparisons remain stable across model families.

\begin{table}[H]
\centering
\caption{
\textbf{Cross-judge robustness on 40 family-stratified Godot tasks.}
Qwen3.8-27B and GPT-5.5 score the same replay recordings with the same rubric.
Both judges preserve the method ordering and yield similar pairwise improvement
margins. Confidence intervals are 95\% task-level bootstrap intervals.
}
\label{tab:crossjudge}
\small
\setlength{\tabcolsep}{4pt}
\begin{tabular*}{\textwidth}{@{\extracolsep{\fill}}lccccc}
\toprule
& \multicolumn{2}{c}{Overall} & \multicolumn{3}{c}{} \\
\cmidrule(lr){2-3}
Method
& Qwen3.8-27B
& GPT-5.5
& & & \\
\midrule
Base (frozen \(P_0\))
& 49.07
& 50.34
& & & \\

\quad \(+\) Play2Code
& 49.63
& 50.62
& & & \\

\quad \(+\) \rsigame{}
& \textbf{63.10}
& \textbf{62.38}
& & & \\

\midrule
Comparison
& \(\Delta\) Qwen
& \(\Delta\) GPT-5.5
& 95\% CI Qwen
& 95\% CI GPT-5.5
& Agreement \\
\midrule

\rsigame{} \( - \) Base
& \(+14.04\)
& \(+12.04\)
& \([+9.84,+18.59]\)
& \([+9.53,+14.60]\)
& 37/40 \\

\rsigame{} \( - \) Play2Code
& \(+13.47\)
& \(+11.76\)
& \([+8.95,+18.19]\)
& \([+7.79,+16.26]\)
& 34/40 \\

Play2Code \( - \) Base
& \(+0.56\)
& \(+0.28\)
& \([-3.45,+4.21]\)
& \([-3.83,+3.83]\)
& 24/40 \\

\bottomrule
\end{tabular*}

\vspace{1mm}
{\footnotesize
\emph{Agreement} counts tasks on which the two judges prefer the same method.
}
\end{table}



\subsection{Free-Play Evaluation}
\label{app:freeplay}

\paragraph{Protocol.}
GameCraft-Bench evaluates a game by replaying the demonstration traces packaged with its submitted project. These traces are visible development resources, rather than hidden benchmark trajectories, and \rsigame{} reuses them for controlled before--after comparisons across checkpoints. Although the hidden rubric, scores, and judge feedback are never exposed during development,
repeated use of the same demonstrations could still favor improvements specific to those interaction trajectories. We therefore evaluate whether the gains persist under independently generated gameplay.

We sample 20 family-stratified Godot tasks. For each task, the frozen base, Play2Code, and \rsigame{} builds are independently played by blind agents with 40 actions and one reset. The agents see only the public game specification and the running game---not the submission demonstrations, rubric, score, source code, development history, or method identity---and return factual reports of their observations. A separate judge compares these reports pairwise, yielding 60 comparisons in total.

\paragraph{Evaluation.}
The pairwise judge follows the same four quality dimensions and relative weighting used in the main evaluation, while judging only evidence observed during free interaction. Importantly, the replay scripts are never shown to the play agents: they choose their own actions, timing, and trajectories. This changes the interaction distribution while keeping the definition of game quality aligned with the main evaluation.

\paragraph{Results.}
As shown in Table~\ref{tab:freeplay}, \rsigame{} is preferred over the frozen base on 17 of 20 tasks and over Play2Code on 18 of 20, while Play2Code does not show a clear advantage over the base. Across all decided comparisons, the free-play evaluation agrees with the benchmark ordering in 45 of 59 cases.
These results indicate that the gains of \rsigame{} persist under independently chosen gameplay trajectories rather than being confined to the fixed replay scripts used during development and benchmark evaluation.

\begin{table}[H]
\centering
\caption{
\textbf{Free-play evaluation on 20 family-stratified Godot tasks.}
Blind agents independently play each build, and a separate judge compares their reports using the same quality dimensions and weighting as the main evaluation.
\(p\) is a two-sided sign test over decided comparisons.
}
\label{tab:freeplay}
\small
\begin{tabular*}{\textwidth}{@{\extracolsep{\fill}}lccc}
\toprule
Comparison & W/L/T & \(p\) & Agreement \\
\midrule
\rsigame{} -- Base
& \textbf{17/3/0} & \textbf{0.003} & \multirow{3}{*}{45/59} \\
\rsigame{} -- Play2Code
& \textbf{18/2/0} & \textbf{\(<0.001\)} & \\
Play2Code -- Base
& 9/11/0 & 0.824 & \\
\bottomrule
\end{tabular*}

\vspace{1mm}
{\footnotesize
W/L/T denotes wins/losses/ties for the first method.
Agreement counts decided comparisons for which free-play evaluation and the
benchmark score prefer the same build.
}
\end{table}

\subsection{Statistical Reliability of Main Results}
\label{app:significance}

\paragraph{Paired task-level uncertainty.}
All main-table comparisons are paired by task. We estimate uncertainty in the
mean Overall difference \(\Delta\) using \(20{,}000\) bootstrap resamples of
the 140 benchmark tasks, and additionally report a Wilcoxon signed-rank test
over the paired task-level differences. We separately resample the three
replay-and-score runs of each fixed task to measure evaluation noise; these
intervals are substantially narrower than the task-bootstrap intervals,
indicating that benchmark-task variation dominates replay and judge noise.

\paragraph{Main comparisons.}
\rsigame{} significantly improves over its frozen starting point in every
engine--generator setting inlcuded in Table \ref{tab:significance}, with gains of \(+8.8\) to \(+14.3\) Overall points
and task-bootstrap intervals well above zero. Against Play2Code, the improvement
is significant for both Godot generators and for GPT-5.5 on Phaser; the
Qwen3.8-27B Phaser comparison remains unresolved
(\(+1.55\), \(p=0.394\)).

\begin{table}[H]
\centering
\caption{
\textbf{Paired significance of the main results.}
\(\Delta\) is the mean task-level Overall difference.
Confidence intervals are obtained from \(20{,}000\) paired bootstrap resamples
of benchmark tasks; \(p\) is from a Wilcoxon signed-rank test.
W/L counts tasks on which the first method scores higher/lower.
}
\label{tab:significance}
\small
\begin{tabular*}{\textwidth}{@{\extracolsep{\fill}}lcccc}
\toprule
Comparison & \(\Delta\) & 95\% CI & \(p\) & W/L \\
\midrule

\multicolumn{5}{l}{\emph{Godot}} \\
GPT-5.5: \rsigame{} -- Base
& +14.26 & [11.49, 17.06] & \(<10^{-4}\) & 112/16 \\

GPT-5.5: \rsigame{} -- Play2Code
& +13.79 & [10.89, 16.65] & \(<10^{-4}\) & 118/19 \\

Qwen3.8-27B: \rsigame{} -- Base
& +10.70 & [7.75, 13.77] & \(<10^{-4}\) & 100/31 \\

Qwen3.8-27B: \rsigame{} -- Play2Code
& +7.24 & [4.49, 10.04] & \(<10^{-4}\) & 92/41 \\

\midrule
\multicolumn{5}{l}{\emph{Phaser}} \\

GPT-5.5: \rsigame{} -- Base
& +8.77 & [6.13, 11.51] & \(<10^{-4}\) & 101/35 \\

GPT-5.5: \rsigame{} -- Play2Code
& +3.11 & [0.53, 5.61] & 0.008 & 83/53 \\

Qwen3.8-27B: \rsigame{} -- Base
& +10.21 & [6.99, 13.70] & \(<10^{-4}\) & 95/36 \\

Qwen3.8-27B: \rsigame{} -- Play2Code
& +1.55 & [-1.50, 4.66] & 0.394 & 73/59 \\

\bottomrule
\end{tabular*}
\end{table}

\subsection{Adaptive Direction Policy Evaluation}
\label{app:direction_eval}

\paragraph{Experimental Setup.}
We compare adaptive development with a round-robin direction policy on
16 paired games initialized from the same frozen base projects. Each policy is
run for 30 development rounds with the same agent models, tools, and per-round
budget; the only difference is how the development focus is selected.
Figure~\ref{fig:direction}a,b analyzes the resulting trajectories using signals
that are not exposed to either policy during development.
We define a \emph{broken build} as a round whose resulting project fails to
produce a runnable game; the following round is then aligned at offset \(+1\)
relative to this failure event.
For state-conditioned analysis, we additionally use a held-out diagnostic
rubric to identify the currently lagging quality dimension. This rubric is
used only for post-hoc analysis and is never provided to the development loop.
Shaded areas in Figure~\ref{fig:direction}a are $\pm 1$ standard error over the corresponding break events or grouped development rounds.

\paragraph{Blind Pairwise Evaluation.}
To compare the final games produced by the two policies, we perform 63 blind
pairwise comparisons using their round-30 checkpoints.
For each comparison, the same demo trace is replayed on both builds and the
resulting observations are presented side by side to the evaluator.
The left--right assignment is randomized independently for each comparison,
and no information about the underlying direction policy is revealed.
Evaluators report an overall preference as well as preferences along the
Mechanics, Depth, Visuals, and Art dimensions; ties are allowed.
To check for position bias, a subset of comparisons is evaluated again with
the two sides swapped. These repeated judgments are used only as a consistency
check and are excluded from the reported 63 comparisons.
The pairwise evaluator is independent of the GameCraft-Bench benchmark
evaluator and has no access to its scores, rubrics, or feedback.

\subsubsection{Pairwise Judging Criteria}
\label{app:pairwise_rubric}

The criteria given to the blind pairwise judge of
Section~\ref{sec:further}, reproduced in full. They are deliberately not the
benchmark's rubric: the judge compares two builds on what its frames show,
while the benchmark scores one build against the task. Rule 1 is the one that
does the most work --- without it a broken build collects credit for the frames
it does manage to render.

\begin{rsipromptbox}{PromptCompare}{Comparison Rubrics. Blind pairwise judge}
You are an expert game reviewer. You compare two builds of the same game, Game
1 and Game 2. Both were developed from the same starting game and the same
design document, which is given below. Each build was played with exactly the
same scripted input (a "demo"), and you see frames sampled at the same fixed
interval from that recording, in time order. Judge only what is visible in the
frames.

For each dimension below, decide which build is better: "1", "2", or "tie".

## Core Mechanics
Does the scripted input visibly produce the gameplay the design document
describes?
- Better: actions have clear on-screen effects (movement, attacks, placement,
  state changes, score or resource changes); the core loop of the design
  document can be observed.
- Worse: input has no visible effect; the game stays on a title or menu screen;
  a scene fails to load; the screen is blank or frozen.

## Content Depth
How much of the designed game is present and reached in this demo?
- Better: more of the distinct mechanics, enemies, levels, events, progression,
  or win/lose states from the design document appear.
- Worse: a single repeated screen; placeholder content; systems that are
  announced in the UI but never shown.

## Functional Visuals
Can a player read the game state?
- Better: HUD, feedback, and important objects are legible and clearly
  separated from the background; text is not clipped or overlapping; sprites
  and animations are stable from frame to frame.
- Worse: unreadable or overlapping text; flickering or inconsistent sprites;
  key objects hidden behind others; no visible feedback for important events.

## Presentation & Art
Does it look like a coherent, finished game?
- Better: a consistent art style across screens; intentional composition;
  assets that match the theme of the design document.
- Worse: primitive shapes where art is expected; mixed or clashing styles;
  visual noise; decoration that makes the game harder to read.

## Overall
Which build would a player of this design document prefer, all things
considered?

## Rules
1. A build that is broken in these frames (blank, crashed, frozen, stuck on a
   screen the demo should have left) loses every dimension it fails to show,
   however good its other frames look.
2. More assets, more effects, or busier frames are not better by themselves.
   Prefer them only when they serve the design and keep the game readable.
3. Ignore which side a build is shown on. Game 1 and Game 2 are in random
   order.
4. Answer "tie" when the two builds are not visibly different on a dimension,
   or when both fail it equally.
5. Each reason must point to something visible, naming the build and the frame
   (for example "Game 2, frame 7: the score counter overlaps the timer").
\end{rsipromptbox}

\subsection{Agentic Verification Audit}
\label{app:verification_eval}

\paragraph{Audit Protocol.}
We evaluate the reliability of verification on development trajectories from
40 games initialized by GPT-5.5. We sample pre- and post-improvement decisions
from these trajectories and provide the complete interaction evidence associated
with each decision to Claude Opus 5~\citep{claudeopus5} for independent
adjudication. Each case is evaluated twice; disagreements between the two
judgments are manually reviewed to obtain the final label. This evaluator is
used only for the verification audit and is fully isolated from the
GameCraft-Bench evaluator and its scores, rubrics, and feedback.

\paragraph{Pre-Improvement Grounding.}
We evaluate whether proposed improvement targets are supported by the evidence
available before editing. For each audited session, a target is labeled as
\emph{grounded} when the observed gameplay evidence supports the claimed
problem or improvement opportunity, and \emph{ungrounded} when it is unsupported
or contradicted by the evidence. We report \emph{grounded precision}, the
fraction of decidable targets that are grounded, and the average number of
\emph{ungrounded targets per round}. We compare the agentic evidence-grounding
procedure used by \rsigame{} against a free-form critic operating on the same
development context.

\paragraph{Post-Improvement Failure Detection.}
We further evaluate whether verification can detect unsuccessful edits after the updated game is executed. An improvement is labeled unsuccessful when the intended change is not observable in gameplay or when the edit introduces a regression. Treating unsuccessful improvements as the positive class, we report failure recall, specificity, and balanced accuracy over 48 adjudicated rounds. We compare replay-based agentic verification with a build-only baseline that checks whether the updated project executes successfully but does not inspect its interactive behavior.

\subsection{Multi-Stage Guidance Evaluation}
\label{app:stage_eval}

Section~\ref{sec:further} tests one stage of development. This appendix
specifies how we ask whether a second and a third stage keep paying: each pair
of stage champions is compared by blind pairwise play, so a later stage has to
be preferred as a game rather than merely score higher.

\begin{wrapfigure}{r}{0.46\textwidth}
  \vspace{-\intextsep}
  \centering
  \includegraphics[width=0.44\textwidth]{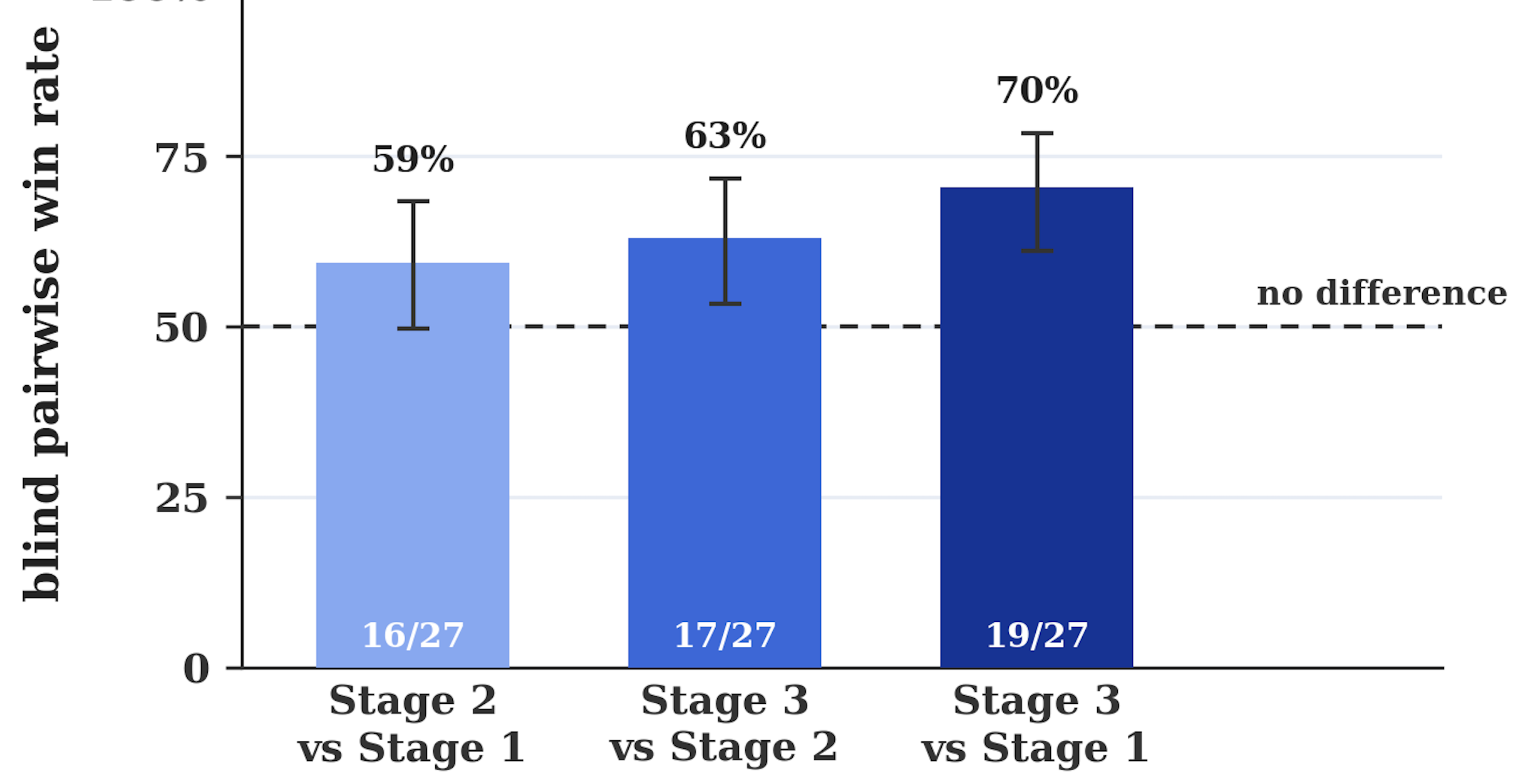}
  \caption{Blind pairwise win
  rate of the later stage in each pair, with $\pm 1$ s.e.}
  \label{fig:stages}
  \vspace{-1.5\intextsep}
\end{wrapfigure}

\paragraph{What is compared.} Three pairs of builds, each the champion the
Global Quality Monitor delivered at the end of a stage: stage~2 against
stage~1, stage~3 against stage~2, and stage~3 against stage~1. Both builds of a pair come from the same game, so each comparison is within-task, with the same generator, the same task specification and the same per-stage budget on both sides.
The result shows that later stages are consistently preferred over earlier ones, indicating that high-level guidance can repeatedly reopen development headroom after saturation and support continued multi-stage evolution.

\paragraph{Why pairwise.} A stage that repairs a broken mechanic and a stage that adds an unused menu can move the benchmark score by the same amount. This evaluation therefore asks which build a player would rather have, and do not use the benchmark score as the outcome.

\paragraph{Three agents, one build each.} A comparison uses three agents. Two \emph{play agents} each receive one build and never see the other, nor any score, source file, round number or development history; each returns a fixed-format factual report --- whether the game starts, whether the core loop can be completed, how inputs respond, what rendered, and what is missing or
broken --- and is explicitly forbidden to give a verdict. A third \emph{judge agent} reads only the two reports, in an order randomised per comparison and recorded, and returns \emph{report~1 better}, \emph{report~2 better} or \emph{tie} with a confidence level. The judge is instructed to weigh a concretely named defect above a general impression, and to treat anything about the review harness --- how much game time a call advances, how long a build takes to boot, the action budget --- as shared by both builds and therefore not evidence about either.

\paragraph{Equal budgets.} Each play agent drives its build through the review tool of Appendix~\ref{app:director} under a fixed budget of 40 actions and one
reset, with \texttt{wait}, \texttt{look} and \texttt{state} free. Sessions are keyed by reviewer id and a reopened session resumes with the budget it has already spent, so each agent is given an id unique to its comparison and checks its own action counter immediately after opening; a session that reports a
non-zero count is discarded and that comparison re-run. Two builds compared
under unequal budgets yield a verdict that follows the budget rather than the build, which is why this check is part of the protocol rather than an
afterthought.

\paragraph{Reporting.} A comparison counts as a win for the later stage when the judge prefers the report belonging to it. Ties are reported separately and excluded from the win rate rather than split, since a tie states that the two reports do not separate the builds. Win rates carry $\pm 1$ standard error in Wilson form, which keeps the interval inside $[0,1]$ near the ends; at a few
dozen games per pair these intervals are wide, and the text says so rather than reading a ranking out of overlapping bars.

\section{Training Data and Experience Internalization}
\label{app:training}

\subsection{Training Corpus Construction and Curation}
\label{app:training_corpus}

The released corpus has two parts. \textbf{Base games} are the generated
projects themselves, with the run that produced each one: 1,998 de-duplicated
games (1,446 Godot, 552 Phaser), 1,070 of them paired with the generation
trajectory, plus 910 complete per-trial run trees over three generation
batches. \textbf{Supervision} is what we train on, extracted from those runs in
three stages and summarised in Table~\ref{tab:training-corpus}: S1 game
generation, S2 planning, and S3 improvement.
Generation trajectories are produced by GPT-5.5 and improvement trajectories by
GLM-5.3-Flash. Plans are not free-written: 1,237 are distilled from the file
tree of a finished game, and 870 come from an earlier plan set and are kept only
where every path they name exists in the matching game.

\begin{table}[h]
\centering
\caption{The released training corpus, counted from the published tables. A
\emph{decision} row is one assistant turn together with the history that
preceded it, so decision rows repeat context and their token counts are not
independent; the trajectory rows are the same recordings un-windowed and are
what to count tokens over. \emph{Verified} improvement rounds are those an
independent post-improvement verification judged to have achieved the stated goal;
the remainder are released as negatives rather than dropped.}
\label{tab:training-corpus}
\small
\begin{tabular*}{\textwidth}{@{\extracolsep{\fill}}lllrrr}
\toprule
Stage & Unit & Teacher & Godot & Phaser & Total \\
\midrule
S1 generation & trajectory & GPT-5.5 & 1,929 & 284 & 2,213 \\
S1 generation & decision row & GPT-5.5 & 66,312 & 28,097 & 94,409 \\
\midrule
S2 planning & plan & distilled & 1,979 & 129 & 2,108 \\
\midrule
S3 improvement & round & GLM-5.3-Flash & 3,730 & 273 & 4,003 \\
\quad of which verified & round & --- & 1,869 & 144 & 2,013 \\
S3 improvement agent & decision row & GLM-5.3-Flash & 111,302 & 9,930 & 121,232 \\
\quad of which verified & decision row & --- & 48,341 & 5,098 & 53,439 \\
\midrule
Games improved & game & --- & 345 & 43 & 388 \\
Distinct tasks & task & --- & 2,189 & 322 & 2,511 \\
Training rows & row & --- & 183,063 & 38,689 & 221,752 \\
\bottomrule
\end{tabular*}
\end{table}

\paragraph{Trajectory Serialization.}
Each generation session is serialized as a single multi-turn agent trajectory
containing the task specification, assistant outputs, tool calls, and tool
results.
Training loss is applied only to assistant-generated outputs, avoiding repeated
supervision over the shared interaction context.
Planning examples are extracted from the agent's initial implementation plan
before coding, while later progress updates to the same plan are discarded.
A decision row keeps a pinned prefix (system prompt, task, and the plan where
one exists), then as many recent whole decisions as fit, then the target
decision last, within a 32k-token budget.

\paragraph{What is filtered out.}
A generation trajectory is retained only when the session produced an executable
artifact; artifact existence is used rather than the agent SDK's reported
success status, which does not reliably indicate whether a runnable project was
produced. Empty trajectories and duplicated tasks are removed. A plan is kept
only if every file path it references exists in the finished game and its engine
matches the brief. An improvement round is dropped when its session failed or its diff
is empty, and is labelled positive only on an independent post-improvement
verification --- 1,869 of 3,730 Godot rounds and 144 of 273 Phaser attempts.
Verbose tool outputs such as build logs are middle-truncated, preserving the
executed command and its verdict. Whole-project diffs are compacted to the
source hunks the improvement actually changed, which shrinks an improvement example by
roughly $48\times$ at the median.

\paragraph{Evaluation contamination.}
Nothing from a baseline, ablation or evaluation run enters the corpus, and the
140 distinct benchmark task ids are excluded at build
time. The published tables were re-scanned against those ids after the fact:
zero overlaps in all 13 configurations, over a union of 2,511 task ids.

\paragraph{Recording gaps are flagged, not hidden.}
One agent SDK stored a one-line UI summary of a tool result instead of the
content the model actually received. The affected rows are kept and carry
explicit flags (\texttt{row\_has\_summary\_only},
\texttt{target\_follows\_summary\_only}, \texttt{tool\_results\_source},
\texttt{trajectory\_incomplete}), so a clean training subset is a filter rather
than a different release. This affects all 284 Phaser generation trajectories
(20.2\% of their rows have the target decision immediately after a summary) and
one of the two Godot improvement arms (31.7\% of multi-turn improvement rows); the other
Godot arm is complete.

\paragraph{Data sanitization.}
Every row, metadata column and document was rewritten member by member. Absolute
machine paths became \texttt{/workspace/game}, \texttt{/workspace/harness} or
\texttt{<path>} (403,280 rewrites); user, host, account and internal project
names became \texttt{<redacted>} (13,610); model-gateway hosts, internal mirrors
and cluster names were replaced; e-mail addresses and non-loopback IP literals
were removed; owner columns in captured directory listings were rewritten; and
26,577 configuration fields were neutralised in place, so the schema is
unchanged. Tool outputs that dumped the build host's process table were
\emph{deleted} rather than scrubbed (26 listings, 379 lines). Scientific
content --- game code and assets, rubric text, per-requirement scores, model
identifiers, reasoning effort, timings --- is untouched. The cleaner was
verified by a second, independently written scanner carrying 47 identity and
credential patterns plus a planted-leak self-test that requires every planted
leak to be removed; both report zero hits.

\subsection{Fine-Tuning Configuration}
\label{app:finetuning}

We fine-tune Qwen3.8-27B~\citep{qwen38} using parameter-efficient supervised fine-tuning with LoRA~\citep{lora} over 4-bit quantized base weights~\citep{qlora}.
Each agentic trajectory is treated as one long training document, allowing the
model to learn planning, coding, tool-use, and improvement decisions within their
original development context.
Unless otherwise specified, all training configurations use the same
hyperparameters; only the composition of the training corpus changes.

\paragraph{Training Mixture.}
The three variants are not three separate mixtures but three nested corpora:
each adds one stage of supervision to the one before it, so
\(\textsc{S1} \subset \textsc{S2} \subset \textsc{S3}\) and any difference
between two rows of Table~\ref{tab:internalize} is attributable to the stage
that was added. Both engines are trained together;
Table~\ref{tab:sft-mixture} gives the composition.

\begin{table}[h]
\centering
\caption{The training corpus of each variant in Table~\ref{tab:internalize}.
Each stage adds rows to the previous corpus; the cumulative column is what that
variant was trained on. Tasks in the held-out brief list are filtered from every
stage.}
\label{tab:sft-mixture}
\small
\begin{tabular*}{\textwidth}{@{\extracolsep{\fill}}llrrrr}
\toprule
Stage added & Unit & Godot & Phaser & Rows added & Cumulative \\
\midrule
S1 generation & agent trajectory & 876 & 418 & 1,294 & 1,294 \\
S2 planning & brief $\to$ file plan & 216 & \phantom{0}92 & \phantom{0,}308 & 1,602 \\
S3 improvement & improvement round & 509 & \phantom{00}0 & \phantom{0,}509 & 2,111 \\
\midrule
\multicolumn{2}{l}{\emph{Total}} & 1,601 & 510 & 2,111 & \\
\bottomrule
\end{tabular*}
\end{table}

\begin{table}[h]
\centering
\caption{Fine-tuning configuration used for experience internalization. These
are the settings of the released adapter; every variant in
Table~\ref{tab:internalize} is trained with them and differs only in the corpus
of Table~\ref{tab:sft-mixture}.}
\label{tab:sft-config}
\small
\begin{tabular}{ll}
\toprule
Configuration & Value \\
\midrule
Base model & Qwen3.8-27B \\
Adaptation & LoRA~\citep{lora} \\
LoRA rank \(r\) & 16 \\
LoRA \(\alpha\) & 32 \\
LoRA dropout & 0.05 \\
LoRA targets & All linear projections \\
Rank-stabilized LoRA (rsLoRA) & Yes \\
Base-weight precision & 4-bit NF4~\citep{qlora} \\
Double quantization & Yes \\
Compute precision & bfloat16 \\
Maximum context length & 57,344 tokens \\
Epochs & 1 \\
Optimizer & AdamW~\citep{adamw}, fused \\
Learning rate & \(5\times10^{-5}\) \\
Learning-rate schedule & Cosine \\
Warmup ratio & 0.03 \\
Weight decay & 0.01 \\
Per-device batch size & 1 \\
Gradient accumulation & None \\
Sequence-parallel degree & 4 \\
Hardware & \(4\times\) H100 80\,GB \\
\bottomrule
\end{tabular}
\end{table}

\paragraph{Long-Context Training.}
Agentic development trajectories can span tens of thousands of tokens.
We therefore use a maximum context length of 57,344 tokens and distribute each
sequence across four GPUs using Ulysses-style sequence parallelism~\citep{ulysses}.
Gradient checkpointing and a fused cross-entropy implementation are used to
reduce activation and output-head memory consumption.

\paragraph{Training Objective.}
For each trajectory, the original interaction context is preserved, while the
supervised loss is applied only to assistant-generated outputs.
This trains the model on the sequence of development decisions without treating
tool observations or environment feedback as prediction targets.
The full \rsigame{} model is trained on the union of planning, generation, and
independently verified improvement examples.

\subsection{Training-Data Ablation}
\label{app:internalize_ablation}
Table~\ref{tab:internalize} gives every variant compared in
Section~\ref{sec:further} (Figure~\ref{fig:global_analysis}c draws \textsc{+Gen} and
the full model). \textsc{Base} and the full model are the Qwen3.8-27B and
Qwen3.8-27B (SFT) rows of Table~\ref{tab:godot_main}.

\begin{table}[h]
\centering
\caption{One-shot Godot generation by training variant. Every row is over
the 140 tasks.}
\label{tab:internalize}
\small
\begin{tabular*}{\textwidth}{@{\extracolsep{\fill}}lccccc}
\toprule
Variant & Mechanics & Depth & Visuals & Art & Overall $\uparrow$ \\
\midrule
\textsc{Base}               & 41.2 & 33.4 & 38.6 & 38.3 & 37.07 \\
\textsc{+Gen}               & 49.0 & 41.4 & 42.0 & 38.0 & 41.45 \\
\textsc{+Gen+Plan}          & 48.5 & 40.8 & 42.6 & 39.5 & 41.77 \\
\textsc{+Gen+Plan+Improve}  & \textbf{56.1} & \textbf{47.5} & \textbf{49.8} & \textbf{44.8} & \textbf{48.22} \\
\bottomrule
\end{tabular*}
\end{table}


\section{Extended Evaluation and Case Studies}
\label{app:extra}

\subsection{Comparison with a Multi-Agent Generate-and-Verify System}
\label{app:vibegame}

\paragraph{Why this comparison.} This paper does not claim that an agent can
improve a game it generated --- several systems do that. It claims that doing so
\emph{reliably} needs a controller: something that decides where the headroom
is, keeps the best version, and stops. The closest public system we could run is
VibeGame\footnote{\url{https://github.com/tettethu/VibeGame}, Apache-2.0.}, an
eight-role agent team whose pipeline already contains verification and
optimisation roles. If a team of that shape closed the gap on its own, a
controller would be an ornament.

\paragraph{Setup.} Forty Phaser tasks of GameCraft-Bench, the VibeGame code
unmodified, all eight of its roles set to GPT-5.5, one trial per task, fully
automatic, no human intervention. The builds are scored with the same verifier,
rubrics and judge as every other Phaser arm in this paper.

\begin{table}[h]
\centering
\caption{VibeGame against one-shot generation and against \rsigame{}. Left: the
twelve tasks shared by VibeGame's forty and our forty-task Phaser development
subset, so every column is the same twelve games. Right: VibeGame's own forty
tasks, against the one-shot reference its release quotes on them.}
\label{tab:vibegame}
\small
\begin{tabular*}{\textwidth}{@{\extracolsep{\fill}}lrr}
\toprule
& the 12 shared tasks & VibeGame's own 40 \\
\midrule
VibeGame                              & 23.3 & 30.5 \\
\midrule
OpenGame $+$ GPT-5.5, one shot        & 50.1 & 46.0 \\
\quad $+$ Play2Code                   & 56.4 & --- \\
\evorow \quad $+$ \rsigame{}          & 62.4 & --- \\
\bottomrule
\end{tabular*}
\end{table}

\paragraph{What it shows.} An agent team that already verifies and optimises
scores \textbf{30.5}, against \textbf{46.0} for a single generation pass on the
same tasks: its loop does not convert rounds into quality. On the twelve tasks
we share it reaches 23.3, where the frozen $P_0$ of our Phaser arm is already at
50.1 and \rsigame{} brings those same twelve to 62.4. The presence of
verification and optimisation \emph{roles} in a pipeline is not what produces
sustained improvement.

\paragraph{Reading it fairly.} One trial per task, and a system we configured
only to the extent of setting its role models; we make no claim about VibeGame
under other models or a larger budget.


\subsection{Qualitative Evolution Case Studies}
\label{app:cases}

We follow three games from the build a code model generates to the build we
ship. Each case opens with a timeline: the official score at every third round
of the 30-round autonomous run, the version the Global Quality Monitor holds, the round
at which the saturation stop ends the run, and, past an axis break, the passes the loop
did not run --- the stage briefs written once the champion stopped being beaten,
and the development passes after them. Each build on that timeline then gets a
card of its own: what that round or pass found, what it changed, and the frames
it is judged on, with boxes placed by eye over the screenshots, green where
something arrived and red where it is still wrong. Every build of a game is
driven through the same script, so its panels are comparable.

\subsubsection{Holding the Best Version Through Regressions}
\label{app:case_hold}

\begin{figure}[H]
  \centering
  \includegraphics[width=\linewidth]{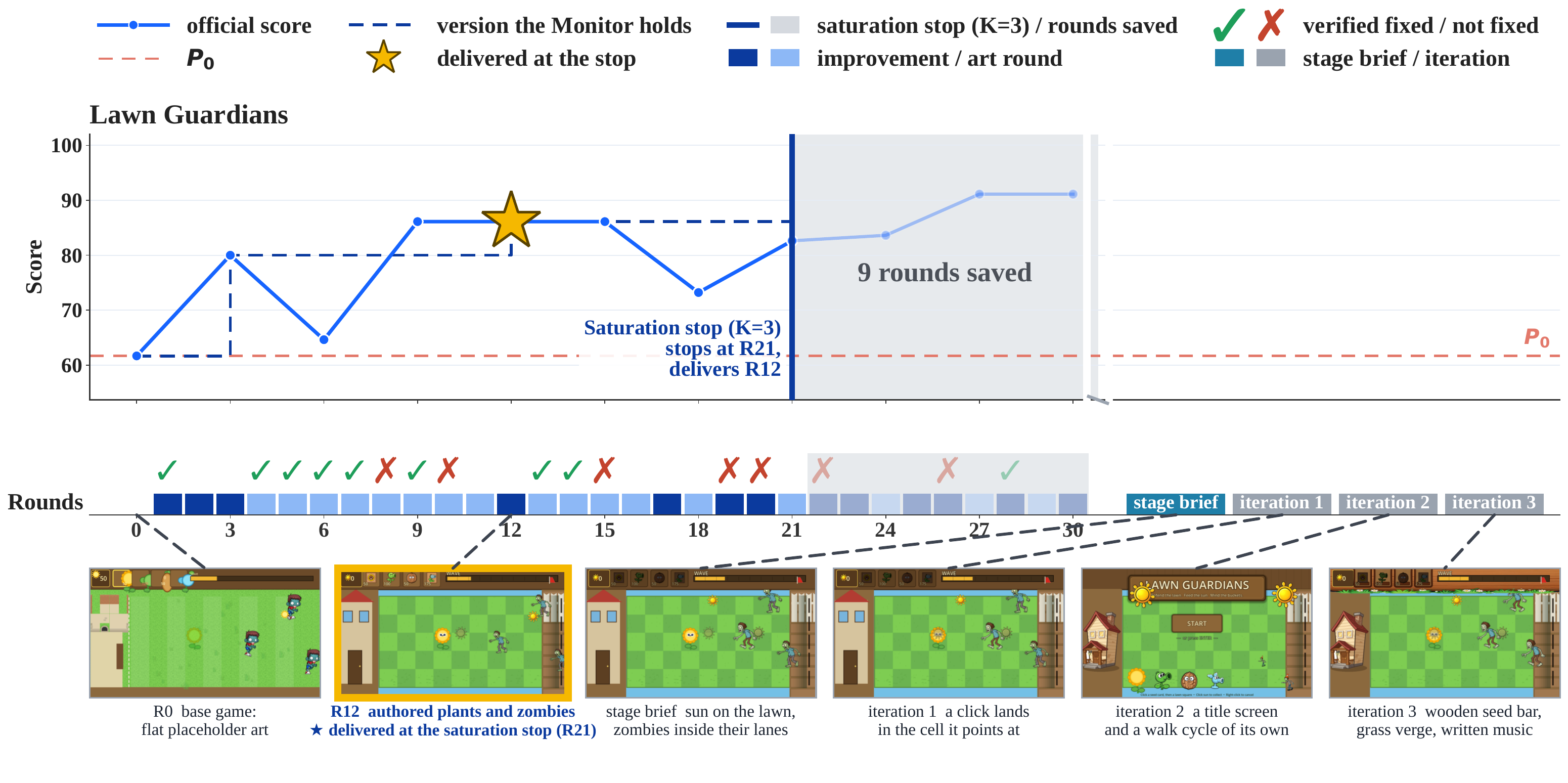}
  \caption{\emph{Lawn Guardians}, from the generated game to the build we
  shipped. The score band gives the official score at every third round of the
  30-round autonomous run, the version the Global Quality Monitor holds, and the star
  where the saturation stop delivers it at round 21, nine rounds early. The round band
  gives one cell per round --- improvement or art --- with the post-improvement verdict
  above it. Past the axis break are the passes the loop did not run: the stage
  brief written once the champion had survived three checkpoints, and the three
  iterations after it, which are development passes rather than loop rounds and
  so carry no round numbers. Each build below is opened up as a card in
  Figures~\ref{fig:case_lawn_stages_a}--\ref{fig:case_lawn_stages_c}.}
  \label{fig:case_lawn_overview}
\end{figure}

\begin{figure}[H]
  \centering
  \includegraphics[width=0.94\linewidth]{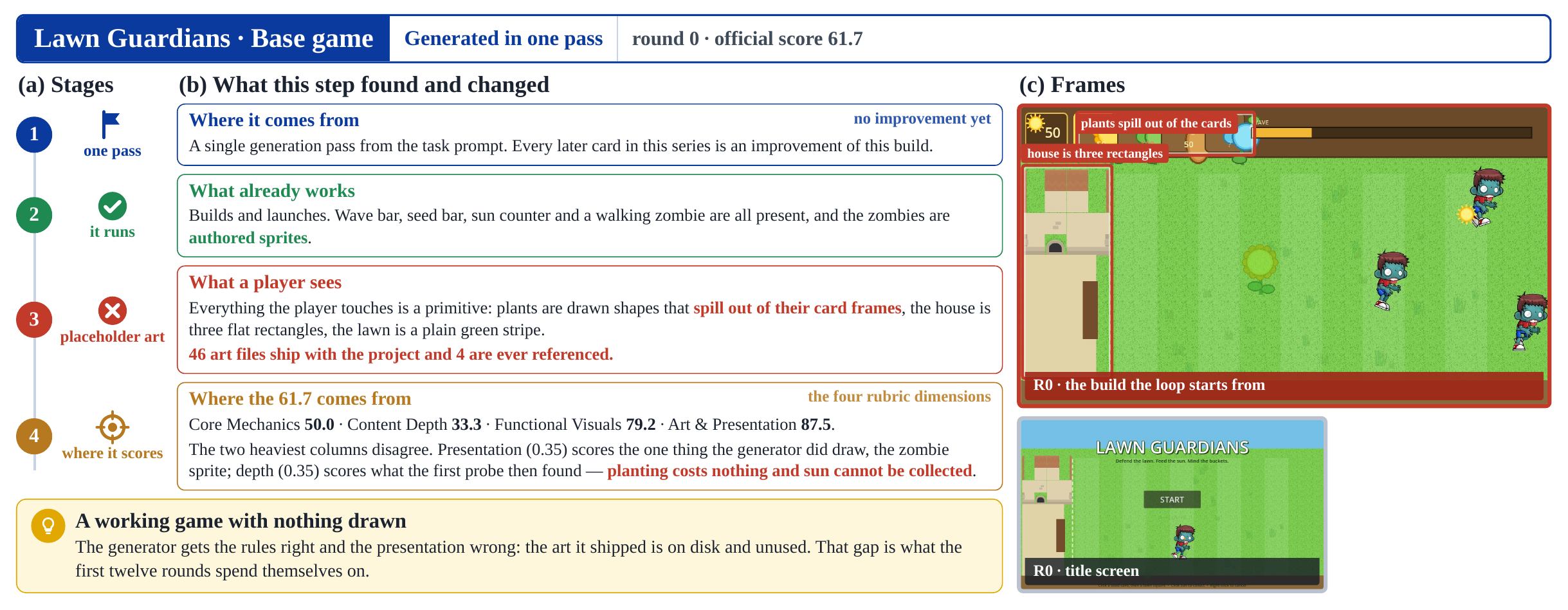}\\[3pt]
  \includegraphics[width=0.94\linewidth]{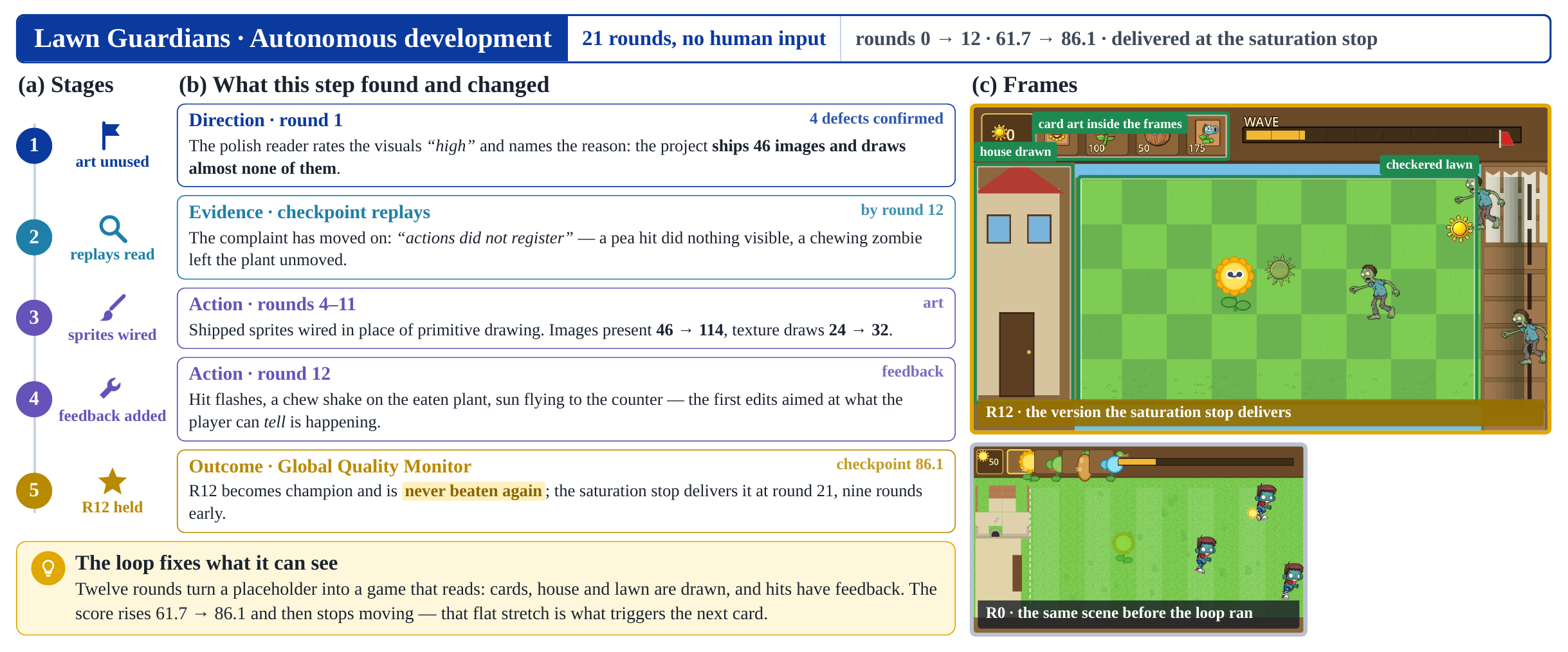}\\[3pt]
  \includegraphics[width=0.94\linewidth]{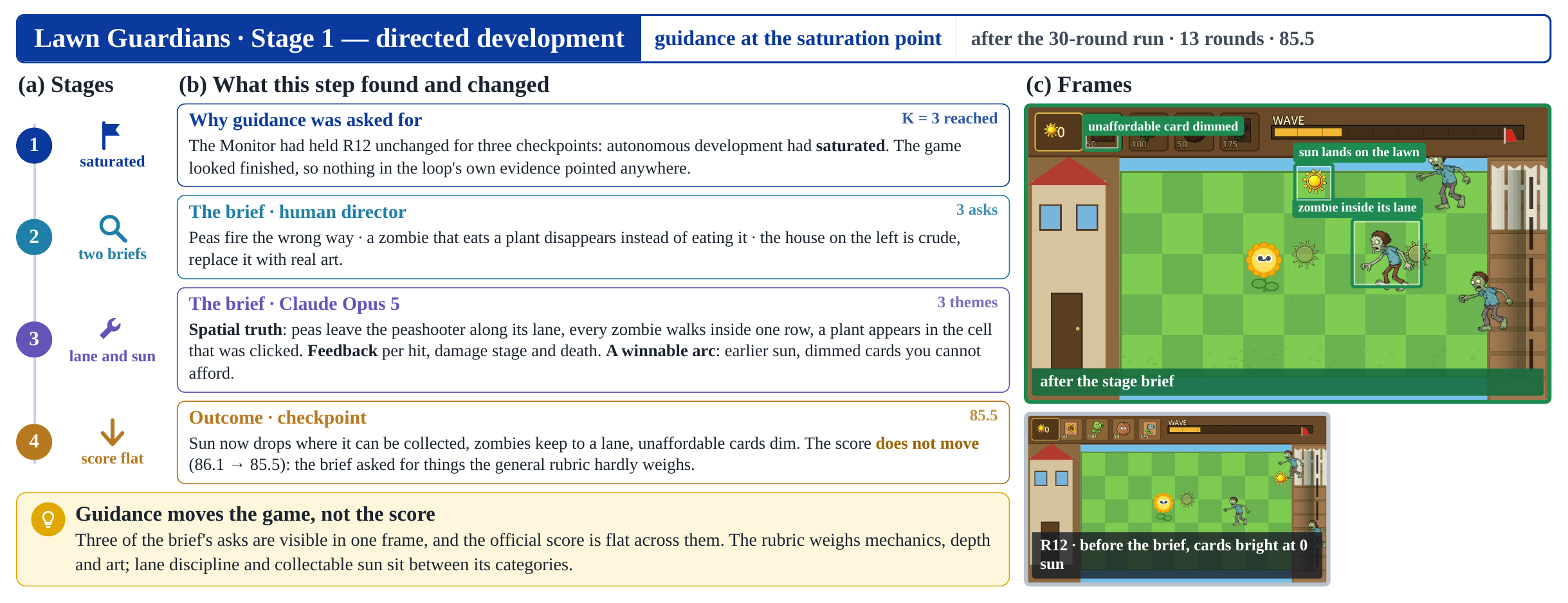}
  \caption{\emph{Lawn Guardians} from the generated build to the stage brief. Each card
  carries what that round or pass found, what it changed, and the frames it is
  judged on; boxes over the screenshots are placed by eye, green where something
  arrived and red where it is still wrong, and every build is driven through the
  same script so the scenes are comparable. \textbf{Base game:} the generator
  has the rules right and the presentation wrong --- the art it shipped is on
  disk and unused. \textbf{Autonomous:} twelve rounds wire that art in and add
  hit feedback, and the champion that emerges is never beaten.
  \textbf{Stage 1:} with the champion held for three checkpoints the loop has
  run out of its own evidence, and a brief written from outside --- by Claude
  Opus 5, alongside the human's --- puts sun on the lawn and zombies inside
  their lanes.}
  \label{fig:case_lawn_stages_a}
\end{figure}

Development does not end where the run does. The Global Quality Monitor's stop
criterion is about the run's own scores, not about the game: once the champion
survives three checkpoints, nothing in the loop's evidence points anywhere, and
a stage brief written from outside supplies the direction the loop no longer
has. Three further passes follow it --- controls, animation, and set dressing
with a written soundtrack --- each changing what the game is like to play rather
than what it does. The six builds are shown one card each in
Figures~\ref{fig:case_lawn_stages_a}--\ref{fig:case_lawn_stages_c}.

\begin{figure}[H]
  \centering
  \includegraphics[width=0.94\linewidth]{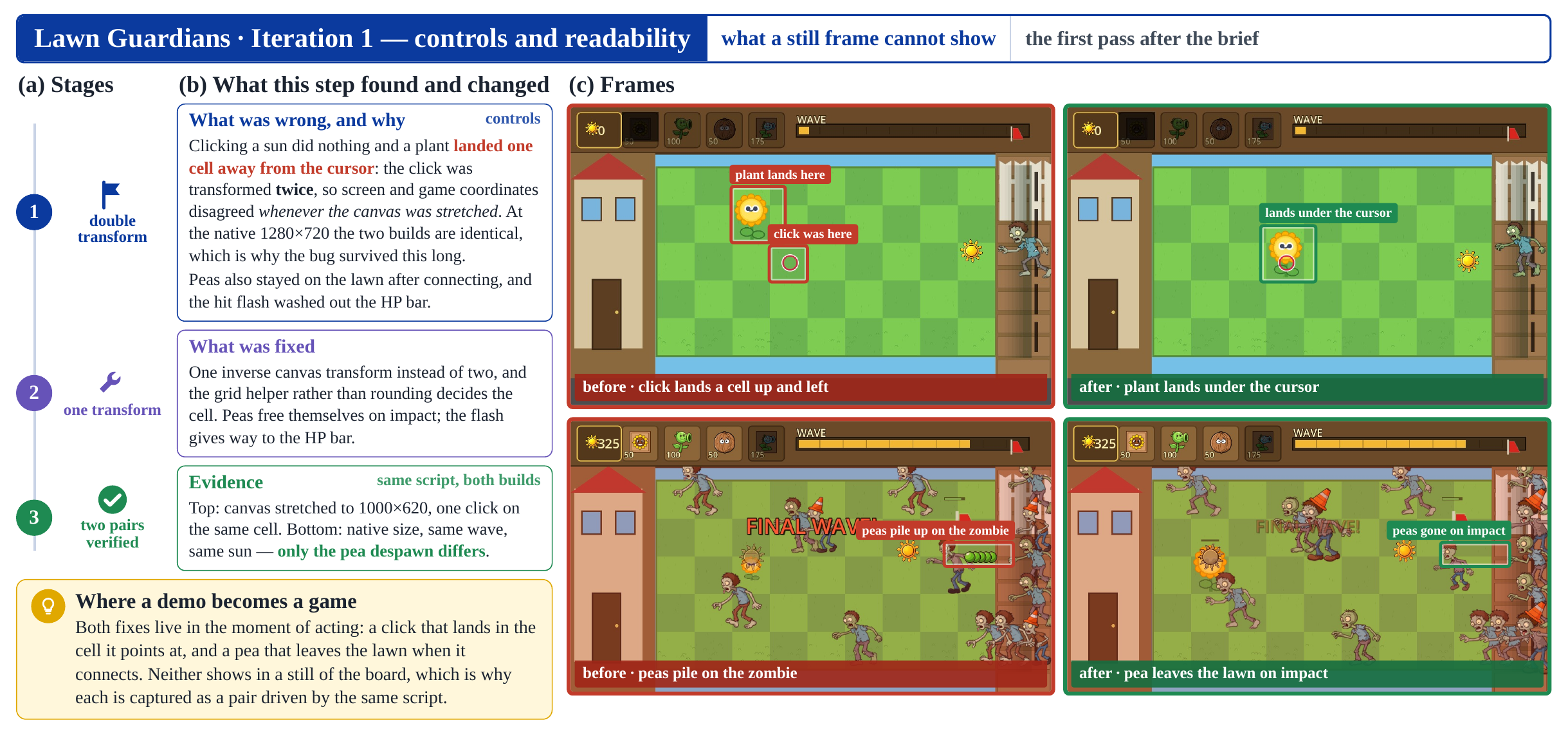}\\[3pt]
  \includegraphics[width=0.94\linewidth]{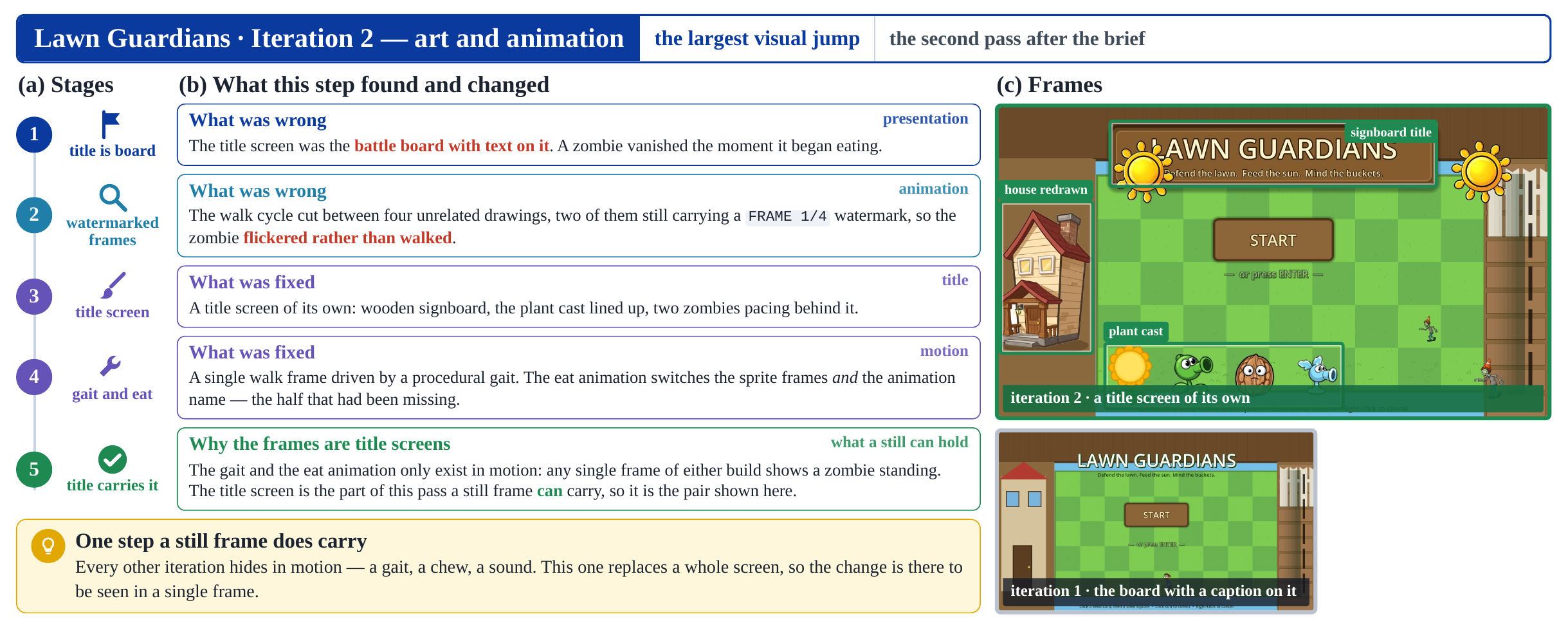}\\[3pt]
  \includegraphics[width=0.94\linewidth]{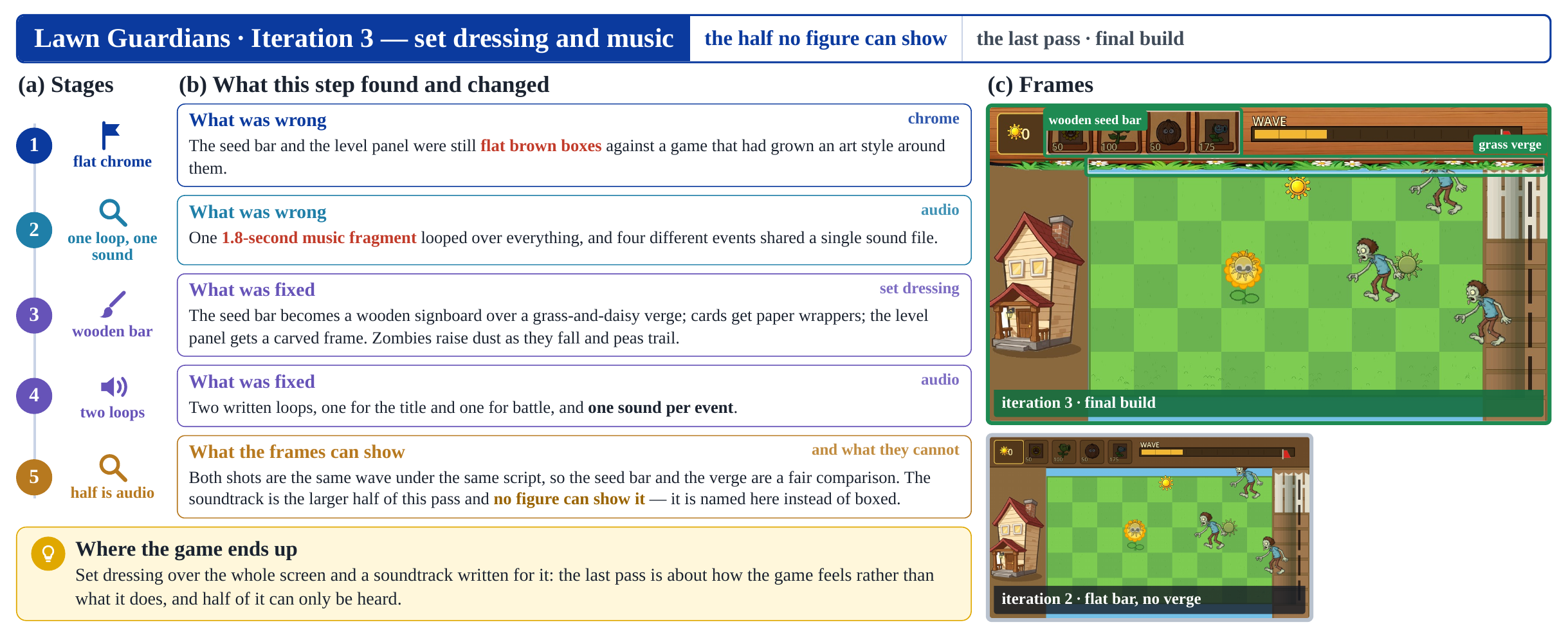}
  \caption{The three development passes that follow the stage brief.
  \textbf{Iteration 1} is a fix no still of the board can show, captured instead
  as two before/after pairs under the same script: at a stretched canvas the
  click was transformed twice, so a plant landed a cell from the cursor, and
  peas stayed on the lawn after connecting. \textbf{Iteration 2} replaces the
  title screen and the walk cycle, the one step a still frame does carry.
  \textbf{Iteration 3} is set dressing and a rewritten soundtrack; the seed bar
  and verge are in the frames, and the audio is the half no figure can show.}
  \label{fig:case_lawn_stages_b}
\end{figure}

\begingroup\phantomsection\label{fig:case_lawn_stages_c}\endgroup


In \emph{Lawn Guardians} (Figure~\ref{fig:case_lawn_overview}), progress is not
monotone: an art round (R6) and a feedback round (R18) each lower the score, and
the Global Quality Monitor adopts neither. It holds R12 (86.1, up from 61.7) until the saturation
stop delivers it at round 21. Two limits also show: a verification replay can be
too short to see a real fix (R26), and the official judge can reward a build
whose art covers the play field (R27). Play2Code peaks at 81.5 and ends at 65.7.

\subsubsection{Two Briefs on the Same Checkpoint}
\label{app:case_alley}

\emph{Alley Brawlers} is the case where the loop's own evidence buys the scenery
and loses the actors. Thirty autonomous rounds draw a lit street behind the
fight --- the largest single change in the run --- and in the same rounds the
fighters stop resolving: in the delivered build a body in motion is a torn
smear. The two cancel, and the champion the saturation stop delivers scores exactly what
the base game scored.

What the loop could not produce is the observation that nobody can win. Both
stage-1 directors played the build and said so, and the run that followed the
model's brief moved the score for the first time. A control branch, restarted
from the same checkpoint with no brief at all, reached 60.0; the briefs reached
61.5 and 62.0. A second brief, written at the stage-1 champion, then went after
what still made the match unplayable --- two bodies standing in the same space
--- and bought a readable exchange rather than a higher score.

\begin{figure}[H]
  \centering
  \includegraphics[width=\linewidth]{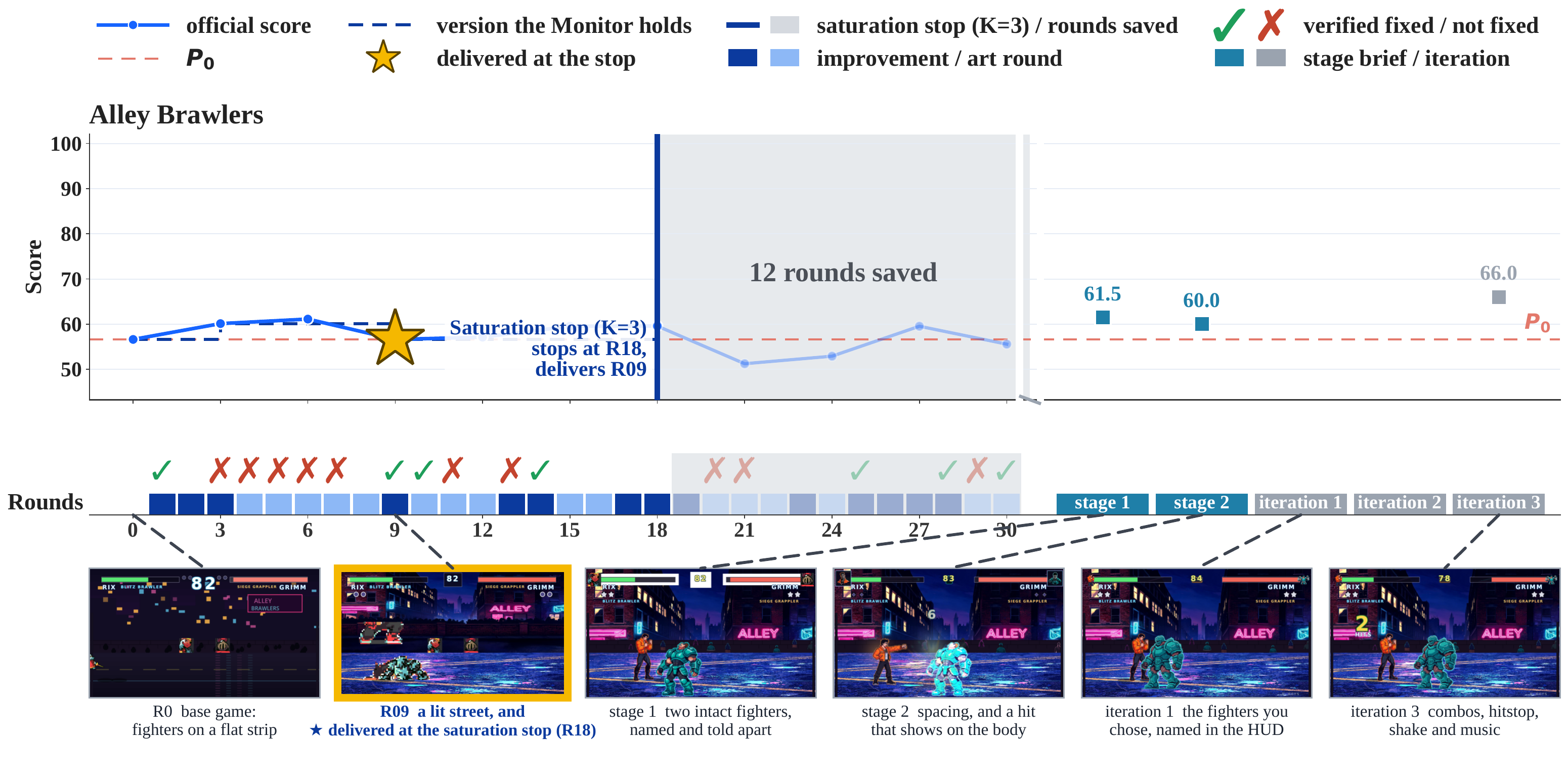}
  \caption{\emph{Alley Brawlers}, from the generated game to the build we
  shipped. Score band, round band and frames as in
  Figure~\ref{fig:case_lawn_overview}. The saturation stop ends the autonomous run at
  round 18 and delivers round 9. Past the axis break are the passes the loop did
  not run: two stage briefs, each written by a director who played the build in
  front of it, and three development passes. Stage 2 and the iterations both
  continue from the stage-1 champion, so they are alternatives rather than a
  sequence.}
  \label{fig:case_alley_overview}
\end{figure}

\begin{figure}[H]
  \centering
  \includegraphics[width=0.94\linewidth]{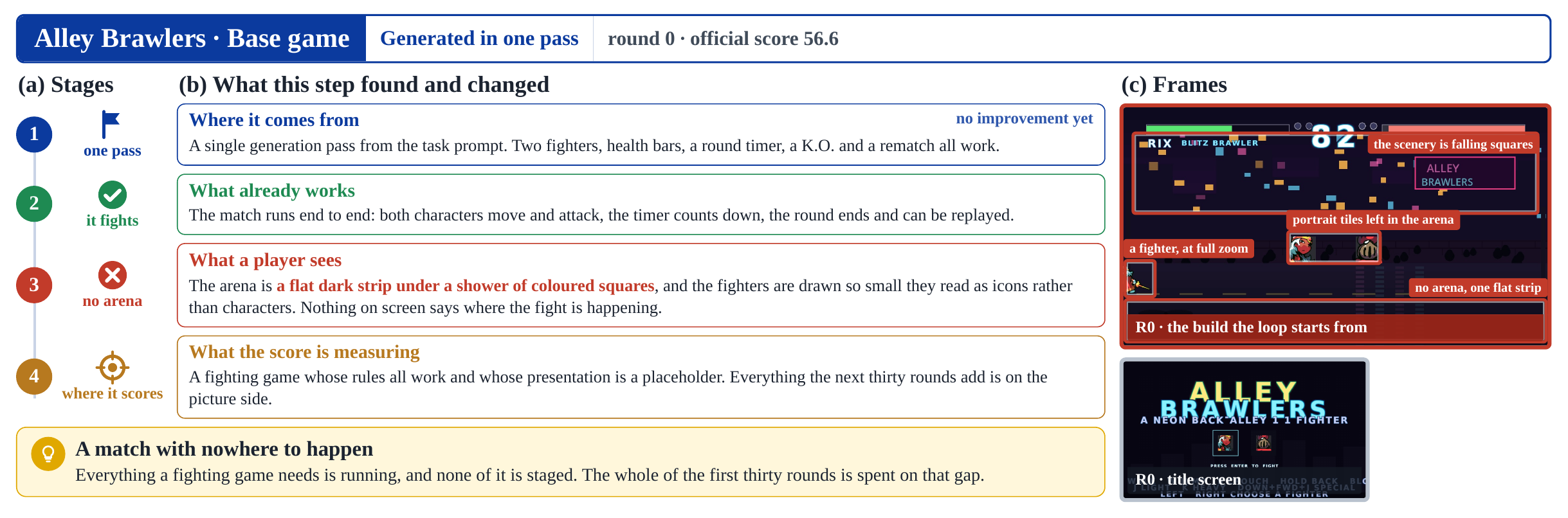}\\[3pt]
  \includegraphics[width=0.94\linewidth]{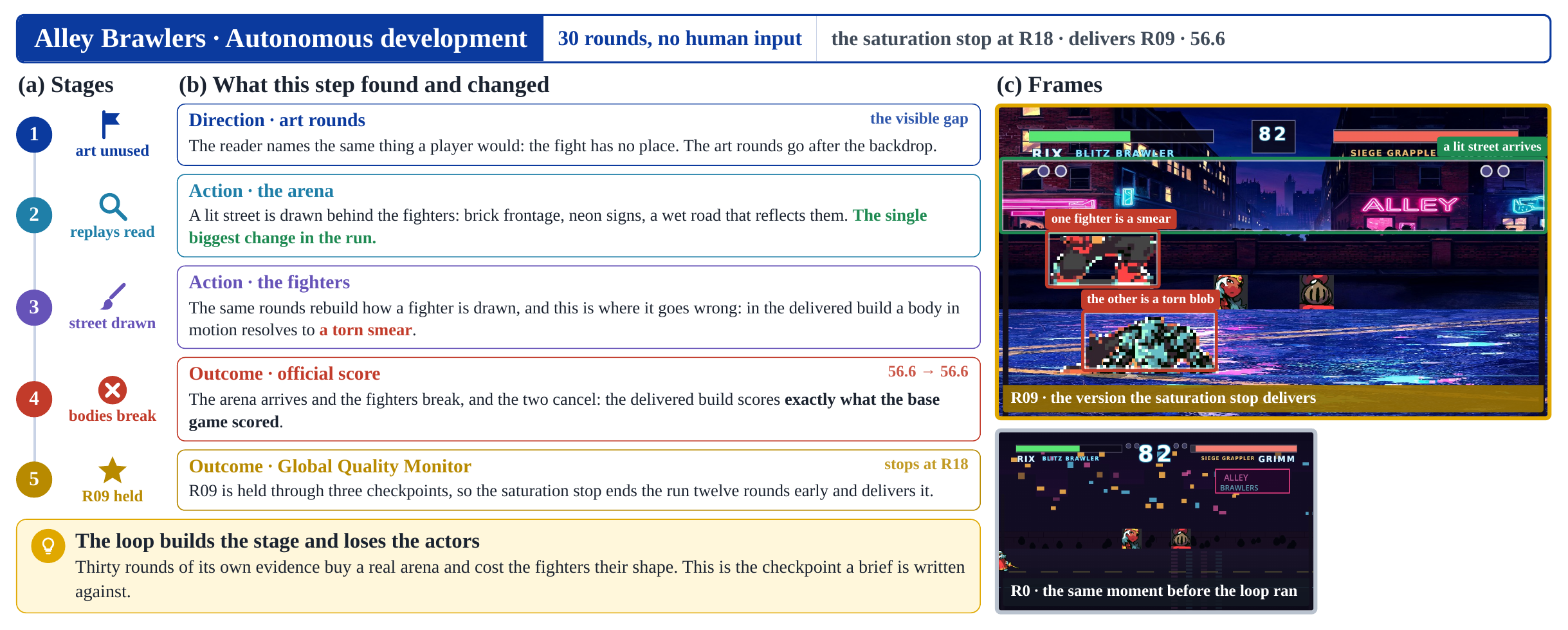}\\[3pt]
  \includegraphics[width=0.94\linewidth]{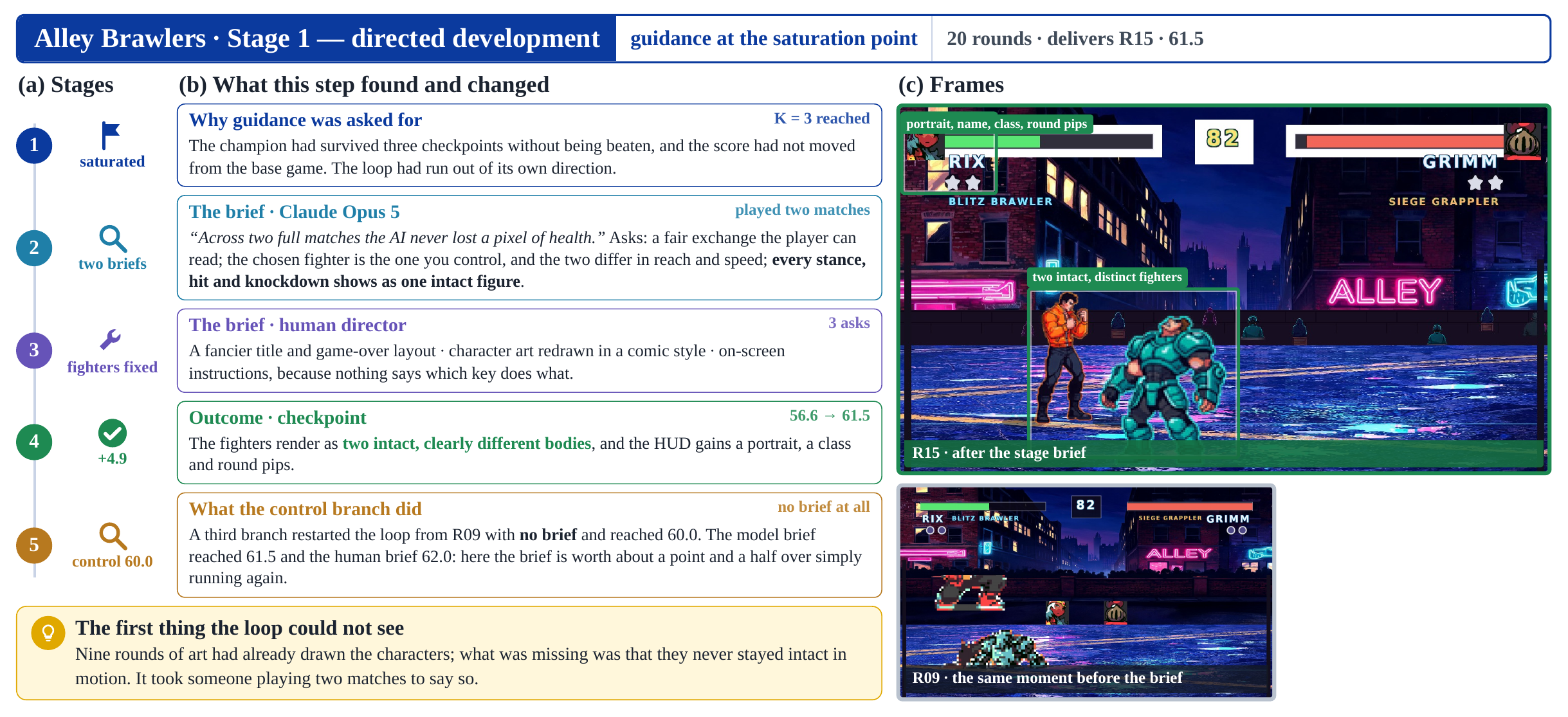}
  \caption{\emph{Alley Brawlers} from the generated build to the first brief. Cards,
  boxes and script as in Figure~\ref{fig:case_lawn_stages_a}.
  \textbf{Base game:} a working match on a flat strip under falling squares.
  \textbf{Autonomous:} the arena arrives and the fighters break, and the
  delivered build scores exactly what the base game scored. \textbf{Stage 1:} a
  director who played two matches asks for a fair exchange and for a fighter
  that stays one intact figure; the score moves for the first time in the run.}
  \label{fig:case_alley_a}
\end{figure}

\begin{figure}[H]
  \centering
  \includegraphics[width=0.94\linewidth]{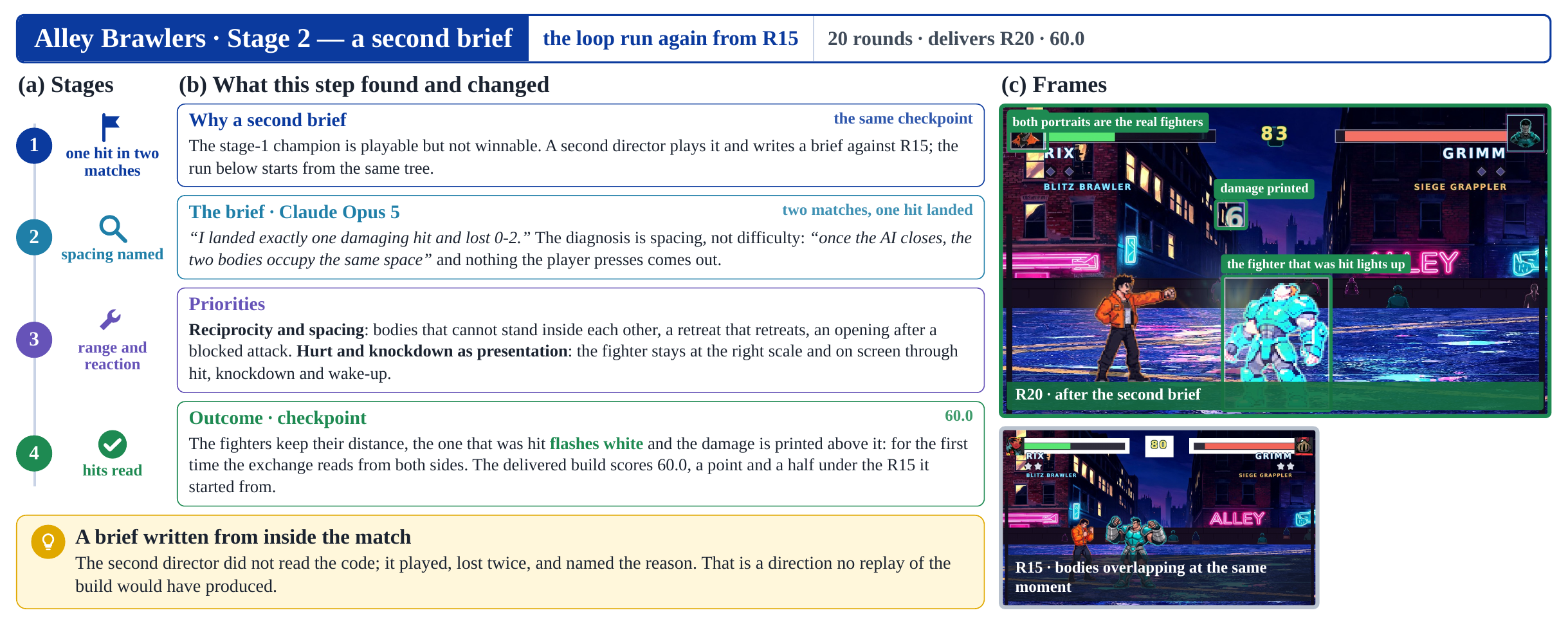}\\[3pt]
  \includegraphics[width=0.94\linewidth]{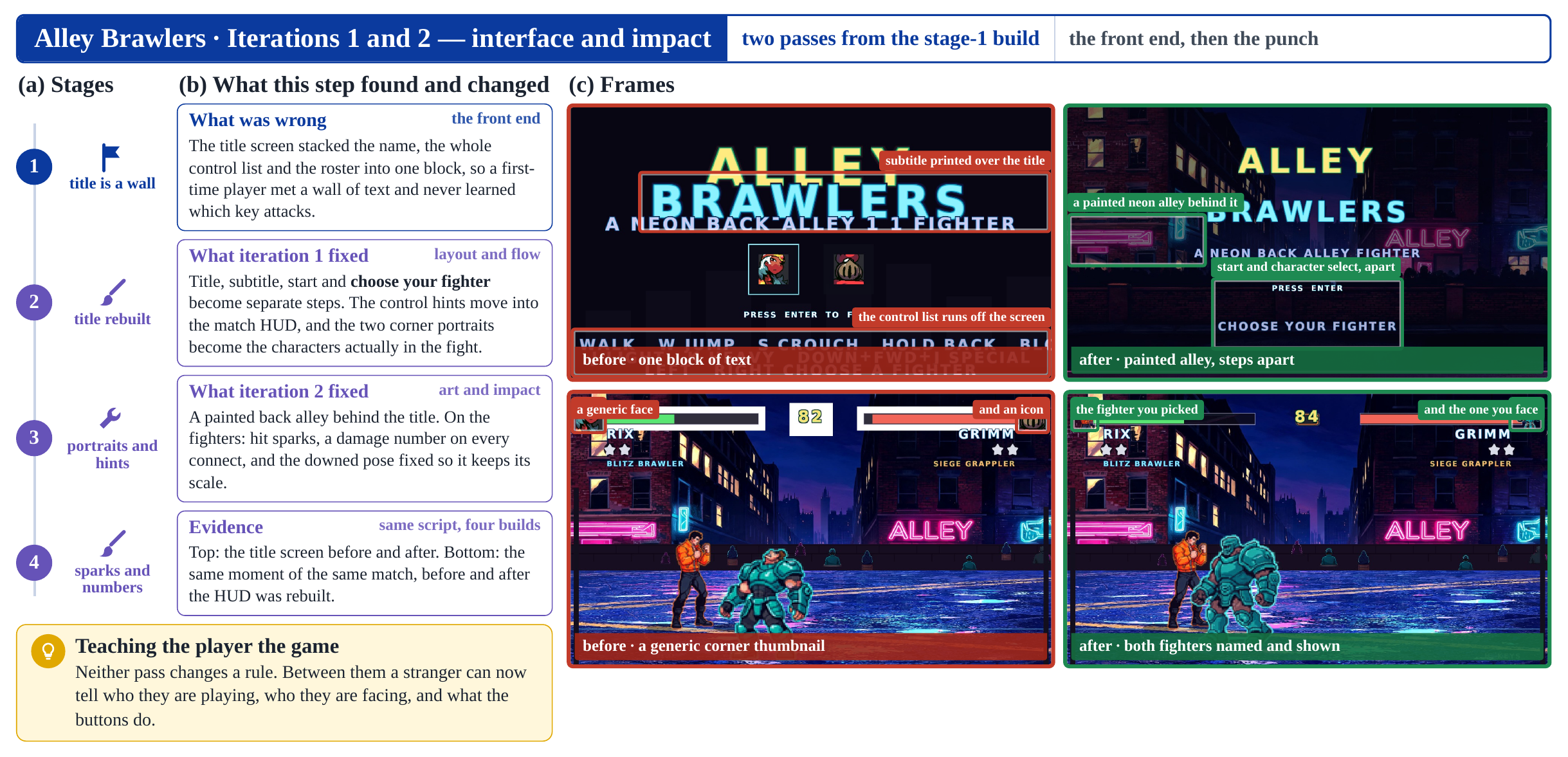}\\[3pt]
  \includegraphics[width=0.94\linewidth]{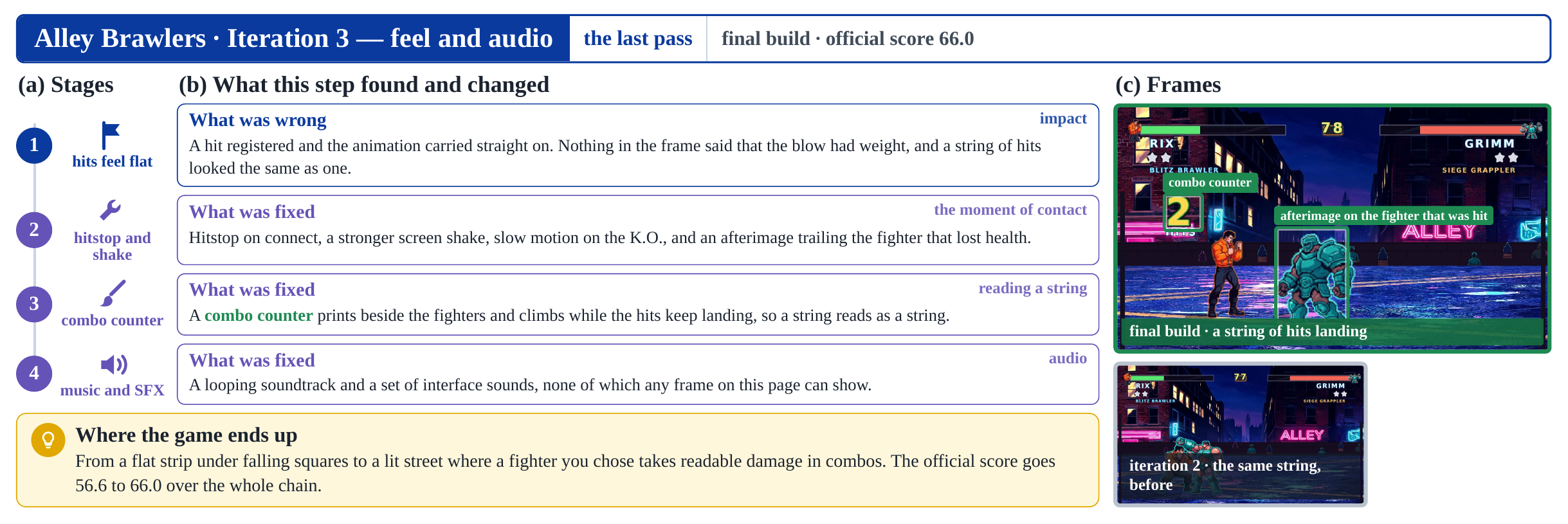}
  \caption{The second brief and the development passes, both continuing from the
  stage-1 champion. \textbf{Stage 2:} a second director names spacing as the
  reason a player cannot act, and the exchange becomes legible from both sides.
  \textbf{Iterations 1--2} rebuild the front end --- the title becomes steps
  rather than a wall of text, and the HUD names and shows both fighters --- and
  add hit sparks and damage numbers. \textbf{Iteration 3} is the feel of
  contact: hitstop, screen shake, K.O. slow motion, a combo counter and a
  soundtrack.}
  \label{fig:case_alley_b}
\end{figure}

\begingroup\phantomsection\label{fig:case_alley_c}\endgroup

\subsubsection{A Game the Loop Cannot Improve}
\label{app:case_cascade}

\emph{Block Cascade} is the opposite starting point. Seven-bag randomiser,
next-three preview, ghost piece, hold swap, wall kicks, line clears, top-out and
a best score all work in the generated build, and thirty rounds of evidence-led
improvement find nothing worth swapping to. The loop is not stuck on a defect; it has
run out of anything to call one.

Restarting from that build is what pays, and the briefs are not what makes it
pay: the two brief-guided branches and the control branch with no brief all
reach 50.0, and what the briefs buy is arriving there at round 6 rather than
round 12. The second brief then asks for stakes --- a named mode, a live clock,
a reward ladder --- and gets them, along with a collision between the new
readouts and the ones already on screen, which the pass after it lays out.

\begin{figure}[H]
  \centering
  \includegraphics[width=\linewidth]{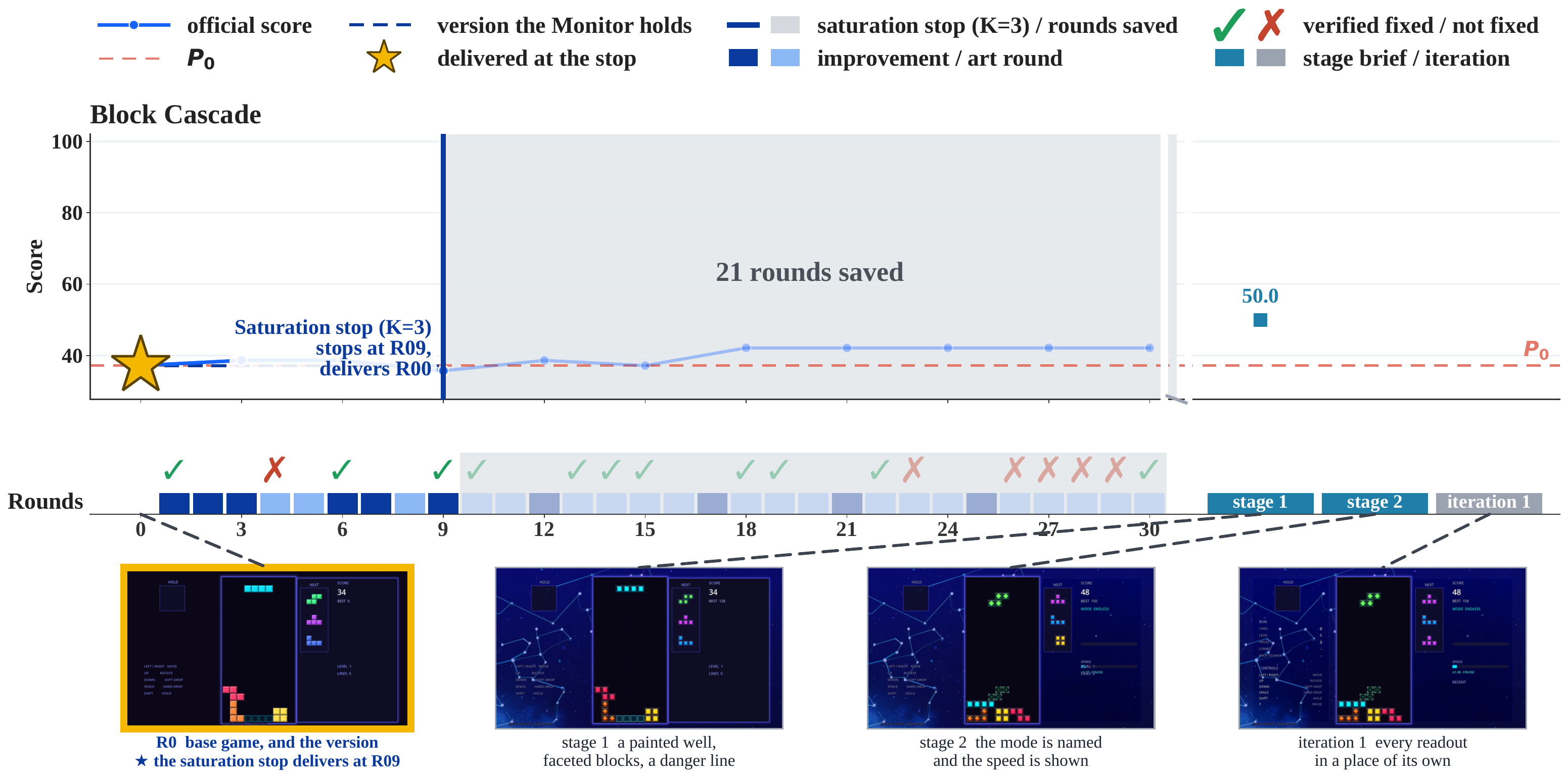}
  \caption{\emph{Block Cascade}, from the generated game to the build we
  shipped. The generated game already implements every mechanic in the spec, so
  the improvement rounds have almost nothing to improve: through nine checkpoints the
  Global Quality Monitor never prefers a later build, the saturation stop ends the run at round 9,
  and what it delivers is round 0 itself. Everything past the axis break follows
  a restart from that build.}
  \label{fig:case_cascade_overview}
\end{figure}

\begin{figure}[H]
  \centering
  \includegraphics[width=0.94\linewidth]{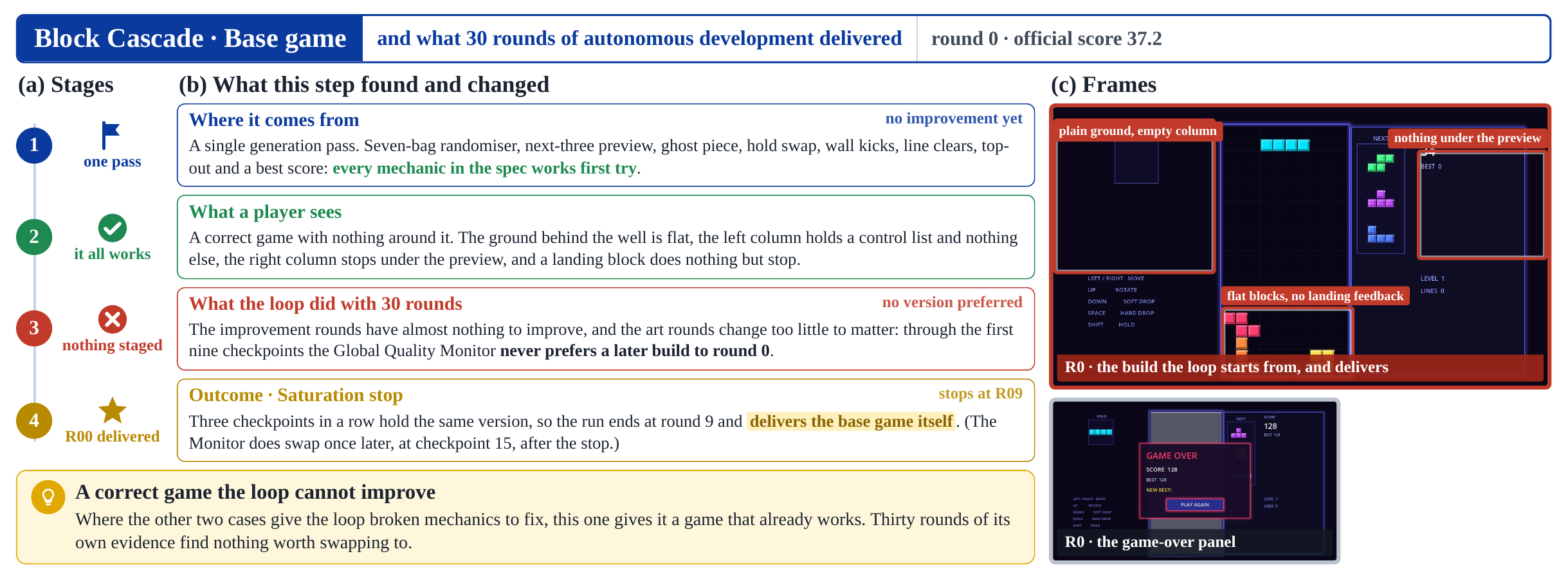}\\[3pt]
  \includegraphics[width=0.94\linewidth]{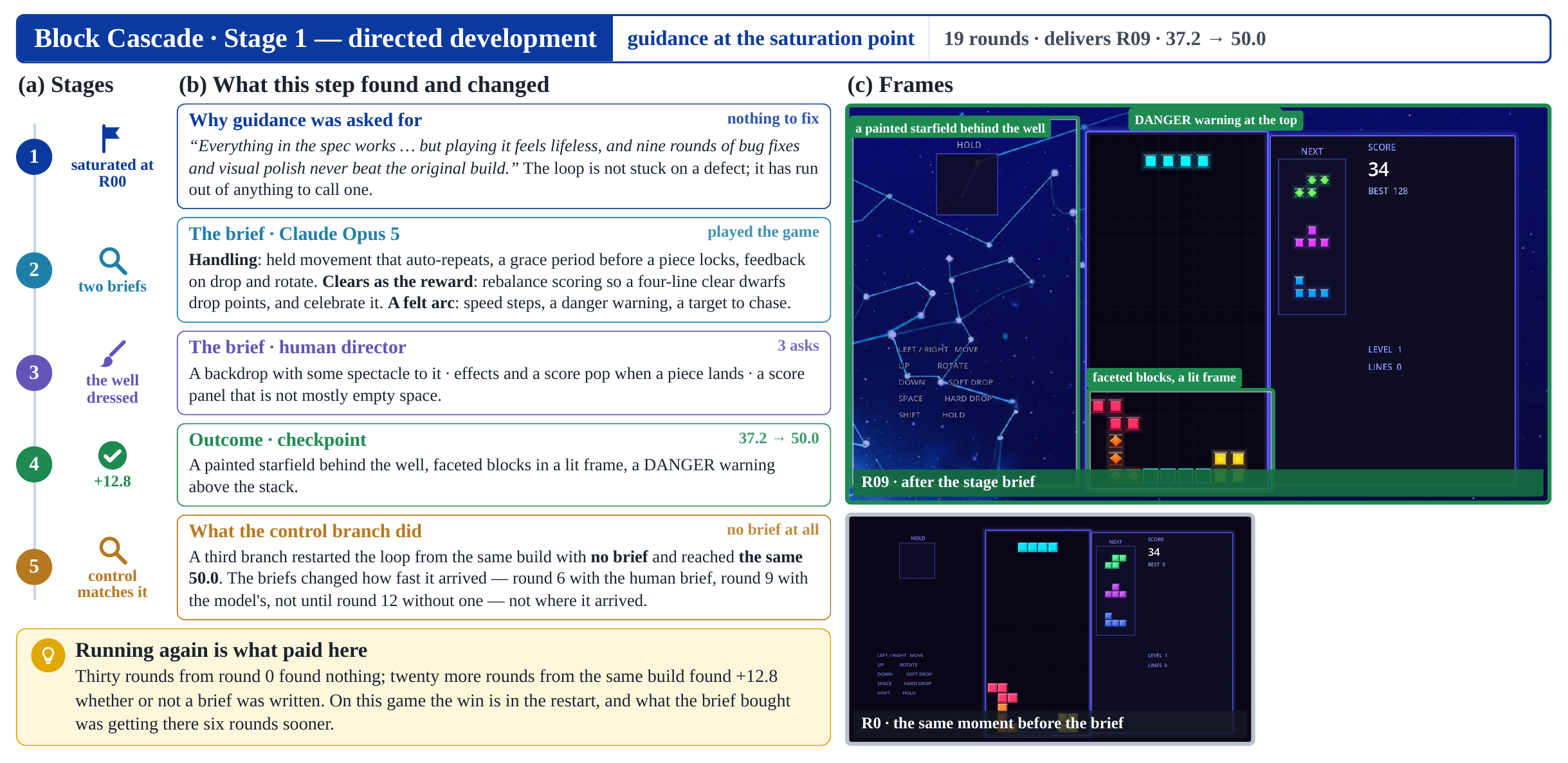}\\[3pt]
  \includegraphics[width=0.94\linewidth]{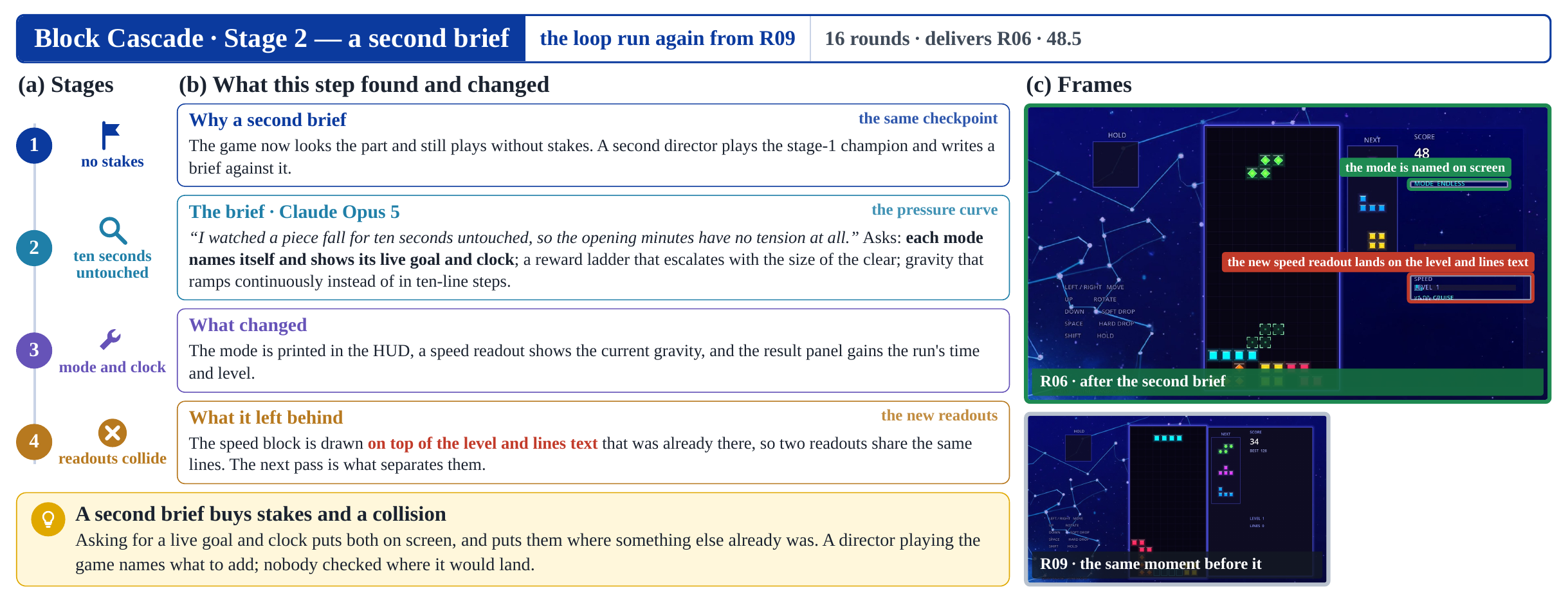}
  \caption{\emph{Block Cascade} from the generated build through both briefs. Cards,
  boxes and script as in Figure~\ref{fig:case_lawn_stages_a}. \textbf{Base
  game:} every mechanic in the spec works first try, and nine rounds later the
  loop delivers this same build. \textbf{Stage 1:} the well gets a painted
  ground, faceted blocks and a danger line --- and a control branch with no
  brief reaches the same score. \textbf{Stage 2:} a named mode and a live speed
  readout arrive, printed on top of the readouts already there.}
  \label{fig:case_cascade_a}
\end{figure}

\begin{figure}[H]
  \centering
  \includegraphics[width=0.94\linewidth]{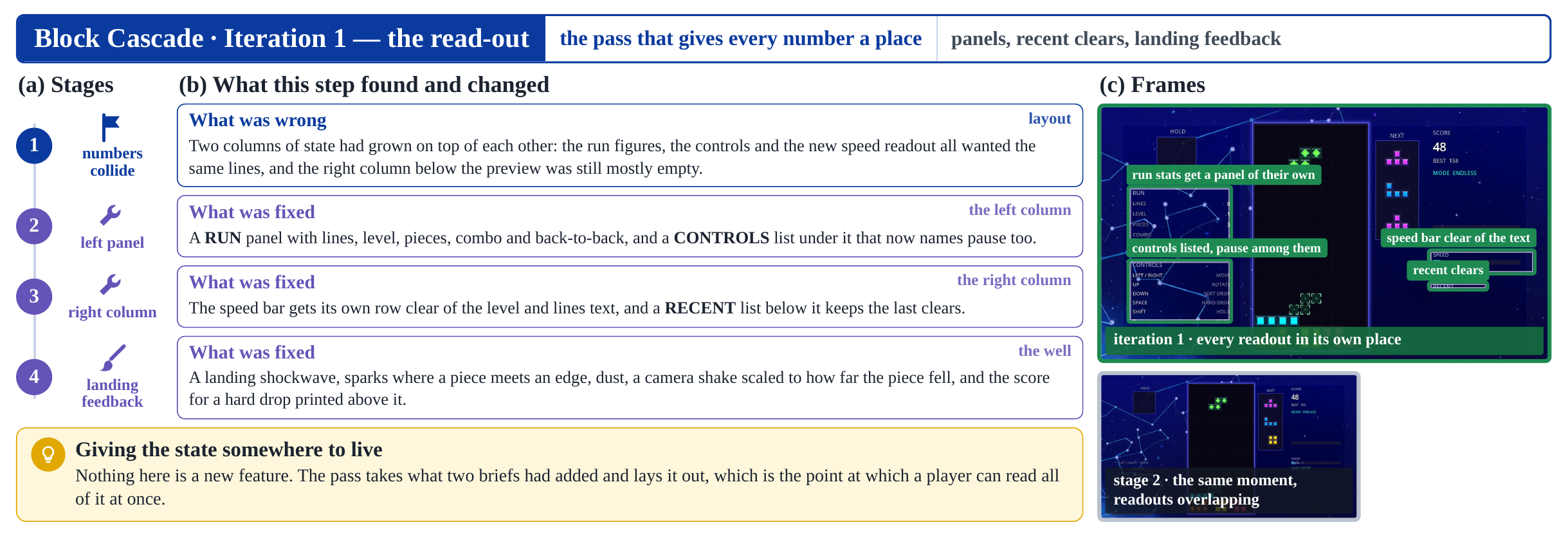}\\[3pt]
  \includegraphics[width=0.94\linewidth]{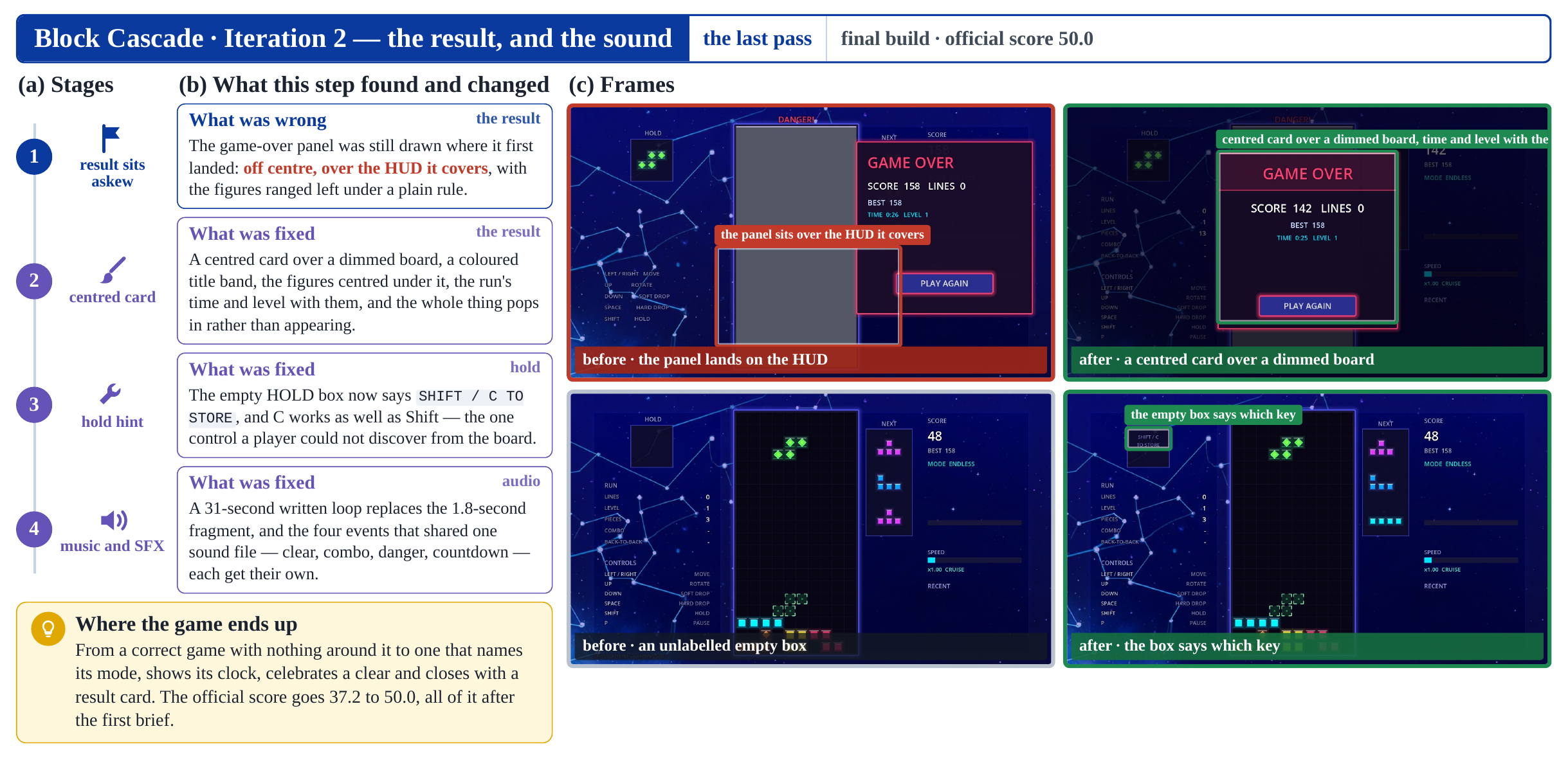}
  \caption{The two development passes that close the run.
  \textbf{Iteration 1} gives every number a place of its own and a landing its
  feedback. \textbf{Iteration 2} adds a centred result card over a dimmed board,
  a HOLD box that says which key fills it, and a written soundtrack in place of
  a 1.8-second fragment.}
  \label{fig:case_cascade_b}
\end{figure}

\begingroup\phantomsection\label{fig:case_cascade_c}\endgroup


\subsection{Per-Family Results}
\label{app:families}

\begin{table}[h]
\centering
\caption{Per-family breakdown on Godot for the Codex $+$ GPT-5.5 (high)
generator (Table~\ref{tab:godot_main}, first row group), by mean Overall ($\uparrow$), Qwen3.8-27B
judge. $N$ is the number of tasks in the family (scored/total while scoring is in
progress); families are ordered by base score; a task with no generated project
scores 0. Abbreviations: \textbf{+P2C} Play2Code;
--: not yet available on all tasks.}
\label{tab:family-godot-gpt}
\small

\begin{tabular*}{\textwidth}{@{\extracolsep{\fill}}lrccc}
\toprule
Family & $N$ & Base $\uparrow$ & +P2C & +\rsigame{} \\
\midrule
idle           &     4 & 81.9 & 86.8 & 87.5 \\
sports         &     4 & 72.8 & 70.5 & 81.2 \\
horror         &     5 & 67.2 & 71.4 & 75.1 \\
openworld      &    15 & 59.2 & 57.9 & 62.2 \\
tycoon         &    16 & 53.3 & 54.4 & 73.2 \\
visualnovel    &    11 & 52.0 & 55.8 & 71.5 \\
roguelike      &    14 & 46.1 & 38.5 & 56.8 \\
shooter        &     7 & 46.0 & 49.7 & 64.1 \\
strategy       &    17 & 46.0 & 46.3 & 61.5 \\
rhythm         &     5 & 45.4 & 44.7 & 67.9 \\
cardgame       &     5 & 44.2 & 44.3 & 71.3 \\
puzzle         &     8 & 43.4 & 45.1 & 53.9 \\
platformer     &    19 & 43.3 & 44.4 & 58.4 \\
racing         &     4 & 43.3 & 53.4 & 68.4 \\
simulation     &     6 & 38.1 & 37.7 & 48.8 \\
\midrule
\emph{All}     & 140 & 50.26 & 50.74 & 64.53 \\
\bottomrule
\end{tabular*}
\end{table}

\begin{table}[h]
\centering
\caption{Per-family breakdown on Godot for the Kimi-K2.6 generator
(Table~\ref{tab:godot_main}, second row group).}
\label{tab:family-godot-kimi}
\small

\begin{tabular*}{\textwidth}{@{\extracolsep{\fill}}lrccc}
\toprule
Family & $N$ & Base $\uparrow$ & +P2C & +\rsigame{} \\
\midrule
idle           &     4 & 53.5 & 50.1 & 60.0 \\
sports         &     4 & 45.9 & 57.3 & 64.4 \\
horror         &     5 & 43.1 & 56.6 & 63.4 \\
tycoon         &    16 & 36.1 & 40.7 & 46.4 \\
rhythm         &     5 & 33.7 & 37.1 & 45.4 \\
shooter        &     7 & 33.4 & 35.9 & 39.1 \\
simulation     &     6 & 30.7 & 29.8 & 34.9 \\
visualnovel    &    11 & 30.1 & 36.8 & 45.3 \\
roguelike      &    14 & 29.4 & 35.1 & 44.5 \\
racing         &     4 & 29.1 & 39.4 & 41.8 \\
openworld      &    15 & 27.0 & 33.6 & 42.5 \\
cardgame       &     5 & 26.2 & 23.6 & 48.8 \\
strategy       &    17 & 24.1 & 33.1 & 47.7 \\
platformer     &    19 & 23.3 & 29.2 & 41.4 \\
puzzle         &     8 & 15.3 & 19.4 & 29.2 \\
\midrule
\emph{All}     & 140 & 29.63 & 35.19 & 44.77 \\
\bottomrule
\end{tabular*}
\end{table}

\begin{table}[h]
\centering
\caption{Per-family breakdown on Godot for the GLM-5.3-Flash generator
(Table~\ref{tab:godot_main}, third row group).}
\label{tab:family-godot-glm}
\small

\begin{tabular*}{\textwidth}{@{\extracolsep{\fill}}lrccc}
\toprule
Family & $N$ & Base $\uparrow$ & +P2C & +\rsigame{} \\
\midrule
sports         &     4 & 49.6 & 61.5 & 72.6 \\
idle           &     4 & 39.0 & 59.2 & 68.6 \\
tycoon         &    16 & 38.8 & 43.1 & 56.1 \\
racing         &     4 & 34.9 & 36.7 & 53.0 \\
horror         &     5 & 34.1 & 43.6 & 49.9 \\
visualnovel    &    11 & 32.7 & 38.8 & 59.0 \\
shooter        &     7 & 32.6 & 42.5 & 54.9 \\
strategy       &    17 & 32.2 & 39.2 & 42.5 \\
simulation     &     6 & 32.0 & 38.6 & 46.5 \\
rhythm         &     5 & 26.3 & 37.2 & 52.9 \\
cardgame       &     5 & 25.5 & 33.1 & 52.7 \\
platformer     &    19 & 25.3 & 35.7 & 48.7 \\
roguelike      &    14 & 24.9 & 37.5 & 40.9 \\
openworld      &    15 & 19.2 & 28.1 & 43.2 \\
puzzle         &     8 & 16.9 & 33.3 & 40.8 \\
\midrule
\emph{All}     &   140 & 29.46 & 38.59 & 49.72 \\
\bottomrule
\end{tabular*}
\end{table}

\begin{table}[h]
\centering
\caption{Per-family breakdown on Godot for the Qwen3.8-27B generator
(Table~\ref{tab:godot_main}, fourth row group).}
\label{tab:family-godot-qwen}
\small

\begin{tabular*}{\textwidth}{@{\extracolsep{\fill}}lrccc}
\toprule
Family & $N$ & Base $\uparrow$ & +P2C & +\rsigame{} \\
\midrule
horror         &     5 & 58.7 & 56.6 & 57.0 \\
idle           &     4 & 48.8 & 53.2 & 73.9 \\
cardgame       &     5 & 48.3 & 53.1 & 59.3 \\
sports         &     4 & 44.8 & 51.2 & 55.6 \\
shooter        &     7 & 42.3 & 44.7 & 61.5 \\
visualnovel    &    11 & 42.0 & 46.0 & 44.9 \\
tycoon         &    16 & 41.0 & 52.0 & 62.6 \\
strategy       &    17 & 39.5 & 35.2 & 44.9 \\
openworld      &    15 & 34.5 & 39.7 & 36.4 \\
rhythm         &     5 & 33.6 & 35.0 & 54.7 \\
simulation     &     6 & 32.1 & 36.4 & 33.3 \\
platformer     &    19 & 31.6 & 31.7 & 43.5 \\
roguelike      &    14 & 31.0 & 36.7 & 42.4 \\
puzzle         &     8 & 23.5 & 32.5 & 43.5 \\
racing         &     4 & 23.3 & 22.7 & 27.9 \\
\midrule
\emph{All}     &   140 & 37.07 & 40.53 & 47.77 \\
\bottomrule
\end{tabular*}
\end{table}

\begin{table}[h]
\centering
\caption{Per-family breakdown on Godot for the Qwen3.8-27B (SFT) generator
(Table~\ref{tab:godot_main}, fifth row group).}
\label{tab:family-godot-sft}
\small

\begin{tabular*}{\textwidth}{@{\extracolsep{\fill}}lrccc}
\toprule
Family & $N$ & Base $\uparrow$ & Play2Code $\uparrow$ & \rsigame{} $\uparrow$ \\
\midrule
idle           &  4  & 80.4 & 78.6 & 84.1 \\
sports         &  4  & 73.3 & 68.7 & 76.1 \\
horror         &  5  & 60.6 & 62.4 & 71.5 \\
tycoon         & 16  & 54.7 & 56.4 & 66.3 \\
strategy       & 17  & 52.1 & 51.0 & 62.0 \\
visualnovel    & 11  & 50.2 & 48.9 & 58.4 \\
openworld      & 15  & 48.1 & 46.7 & 58.2 \\
roguelike      & 14  & 46.4 & 47.8 & 60.2 \\
platformer     & 19  & 42.3 & 46.9 & 56.0 \\
racing         &  4  & 41.3 & 38.5 & 52.1 \\
puzzle         &  8  & 40.6 & 44.8 & 60.3 \\
simulation     &  6  & 40.4 & 45.1 & 61.2 \\
rhythm         &  5  & 37.4 & 43.9 & 63.8 \\
cardgame       &  5  & 36.1 & 41.8 & 58.9 \\
shooter        &  7  & 35.3 & 39.8 & 55.2 \\
\midrule
\emph{All}     & 140 & 48.22 & 49.71 & 61.38 \\
\bottomrule
\end{tabular*}
\end{table}


\begin{table}[h]
\centering
\caption{Per-family breakdown on Phaser for the OpenGame $+$ GPT-5.5 generator
(Table~\ref{tab:phaser_main}, Phaser block, first row group).}
\label{tab:family-phaser}
\small

\begin{tabular*}{\textwidth}{@{\extracolsep{\fill}}lrccc}
\toprule
Family & $N$ & Base $\uparrow$ & +P2C & +\rsigame{} \\
\midrule
idle           &     4 & 72.7 & 73.8 & 78.2 \\
cardgame       &     5 & 68.9 & 68.9 & 61.8 \\
visualnovel    &    11 & 68.0 & 68.6 & 73.5 \\
tycoon         &    16 & 59.9 & 62.9 & 74.3 \\
sports         &     4 & 57.2 & 59.8 & 64.8 \\
roguelike      &    14 & 56.5 & 58.2 & 66.7 \\
horror         &     5 & 53.5 & 52.7 & 57.6 \\
simulation     &     6 & 51.8 & 57.7 & 55.4 \\
rhythm         &     5 & 50.8 & 52.5 & 58.2 \\
strategy       &    17 & 45.7 & 56.8 & 54.8 \\
shooter        &     7 & 45.2 & 58.6 & 60.4 \\
racing         &     4 & 39.8 & 60.4 & 52.8 \\
platformer     &    19 & 38.1 & 43.7 & 45.0 \\
openworld      &    15 & 32.8 & 41.0 & 39.5 \\
puzzle         &     8 & 32.5 & 40.6 & 51.6 \\
\midrule
\emph{All}     & 140 & 49.44 & 55.10 & 58.21 \\
\bottomrule
\end{tabular*}
\end{table}

\begin{table}[h]
\centering
\caption{Per-family breakdown on Phaser for the Qwen3.8-27B generator
(Table~\ref{tab:phaser_main}, Phaser block, second row group).}
\label{tab:family-phaser-qwen}
\small

\begin{tabular*}{\textwidth}{@{\extracolsep{\fill}}lrccc}
\toprule
Family & $N$ & Base $\uparrow$ & +P2C & +\rsigame{} \\
\midrule
sports         &     4 & 56.1 & 68.8 & 65.4 \\
horror         &     5 & 55.6 & 52.9 & 57.6 \\
simulation     &     6 & 55.3 & 53.3 & 67.8 \\
visualnovel    &    11 & 53.3 & 54.2 & 52.4 \\
rhythm         &     5 & 50.9 & 58.4 & 61.4 \\
tycoon         &    16 & 47.5 & 58.9 & 56.7 \\
cardgame       &     5 & 47.0 & 72.3 & 61.4 \\
puzzle         &     8 & 42.6 & 44.2 & 39.5 \\
idle           &     4 & 41.3 & 61.5 & 65.2 \\
racing         &     4 & 40.9 & 45.3 & 51.6 \\
strategy       &    17 & 37.3 & 50.0 & 50.6 \\
platformer     &    19 & 37.1 & 40.2 & 46.2 \\
roguelike      &    14 & 29.6 & 40.7 & 48.8 \\
shooter        &     7 & 27.7 & 46.0 & 48.1 \\
openworld      &    15 & 21.1 & 32.0 & 29.3 \\
\midrule
\emph{All}     & 140 & 40.03 & 48.70 & 50.24 \\
\bottomrule
\end{tabular*}
\end{table}

\begin{table}[h]
\centering
\caption{Per-family breakdown on Phaser for the Qwen3.8-27B (SFT) generator
(Table~\ref{tab:phaser_main}, Phaser block, third row group).}
\label{tab:family-phaser-qwen-sft}
\small

\begin{tabular*}{\textwidth}{@{\extracolsep{\fill}}lrccc}
\toprule
Family & $N$ & Base $\uparrow$ & Play2Code $\uparrow$ & \rsigame{} $\uparrow$ \\
\midrule
idle           &  4  & 73.0 & 75.2 & 78.5 \\
horror         &  5  & 56.0 & 60.7 & 67.5 \\
tycoon         & 16  & 53.3 & 62.0 & 66.5 \\
sports         &  4  & 51.6 & 49.9 & 56.8 \\
shooter        &  7  & 41.8 & 51.4 & 55.4 \\
rhythm         &  5  & 46.8 & 61.3 & 66.2 \\
simulation     &  6  & 46.0 & 50.7 & 60.1 \\
roguelike      & 14  & 45.5 & 56.8 & 60.2 \\
openworld      & 15  & 42.2 & 46.0 & 50.7 \\
puzzle         &  8  & 43.8 & 58.7 & 64.8 \\
strategy       & 17  & 42.9 & 51.3 & 54.9 \\
cardgame       &  5  & 41.6 & 45.1 & 58.7 \\
visualnovel    & 11  & 40.0 & 46.2 & 54.3 \\
platformer     & 19  & 39.9 & 50.4 & 54.1 \\
racing         &  4  & 32.8 & 39.1 & 49.6 \\
\midrule
\emph{All}     & 140 & 45.14 & 53.15 & 58.53 \\
\bottomrule
\end{tabular*}

\end{table}

\end{document}

%% file: math_commands.tex
\usepackage{amsmath,amsfonts,bm}

\def\eqref#1{equation~\ref{#1}}

\def\1{\bm{1}}

\DeclareMathAlphabet{\mathsfit}{\encodingdefault}{\sfdefault}{m}{sl}
\SetMathAlphabet{\mathsfit}{bold}{\encodingdefault}{\sfdefault}{bx}{n}

